\documentclass[nohyperref]{article}

\usepackage{microtype}
\usepackage{graphicx}
\usepackage{subfigure}
\usepackage{booktabs} 

\usepackage[backref]{hyperref}

\usepackage[dvipsnames]{xcolor}

\usepackage[accepted]{icml2022}

\usepackage{amsmath}
\usepackage{amssymb}
\usepackage{mathtools}
\usepackage{amsthm}

\usepackage[capitalize,noabbrev]{cleveref}

\usepackage{amsfonts}
\usepackage{tikz}
\usepackage{longtable}
\usetikzlibrary{fit,arrows,arrows.meta,calc,positioning}
\usepackage{dblfloatfix}

\newsavebox\CBox

\DeclareMathOperator*{\argmin}{arg\,min}

\def\genbox#1#2#3#4#5#6{
    \leavevmode\raise#4bp\hbox to#5bp{\vrule height#5bp depth0bp width0bp
    \pdfliteral{q .5 w \csname #2COLOR\endcsname\space RG
                       \csname #3PDF\endcsname{#5}{#6} S Q
             \ifx1#1 q \csname #2COLOR\endcsname\space rg 
                       \csname #3PDF\endcsname{#5}{#6} f Q\fi}\hss}}

\def\trianbox   #1#2{\genbox{#1}{#2}  {trian}    {0.5}   {5}    {2.5}}
\def\uptrianbox #1#2{\genbox{#1}{#2}  {uptrian}  {0.5}   {5}    {2.5}}

\theoremstyle{plain}
\newtheorem{theorem}{Theorem}[section]

\theoremstyle{definition}
\newtheorem{definition}[theorem]{Definition}

\theoremstyle{remark}

\usepackage[textsize=tiny]{todonotes}

\icmltitlerunning{Closed-Form Diffeomorphic Transformations for Time Series Alignment}

\begin{document}

\twocolumn[
\icmltitle{Closed-Form Diffeomorphic Transformations for Time Series Alignment}



\icmlsetsymbol{equal}{*}

\begin{icmlauthorlist}
\icmlauthor{I\~nigo Martinez}{VICOMTECH}
\icmlauthor{Elisabeth Viles}{TECNUN,ICDIA}
\icmlauthor{Igor G. Olaizola}{VICOMTECH}
\end{icmlauthorlist}

\icmlaffiliation{VICOMTECH}{Vicomtech Foundation, Basque Research and Technology Alliance (BRTA), San Sebastian, Spain}
\icmlaffiliation{TECNUN}{TECNUN School of Engineering, University of Navarra, San Sebastian, Spain}
\icmlaffiliation{ICDIA}{Institute of Data Science and Artificial Intelligence, University of Navarra, Pamplona, Spain}

\icmlcorrespondingauthor{I\~nigo Martinez}{imartinez@vicomtech.org}

\icmlkeywords{time series, nonlinear time warping, diffeomorphisms, machine learning, deep learning, nearest centroid classification, time series averaging, temporal distortion, time series analysis, sensitivity analysis, ordinary differential equations}

\vskip 0.3in
]



\printAffiliationsAndNotice{}  

\begin{abstract}
Time series alignment methods call for highly expressive, differentiable and invertible warping functions which preserve temporal topology, i.e diffeomorphisms. Diffeomorphic warping functions can be generated from the integration of velocity fields governed by an ordinary differential equation (ODE). Gradient-based optimization frameworks containing diffeomorphic transformations require to calculate derivatives to the differential equation's solution with respect to the model parameters, i.e. sensitivity analysis. Unfortunately, deep learning frameworks typically lack automatic-differentiation-compatible sensitivity analysis methods; and implicit functions, such as the solution of ODE, require particular care. Current solutions appeal to adjoint sensitivity methods, ad-hoc numerical solvers or ResNet's Eulerian discretization. In this work, we present a closed-form expression for the ODE solution and its gradient under continuous piecewise-affine (CPA) velocity functions. We present a highly optimized implementation of the results on CPU and GPU. Furthermore, we conduct extensive experiments on several datasets to validate the generalization ability of our model to unseen data for time-series joint alignment. Results show significant improvements both in terms of efficiency and accuracy.
\end{abstract}

\section{Introduction and Context}
\label{sec:introduction}

\begin{figure}[!ht]
    \vskip 0.2in
    \begin{center}
    \centerline{\includegraphics[width=\columnwidth]{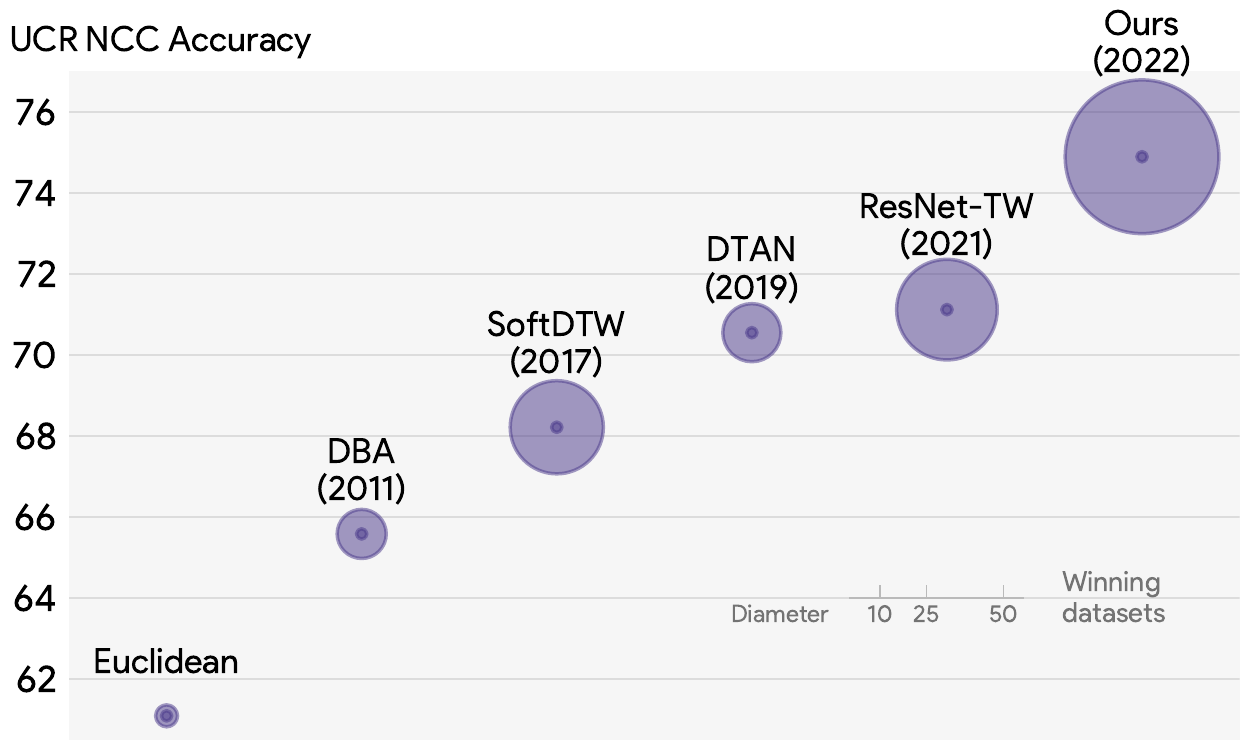}}
    \caption{
    Correct classification rates using NCC on UCR archive \cite{dau2019ucr}. Radius denotes the number of datasets in which each method achieved top accuracy. Our method was better or no worse than 94\% of the datasets compared with Euclidean, DBA (77\%), SoftDTW (69\%),  DTAN (76\%),  ResNet-TW (70\%).
    }
    \label{fig:ucr:accuracy}
    \end{center}
    \vskip -0.3in
\end{figure}

Time series data recurrently appears misaligned or warped in time despite exhibiting amplitude and shape similarity. 
Temporal misalignment, often caused by differences in execution, sampling rates, or the number of measurements, confounds statistical analysis to the point where even the sample mean of semantically-close time series can obfuscate actual peaks and valleys, generate non-existent features and be rendered almost meaningless. 
Indeed, ignoring temporal variability can greatly decrease recognition and classification performance \cite{veeraraghavan2009rate}.
 
\textbf{Time series alignment} can be solved by finding a set of optimal and plausible warping functions $\boldsymbol{\phi}$ between observations of the same process so that their temporal variability is minimized. 
Alignment methods have become critical for many applications involving time series data, such as bioinformatics, computer vision, industrial data, speech recognition \& synthesis, or human action recognition \cite{srivastava2016functional}.
Temporal warping functions extend from low dimensional global transformations described by a few parameters (e.g. affine transformations) to high dimensional non-rigid transformations specified at each point of the domain (e.g. diffeomorphisms). 
A ($C^1$) \textbf{diffeomorphism} is a differentiable invertible map with a differentiable inverse \cite{duistermaat2000lie}. Diffeomorphisms belong to the group of homeomorphisms that preserve topology by design, and thereby are a natural choice in the context of nonlinear time warping, where continuous, differentiable and invertible functions are preferred \cite{mumford2010pattern, bertrand2016}. 
Diffeomorphic warping functions $\boldsymbol{\phi}$ can be generated via integration of regular stationary or time-dependent velocity fields $v$ specified by the ordinary differential equation (ODE) $\partial\boldsymbol{\phi}/\partial t = v(\boldsymbol{\phi})$ \cite{Freifeld2017, Detlefsen2018, Ouderaa2021}. 

Temporal alignment problems have been widely stated as the estimation of diffeomorphic transformations between input and output data \cite{Huang2021}. 
Finding the optimal warping function for pairwise or joint alignment requires solving an optimization problem. 
However, aligning a batch of time series data is insufficient because new optimization problems arise as new data batches arrive. 
Inspired by Spatial Transformer Networks (STN), \textbf{Temporal Transformer Networks} (TTN) such as \cite{Weber2019,Lohit2019,Nunez2020} and recently \cite{Huang2021}, generalize inferred alignments from the original batch to the new data without having to solve a new optimization problem each time. 
Temporal transformer's neural network training proceeds via gradient descent on model parameters. 
Under such gradient-based optimization frameworks, neural networks that include diffeomorphic transformations require to calculate derivatives to the differential equation's solution with respect to the model parameters, i.e. sensitivity analysis. 
Unfortunately, deep learning frameworks typically lack automatic-differentiation-compatible sensitivity analysis methods; and implicit functions, such as the solution of ODE, require particular care. 
Current solutions appeal to adjoint sensitivity methods \cite{chen2019neural}, ResNet's Eulerian discretization  \cite{Huang2021} or ad-hoc numerical solvers and automatic differentiation \cite{Freifeld2017}. More on this in \cref{sec:related_work}.

In contrast, our proposal is to formulate the \textbf{closed-form}\footnote{
A closed-form expression may use
a finite number of standard operations 
($+,-,\times,\div$), and functions 
(e.g., $\sqrt[n]{\;}$, $\exp$, $\log$, $\sin$, $\sinh$)
, but no limit, differentiation, or integration.
\vspace{-6pt}
}
\textbf{expression for the ODE's diffeomorphic solution and its gradient} under continuous piecewise-affine (CPA) velocity functions. 
CPA velocity functions yield well-behaved parametrized diffeomorphic transformations, which are efficient to compute, accurate and highly expressive \cite{Freifeld2015}. These finite-dimensional transformations can handle optional constraints (zero-velocity at the boundary) and support convenient modeling choices such as smoothing priors and coarse-to-fine analysis. The term “piecewise" refers to a partition of the temporal domain into subintervals. The fineness of the partition controls the trade-off between expressiveness and computational complexity.
Unlike other spaces of highly-expressive velocity fields, integration of CPA velocity fields is given in either closed form (in 1D) or “essentially” closed form for higher dimensions \cite{Freifeld2017}. However, the gradient of CPA-based transformations is only available via the solution of a system of coupled integral equations \cite{freifeld2018deriving}.
Following the first principle of automatic differentiation --- \emph{if the analytical solution to the derivative is known, then replace the function with its derivative}---, in this work we present a closed-form solution for the ODE gradient in 1D that is not available in the current literature. 

A closed-form solution provides efficiency, speed and precision. Fast computation is essential if the transformation must be repeated several times, as is the case with iterative gradient descent methods used in deep learning training. Furthermore, a precise (exact) gradient of the transformation translates to efficient search in the parameter space, which leads to better solutions at convergence. 
Encapsulating the ODE solution (forward operation) and its gradient (backward operation) under a closed-form transformation block shortens the chain of operations and decreases the overhead of managing a long tape with lots of scalar arithmetic operations. Indeed, explicitly implementing the backward operation is more efficient than letting the automatic differentiation system naively differentiate the closed-form forward function. 

Furthermore, we integrate closed-form diffeomorphic transformations into a temporal transformer network (see \cref{fig:architecture}) for time series alignment. Similarly to \cite{Weber2019}, we formulate the joint alignment problem as to simultaneously compute the centroid and align all sequential data within a class, under a semi-supervised schema.

\subsection*{Contributions}

\begin{itemize}
    \setlength{\itemsep}{6pt}
    \setlength{\parskip}{0pt}
    \setlength{\parsep}{0pt}
    \item We propose a novel closed-form expression for the gradient of CPA-based one-dimensional diffeomorphic transformations, providing efficiency, speed and precision. 
    \item We present Diffeomorphic Fast Warping (\textit{DIFW}), an open-source library and highly optimized implementation of 1D diffeomorphic transformations on multiple backends for CPU (\textit{NumPy} and \textit{PyTorch} with \textit{C++}) and GPU (\textit{PyTorch} with \textit{CUDA}). 
    Speed tests show an x18 (\textcolor{Sepia}{x260}) and x10 (\textcolor{Sepia}{x30}) improvement on CPU (\textcolor{Sepia}{GPU}) over current solutions for forward and backward operations respectively.
    \item We incorporate these improvements into a diffeomorphic temporal transformer network resembling \cite{Weber2019} for time-series alignment and classification.
    \item We conduct extensive experiments on 84 datasets from the UCR archive \cite{dau2019ucr} to validate the generalization ability of our model to unseen data for time-series joint alignment. Results show significant improvements both in terms of efficiency and accuracy (see \cref{fig:ucr:accuracy}).
\end{itemize}
\clearpage
The paper is structured as follows.
We discuss related work in \cref{sec:related_work} and outline the mathematical formalism relevant to our method in \cref{sec:method}.
The experiments with extensive results are described in \cref{sec:results} and final remarks are included in \cref{sec:conclusions}.

\section{Related Work}
\label{sec:related_work}

\subsection{Time Series Alignment}\label{sec:time_series_alignment}

\textbf{Pairwise alignment:}\label{sec:pairwise_alignment}
Given two time series observations $\textbf{y}$ and $\textbf{z}$: $\Omega \rightarrow \mathbb{R}^d$, where $\Omega \subseteq \mathbb{R}$ is the temporal domain, the goal of pairwise alignment is to compute an optimal transformation $\boldsymbol{\phi}^{*}: \Omega \rightarrow \Omega$, such than the data fit between the query $\textbf{y}$ and transformed target $\textbf{z} \circ \boldsymbol{\phi}$ is high, that is, their difference according to a distance measure $\mathcal{D}$ is minimum\footnote{The $\circ$ operator refers to function composition\vspace{-14pt}}.
This problem is typically solved as an optimization problem. The “goodness” of the transformation is measured by a cost of the form:
$\boldsymbol{\phi}^{*} = \argmin_{\boldsymbol{\phi} \in \Phi} \mathcal{D}(\textbf{y}, \textbf{z} \circ \boldsymbol{\phi}) + \mathcal{R}(\boldsymbol{\phi})$.
The set $\Phi$ is the set of all monotonically-increasing functions from $\Omega$ to itself. 
The first term $\mathcal{D}(\textbf{y}, \textbf{z} \circ \boldsymbol{\phi})$ evaluates the similarity between $\textbf{y}$ and $\textbf{z} \circ \boldsymbol{\phi}$, whereas the second term $\mathcal{R}(\boldsymbol{\phi})$ imposes constraints on the warping function $\boldsymbol{\phi}$, as smoothness, monotonicity preserving and boundary conditions.

Dynamic Time Warping (DTW) \cite{sakoe1978dynamic} is a popular method that finds the optimal alignment by first computing a pairwise distance matrix and then solving a dynamic program using Bellman's recursion with a quadratic cost. However, DTW is not differentiable everywhere, is sensitive to noise and leads to bad local optima when used as a loss \cite{blondel2021differentiable}. SoftDTW \cite{cuturi2017soft} is a differentiable loss function that replaces the minimum over alignments in DTW with a soft minimum, which induces a probability distribution over alignments.

\textbf{Joint alignment:}\label{sec:joint_alignment}
Given a set of $N$ time series observations $Y=\{y_i\}_{i=1}^N$ the goal of joint alignment is to find a set of optimal warping functions $\{\phi_{i}^{*}\}_{i=1}^N$ such that the average sequence, $\bar{\textbf{y}}=\argmin_{y} \sum_{i=1}^{N} \mathcal{D}(y,y_i)$, minimizes the sum of distance to all elements in the set $Y$: $\{\boldsymbol{\phi}_{i}^{*}\}_{i=1}^{N} = \argmin_{\boldsymbol{\phi}_i \in \Phi} \sum_{i=1}^{N} \mathcal{D}(\bar{\textbf{y}}, \textbf{y}_i \circ \boldsymbol{\phi}_i) + \mathcal{R}(\boldsymbol{\phi}_i)$. 

DTW Barycenter Averaging (DBA) \cite{Petitjean2011-DBA} minimizes the sum of the DTW discrepancies to the input time-series and iteratively refines an initial average sequence. 
Generalized Time Warping (GTW) \cite{Zhou2012} approximates the optimal temporal warping by a linear combination of monotonic basis functions and a Gauss-Newton-based method is used to learn the weights of the basis functions.
However, GTW requires a large number of complex basis functions to be effective.
\cite{kawano2020neural} relaxed the DTW-based discrete formulation of the joint alignment problem to a continuous optimization in which a neural network learns the optimal warping functions.
Also related is the square-root velocity function (SRVF) representation \cite{Srivastava2011} for analyzing shapes of curves in euclidean spaces under an elastic metric that is invariant to reparametrization. 
Note that these solutions are cast as an optimization problem, and as a result, lack a generalization mechanism and must compute alignments from scratch for new data. 

\textbf{Temporal Transformer Networks} can generalize inferred alignments from the original batch to the new data without having to solve a new optimization problem each time. 
Diffeomorphisms have been actively studied in the field of image registration and have been successfully applied in a variety of methods \cite{Beg2005,vercauteren2009diffeomorphic,dalca2018unsupervised,fu2020deep}, 
such as the log-Euclidean polyaffine method \cite{Arsigny2006, Arsigny2006a}.
Hereof \cite{Detlefsen2018} first implemented flexible CPA-based diffeomorphic image transformations within STN, and were later extended to variational autoencoders \cite{Detlefsen2019}. 
On this subject, \cite{Ouderaa2021} presented an STN with diffeomorphic transformations and used the scaling-and-squaring method for solving 2D stationary velocity fields and the Baker-Campbell-Hausdoorff formula for solving time-dependent velocity fields. 

The counterpart STN model for time series alignment is Diffeomorphic Temporal Alignment Nets (DTAN) \cite{Weber2019} and its recurrent variant R-DTAN. DTAN closely resembles the TTN model presented in this paper, even though the core diffeomorphic transformations are computed differently, as presented in \cref{sec:method}.
In this regard, \cite{Lohit2019} generated input-dependent warping functions that lead to rate-robust representations, reduce intra-class variability and increase inter-class separation.

Also related is \cite{Rousseau2019}, a residual network for the numerical approximation of exponential diffeomorphic operators. The velocity field in this model is a linear combination of basis functions which are parametrized with convolutional and \textit{ReLu} layers.
Very recently, \cite{Huang2021} proposed a residual network (ResNet-TW) that echoes an Eulerian discretization of the flow equation (ODE) for CPA time-dependent vector fields to build diffeomorphic transformations.
As an alternative approach, \cite{Nunez2020} proposed a deep neural network for learning the warping functions directly from DTW matches, and used it to predict optimal diffeomorphic warping functions.
Regarding the SRVF framework, \cite{Nunez2021} presented an unsupervised generative encoder-decoder architecture (SRVF-Net) to produce a distribution space of SRVF warping functions for the joint alignment of functional data.
Beyond diffeomorphic transformations, Trainable Time Warping \cite{Khorram2019} performs alignment in the continuous-time domain using a \textit{sinc} convolutional kernel and a gradient-based optimization technique.

\subsection{Computing Derivatives to ODE's solution}\label{sec:computing_derivatives}
In this section we review available solutions to compute derivatives to the ODE's solution:

\textbf{Numerical differentiation} should be avoided because it requires two numerical ODE solutions for each parameter (very inefficient) and is prone to numerical error: if the step size is too small, it may exhibit floating point cancellation; if the step size is chosen too large, then the error term of the approximation is large. 

\textbf{Forward-mode continuous sensitivity analysis} extends the ODE system and studies the model response when each parameter is varied while holding the rest at constant values. However, since the number of ODEs in the system scales proportionally with the number of parameters, this method is impractical for systems with a large number of parameters.

\textbf{Residual networks} (ResNets) can be interpreted as discrete numerical integrators of continuous dynamical systems, with each residual unit acting as one step of Euler's forward method. ResNet-TW \cite{Huang2021} use this Eulerian discretization schema of the ODE to generate smooth and invertible transformations. The gradient is computed using reverse-mode AD backpass, also known as backpropagation.
However, ResNets are difficult to compress without significantly decreasing model accuracy, and do not learn to represent continuous dynamical systems in any meaningful sense \cite{queiruga2020continuousindepth}. Even though highly expressive diffeomorphic transformations can be generated via non-stationary velocity functions, the integration error is proportional to the number of residual units. Computing precise ODE solutions with ResNets require increasing the number of layers and as a result, the memory use of the model.

Neural ODEs \cite{chen2019neural}, the continuous version of ResNets, compute gradients via \textbf{continuous adjoint sensitivity analysis}, i.e. solving a second augmented ODE backwards in time. The efficiency issue with adjoint sensitivity analysis methods is that they require multiple forward ODE solutions, which can become prohibitively costly in large models. Neural ODEs reduce the computational complexity to a single solve, while retaining low memory cost by solving the adjoint gradients jointly backward-in-time alongside the ODE solution. 
However, this method implicitly makes the assumption that the ODE integrator is time-reversible and sadly there are no reversible adaptive integrators for first-order ODEs solvers, so this method diverges on some systems \cite{rackauckas2019diffeqfluxjl}. Other notable approaches for solving the adjoint equations with different memory-compute trade-offs are
interpolation schemes \cite{daulbaev2020interpolation},
symplectic integration \cite{zhuang2021mali},
storing intermediate quantities \cite{zhang2014fatode} and
checkpointing \cite{zhuang2020adaptive}. 

\textbf{Discrete sensitivity analysis} calculates model sensitivities by directly differentiating the numerical method's steps. Yet, this approach requires specialized implementation of the first-order ODE solvers to propagate said derivatives. Automatic differentiation (AD) can be used on a solver implemented fully in a language with AD (a differential programming approach) \cite{ma2021comparison}. However, pure tape-based reverse-mode AD software libraries (such as PyTorch \cite{paszke2019pytorch}, ReverseDiff.jl \cite{reverseDiff} and Tensorflow Eager \cite{agrawal2019tensorflow}), 
have been generally optimized for large linear algebra and array operations, which decrease the size of the tape in relation to the amount of work performed. Given that ODEs are typically defined by nonlinear functions with scalar operations, the tape handling to work ratio gets cut down and is no longer competitive with other forms of derivative calculations \cite{ma2021comparison}.
Other frameworks, such as JAX \cite{jax2018github} cannot JIT optimize the non-static computation graphs of a full ODE solver. These issues may be addressed in the next generation reverse-mode source-to-source AD packages like Zygote.jl \cite{Zygote.jl-2018} or Enzyme.jl \cite{NEURIPS2020_9332c513} by not relying on tape generation. 


\section{Method}
\label{sec:method}

Time series alignment seeks to find a plausible warping function $\phi$ that minimizes the observed temporal variability. We propose spaces of diffeomorphic warping transformations that are based on fast and exact integration of CPA velocity fields. For this section, we inherit the notation used by \cite{Freifeld2015,Freifeld2017}, who originally proposed CPA-based diffeomorphic transformations.

\subsection{CPA Velocity Functions}\label{sec:cpa_velocity_functions}

Let $\Omega \subseteq \mathbb{R}$ be the Cartesian product of 1D compact intervals that represents the temporal domain. 
\begin{definition}\label{def:tessellation}
A finite tessellation $\mathcal{P} = \{U_{c}\}_{c=1}^{N_\mathcal{P}}$ is a set of $N_\mathcal{P}$ closed subsets of $\Omega$, also called cells $U_c$, such that their union is $\Omega$ and the intersection of any pair of adjacent cells is their shared border. 
Notation: $N_\mathcal{P}$ is the number of cells, $N_v = N_p + 1$ number of vertices and $N_e = N_v - 2$ number of shared vertices (all three quantities positive integers). 
\end{definition}

Fix a tessellation $\mathcal{P}$, let $x \in \Omega$ and define the membership function $\gamma: \Omega \rightarrow \{1,...,N_\mathcal{P}\}$, $\gamma: x \rightarrow \text{min}\{c: x \in U_c\}$. If $x$ is not on an inter-cell border, then $\gamma(x) = c \leftrightarrow x \in U_c$.

\begin{definition}\label{def:piecewise_affine}
A map $f: \Omega \rightarrow \mathbb{R}$ is called piecewise affine (PA) w.r.t $\mathcal{P}$ if $\{f|_{U_c}\}_{c=1}^{N_\mathcal{P}}$ are affine, i.e., $f(x) = \textbf{A}_{\gamma(x)} \tilde{x}$ where $\tilde{x} \stackrel{\Delta}{=} \begin{bmatrix} x \\ 1 \end{bmatrix} \in \mathbb{R}^2 ,\quad \textbf{A}_c \in \mathbb{R}^{1 \times 2} \quad \forall c \in \{1,...,N_\mathcal{P}\}$.
Let's define $\textbf{A}_c = \begin{bmatrix} a_c & b_c \end{bmatrix} \in \mathbb{R}^{2 \times 1}$, then $f(x) = a_{\gamma(x)}x + b_{\gamma(x)} = a_c x + b_c$. 
\end{definition}

\begin{figure}[t]
    \begin{center}
    \scalebox{0.9}{
    \begin{tikzpicture}
        \pgfmathsetmacro{\N}{6};
        \pgfmathsetmacro{\M}{5};
        \pgfmathsetmacro{\P}{\N-1};
        \pgfmathsetmacro{\h}{-0.5};
        \draw[stepx=1, stepy=1, black!30, thin] (0,0) grid (\N, \M);
        \draw[-latex, very thick] (-0.1,0) -- (\N,0) node[right] {$x$};
        \draw[-latex, very thick] (0,-0.1) -- (0,\M) node[above] {$v(x)$};

        \draw (0,0.25+\h) -- (0,-0.25+\h)  node[anchor=north, pos=1.25] {$x_{1}$};
        \draw (1,0.25+\h) -- (1,-0.25+\h)  node[anchor=north, pos=1.25] {$x_{2}$};
        \draw (2,0.25+\h) -- (2,-0.25+\h)  node[anchor=north, pos=1.25] {$x_{3}$};
        \draw (3,0.25+\h) -- (3,-0.25+\h)  node[anchor=north, pos=1.25] {$x_{c}$};
        \draw (4,0.25+\h) -- (4,-0.25+\h)  node[anchor=north, pos=1.25] {$x_{c+1}$};
        \draw (5,0.25+\h) -- (5,-0.25+\h)  node[anchor=north, pos=1.25] {$x_{N_\mathcal{P}}$};
        \draw (6,0.25+\h) -- (6,-0.25+\h)  node[anchor=north, pos=1.25] {$x_{N_\mathcal{P}+1}$};
    
        \draw[-, thin] (0,\h) -- (1,\h) node[midway,fill=white, font=\footnotesize] {$U_{1}$};
        \draw[-, thin] (1,\h) -- (2,\h) node[midway,fill=white, font=\footnotesize] {$U_{2}$};
        \draw[-, thin] (2,\h) -- (3,\h) node[midway,fill=white, font=\footnotesize] {$...$};
        \draw[-, thin] (3,\h) -- (4,\h) node[midway,fill=white, font=\footnotesize] {$U_{c}$};
        \draw[-, thin] (4,\h) -- (5,\h) node[midway,fill=white, font=\footnotesize] {$...$};
        \draw[-, thin] (5,\h) -- (6,\h) node[midway,fill=white, font=\footnotesize] {$U_{N_\mathcal{P}}$};
    
        \node[black, font=\footnotesize, left] at (0,0) {$0$};
        \node[red, font=\normalsize] at (3.5,4.5) {$A_{c}$};
        \node[red,   align=left, font=\footnotesize] at (7.0,4.4) {Piecewise-\\Affine (PA)};
        \node[blue,  align=left, font=\footnotesize] at (7.1,2.25) {Continuous\\Piecewise-\\Affine (CPA)};
        \node[black, align=left, font=\footnotesize] at (7.0,0.7) {CPA \& zero\\at boundary};
    
        \pgfmathsetseed{2}
        \foreach \i in {0,...,\P}{
            \pgfmathparse{3.0 + \M * random() / 2.0};
            \coordinate (A) at (\i,\pgfmathresult);
            \pgfmathparse{3.0 + \M * random() / 2.0};
            \coordinate (B) at (\i+1,\pgfmathresult);
            \filldraw[red] (A) circle (1.5pt);
            \draw[dotted, red, thick] (A) -- (B);
            \filldraw[red] (B) circle (1.5pt);
        }
        \pgfmathsetseed{2}
        \pgfmathparse{\M * random()}\let\var=\pgfmathresult;
        \foreach \i in {0,...,\P}{
            \coordinate (A) at (\i,\var);
            \pgfmathparse{1.25 + \M * random() / 2.0};
            \coordinate (B) at (\i+1,\pgfmathresult);
            \filldraw[blue] (A) circle (1.5pt);
            \draw[-, solid, blue, thin] (A) -- (B);
            \filldraw[blue] (B) circle (1.5pt);
            \global\let\var=\pgfmathresult
        }
        \pgfmathsetseed{3}
        \pgfmathsetmacro{\var}{0}
        \foreach \i in {0,...,\P}{
            \coordinate (A) at (\i,\var);
            \pgfmathparse{\M * random() / 4.0};
            \coordinate (B) at (\i+1,\pgfmathresult);
            \ifnum\i=\P
                \coordinate (B) at (\i+1,0);
            \fi
            \filldraw[black] (A) circle (1.5pt);
            \draw[-, densely dashed, thin] (A) -- (B);
            \filldraw[black] (B) circle (1.5pt);
            \global\let\var=\pgfmathresult
        }
    \end{tikzpicture}
    }
    \vskip -0.1in
    \caption{Velocity functions $v(x)$. Each cell $U_c$ in the tessellation $\mathcal{P}$ defines an affine transformation $\textbf{A}_{c}=\begin{bmatrix} a_c & b_c \end{bmatrix} \in \mathbb{R}^{1 \times 2}$.}
    \label{fig:velocity_functions}
    \end{center}
\end{figure}
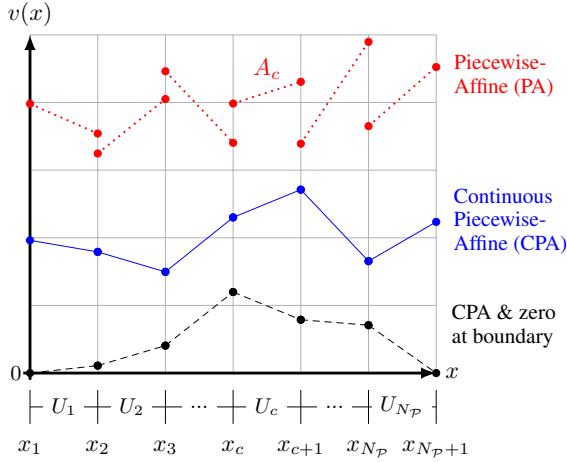

\begin{definition}\label{def:continuous_piecewise_affine}
$f$ is called CPA if it is continuous and piecewise affine.
\end{definition}

Let $\mathcal{V}_{\Omega, \mathcal{P}}$ be the spaces of CPA velocity fields on $\Omega$ w.r.t $\mathcal{P}$ and $d = \text{dim}(\mathcal{V}) = N_v$ its dimensionality.
A generic element of $\mathcal{V}_{\Omega, \mathcal{P}}$ is denoted by $v_\textbf{A}$ where 

$\textbf{A}=(\textbf{A}_1, ...\,, \textbf{A}_{N_\mathcal{P}}) = \begin{bmatrix}
a_1 \,\cdots\, a_c\, \cdots \, a_{N_\mathcal{P}} \\
b_1 \,\cdots\, b_c\, \cdots \, b_{N_\mathcal{P}} 
\end{bmatrix}^T \in \mathbb{R}^{N_\mathcal{P} \times 2}$
The vectorize operation $vec(\textbf{A})$ computes the row by row flattening to a column vector. If $\textbf{A}_c = \begin{bmatrix} a_c & b_c \end{bmatrix} \in \mathbb{R}^{1 \times 2}$ then $vec(\textbf{A}_c)=\textbf{A}_{c}^{T} = \begin{bmatrix} a_c & b_c \end{bmatrix}^T \in \mathbb{R}^{2 \times 1}$. Also, if $\textbf{A} \in \mathbb{R}^{{N_\mathcal{P}} \times 2}$ then $vec(\textbf{A}) \in \mathbb{R}^{2{N_\mathcal{P}} \times 1}$, as follows:
\begin{equation*}\label{eq:vectorize}
\begin{split}
vec(\textbf{A}) &= 
\begin{bmatrix} vec(\textbf{A}_1)^T \, \cdots \, vec(\textbf{A}_c)^T  \, \cdots & vec(\textbf{A}_{N_\mathcal{P}})^T \end{bmatrix}^T \\ &= 
\begin{bmatrix} \textbf{A}_1 \, \cdots \, \textbf{A}_c \, \cdots \, \textbf{A}_{N_\mathcal{P}} \end{bmatrix}^T \\ &=
\begin{bmatrix} a_1 \, b_1 \, \cdots \, a_c \, b_c \, \cdots \, a_{N_\mathcal{P}} \, b_{N_\mathcal{P}}
\end{bmatrix}^T
\end{split}
\end{equation*}

\paragraph{Continuity Constraints}
Since matrix $\textbf{A}$ defines a CPA transformation, $vec(\textbf{A})$ must satisfy a linear system of constraints to be continuous everywhere: $\textbf{L} * vec(\textbf{A}) = \vec{\textbf{0}}$ where $\textbf{L}$ is the constraint matrix. For a detailed explanation on this matter we refer the reader to \cref{apx:velocity_continuity_constraints}.
The null space of $\textbf{L}$ coincides with the CPA vector-field space. Let $\textbf{B} = \begin{bmatrix} \textbf{B}_1 & \textbf{B}_2 & \cdots & \textbf{B}_d \end{bmatrix} \in \mathbb{R}^{2N_\mathcal{P} \times N_v}$ be the orthonormal basis of the null space of $\textbf{L}$.
Under this setting, $\boldsymbol{\theta} = \begin{bmatrix} \theta_1 & \theta_2 & \cdots & \theta_d \end{bmatrix} \in \mathbb{R}^{d} = \mathbb{R}^{N_v}$ are the coefficients (parameters) of each basis vector, and we can compute the matrix $\textbf{A}$ as follows: $vec(\textbf{A}) = \textbf{B} \cdot \boldsymbol{\theta} = \theta_1 \cdot \textbf{B}_1 + \theta_2 \cdot \textbf{B}_2 + \cdots + \theta_d \cdot \textbf{B}_d = \sum_{j=1}^{d} \theta_j \cdot \textbf{B}_j$. If the velocity field is built using the orthonormal basis $\textbf{B}$ such that $vec(\textbf{A}) = \textbf{B} \cdot \boldsymbol{\theta}$, then $v_\textbf{A}$ is CPA.

In this work, we implement four different methods to obtain the null space of $\textbf{L}$: SVD decomposition, QR decomposition, Reduced Row Echelon Form (RREF) and Sparse Form. We refer the reader to \cref{apx:null_space} for a comparison of these four spaces about sparsity and computation times.

\paragraph{Additional Constraints}
To allocate additional constraints, we must extend the constraint matrix $\textbf{L}$ to have more rows. The null space of the extended $L$ is a linear subspace of the null space of the original $\textbf{L}$. For instance, constraining the velocity field to be zero at the border of $\Omega$ : $v(\delta \Omega) = 0$ (zero-boundary constraint) adds two additional equations (one at each limit of $\Omega$), thus the number of constraints $d'=d-2$ and basis $\textbf{B} = \begin{bmatrix} \textbf{B}_1 & \textbf{B}_2 & \cdots & \textbf{B}_d' \end{bmatrix} \in \mathbb{R}^{2N_\mathcal{P} \times d'}$

\paragraph{Smoothness Priors}
We include smoothness priors on CPA velocity functions as done by \cite{Freifeld2017}: First sampling a zero-mean gaussian with $D \times D$ covariance matrix $\boldsymbol{\Sigma}_{PA}$ whose correlations decay with inter-cell distances $vec(\textbf{A}) \sim \mathcal{N}(0_{D \times 1}, \boldsymbol{\Sigma}_{PA})$; and then projecting it into the CPA space: $\boldsymbol{\theta} = \textbf{B}^T \cdot vec(\textbf{A})$. With this procedure we can sample transformation parameters $\boldsymbol{\theta}$ from a prior distribution:
$p(\boldsymbol{\theta}) = \mathcal{N}(0_{d \times 1}, \boldsymbol{\Sigma}_{CPA})$, where $\boldsymbol{\Sigma}_{CPA} = \textbf{B}^T \cdot \boldsymbol{\Sigma}_{PA} \cdot \textbf{B}$ uses the squared exponential kernel and has two parameters: $\lambda_{\sigma}$ which controls the overall variance and $\lambda_{s}$ which controls the kernel's length-scale. Small $\lambda_{\sigma}$ generate close to the identity warps and vice versa. Large $\lambda_{s}$ favors purely affine velocity fields.

\subsection{CPA Diffeomorphic Transformations}\label{sec:cpa_diffeomorphic_transformations}

\begin{definition}\label{def:diffeomorphism}
A map $T: \Omega \rightarrow \Omega$ is called a ($\mathcal{C}^1$) diffeomorphism on $\Omega$ if $T^{-1}$ exists and both  $T$ and $T^{-1}$ are differentiable. A diffeomorphism can be obtained, via integration, from uniformly continuous stationary velocity fields $T^{\theta}(x) = \phi^{\theta}(x,1)$ 
where $\phi^{\theta}(x,t) = x + \int_0^t v^{\theta}(\phi^{\theta}(x,\tau)) d\tau$ for uniformly continuous $v: \Omega \rightarrow \mathbb{R}$ and integration time $t$.
\end{definition}

Any continuous velocity field, whether piecewise-affine or not, defines differentiable $\mathbb{R} \rightarrow \Omega$ trajectories. 
Let
\begin{equation*}\label{eq:trajectory}
\left.\begin{matrix}
    x \in \Omega \\
    v^\theta \in \mathcal{V}_{\Omega, \mathcal{P}}
\end{matrix}\right\}
\text{ define a function: } t \rightarrow \phi^\theta(x,t)
\end{equation*}
such that $\phi^\theta(x,0) = x$ and $\phi^\theta(x,t)$ solves the integral equation:
\begin{equation}\label{eq:integral}
\phi^\theta(x,t) = x + \int_0^t v^\theta(\phi^\theta(x,\tau)) d\tau
\end{equation}
or, the equivalent ordinary differential equation (ODE):
\begin{equation}\label{eq:ode}
\frac{d\phi^\theta(x,t)}{dt} = v^\theta(\phi^\theta(x,t))
\end{equation}
This integral equation should not be confused with the piecewise-quadratic $\mathbb{R} \rightarrow \Omega$ map, $y \rightarrow \int_0^y v^\theta(x) dx$. Both $x \rightarrow \phi^\theta(x,t)$ and $t \rightarrow \phi^\theta(x,t)$ are not piecewise quadratic.

\begin{figure*}[!b]
\vskip -0.1in
\begin{equation}\label{eq:derivative_complete}\tag{5}
\begin{split}
\frac{\partial \phi^\theta(x,t)}{\partial \theta_k} &= 
\bigg(
\frac{\partial \psi^\theta(x,t)}{\partial \theta_k} + 
\frac{\partial \psi^\theta(x,t)}{\partial t^\theta} \cdot
\frac{\partial t^\theta}{\partial \theta_k} + 
\frac{\partial \psi^\theta(x,t)}{\partial x} \cdot
\frac{\partial x}{\partial \theta_k}
\bigg)_{\substack{x = x_m \\ t = t_m}} = \\
 &= a_{c_m}^{(k)} \, t_m \, e^{t_m a_{c_m}} \Big(x_m + \frac{b_{c_m}}{a_{c_m}} \Big) + 
\Big(e^{t_m a_{c_m}}-1\Big)\frac{b_{c_m}^{(k)} \, a_{c_m} - a_{c_m}^{(k)} \, b_{c_m}}{a_{c_m}^2} - \\ & \quad 
e^{t_m a_{c_m}} \Big( a_{c_m} x_m + b_{c_m} \Big) 
\sum_{i=1}^{m-1} 
\Bigg(
-\frac{a_{c_i}^{(k)}}{a_{c_i}^2} \log \bigg( \frac{a_{c_i} x_{c_i} + b_{c_i}}{a_{c_i} x_i + b_{c_i}} \bigg) +
 \frac{(x_{c_i} - x_i)(b_{c_i}^{(k)} a_{c_i} - a_{c_i}^{(k)} b_{c_i})}{a_{c_i}(a_{c_i} x_i + b_{c_i})(a_{c_i} x_{c_i} + b_{c_i})}
\Bigg)
\end{split}
\end{equation}
\vskip -0.25in
\end{figure*}

Letting $x$ vary and fixing $t$, $x \rightarrow \phi^\theta(x,t)$ is an $\Omega \rightarrow \Omega$ transformation. Without loss of generality, we may set $t=1$ and define $T^{\theta}(x) = \phi^\theta(x,1)$ where $\theta \in \mathbb{R}^d$.
The solution $\phi$ to this ODE is the composition of a finite number of solutions $\psi$:
\begin{equation}\label{eq:ode_solution}
\phi^\theta(x,t) = \Big(\psi_{\theta,c_m}^{t_m} \circ \psi_{\theta,c_{m-1}}^{t_{m-1}} \circ \cdots \circ \psi_{\theta,c_2}^{t_2} \circ \psi_{\theta,c_1}^{t_1} \Big)(x)
\end{equation}
where $m$ is the number of cells visited and $\psi_{\theta,c}^{t}$ is the solution of a basic ODE $\frac{d\psi}{dt}=v^\theta(\psi)$
with an $\mathbb{R} \rightarrow \mathbb{R}$ affine velocity field: 
$v^\theta(\psi) = a^\theta \psi + b^\theta$ and an initial condition $\psi(x,0) = x$.
The integration details for this ODE are included in \cref{apx:integration_details}.
\begin{equation*}\label{eq:integral_simple}
\frac{d\psi}{dt}=v^\theta(\psi)=a^\theta \psi + b^\theta \longrightarrow 
\psi = x e^{t a^\theta} + \Big(e^{t a^\theta}-1\Big) \frac{b^\theta}{a^\theta}
\end{equation*}

\subsection{Closed-Form Integration}\label{sec:closed_form_integration}

\textbf{Step 1:} 
Given inputs $x$, $\theta$, $t$ and the membership function $\gamma$, compute the cell index $c = \gamma(x)$ and the cell boundary points $\left\{\begin{matrix}
    x_c = x_{c}^{max}, \text{ if }  v(x) \geq 0 \\
    x_c = x_{c}^{min}, \text{ if }  v(x) < 0
\end{matrix}\right.$

\textbf{Step 2:}
Calculate the \textit{hitting time} $t_{hit}$ at the boundary $x_c$:
$$
\psi_c^\theta(x,t_{hit}) = x_c 
\longrightarrow
t_{hit}^\theta = \frac{1}{a_c^\theta} \log \bigg( \frac{a_c^\theta x_c + b_c^\theta}{a_c^\theta x + b_c^\theta} \bigg)
$$

\textbf{Step 3:} 
If $t_{hit}^\theta > t$ then $\phi^\theta(x,t) = \psi_c(x,t)$, otherwise repeat from step 1, with new values for
$t = t - t_{hit}^\theta$, $x = x_c$ and
$\left\{\begin{matrix}
    c = c+1, \text{ if }  v(x) \geq 0 \\
    c = c-1, \text{ if }  v(x) < 0
\end{matrix}\right.$

This is an iterative process until convergence. The upper bound for $m$ is $\max(c_{1}, N_{\mathcal{P}}-c_{1} + 1)$, where $c_1$ refers to the first visited cell index.

\subsection{Closed-Form Derivatives of $T^{\theta}(x)$ w.r.t. $\theta$}\label{sec:closed_form_derivatives}

Let's recall that the trajectory (the solution for the integral equation \ref{eq:integral}) is a composition of a finite number of solutions $\psi$, as given by \cref{eq:ode_solution}. During the iterative process of integration, several cells are crossed, starting from cell $c_1$ at integration time $t_1=1$, and finishing at cell $c_m$ at integration time $t_m$. The integration time $t_m$ of the last cell $c_m$ can be calculated by subtracting from the initial integration time the accumulated boundary hitting times: $t_m = t_1 - \sum_{i=1}^{m-1} t_{hit}^\theta(c_i, x_i)$. The final integration point $x_m$ is the boundary of the penultimate cell $c_{m-1}$: $x_m = x_{c_{m-1}}$. In case only one cell is visited, both time and space remain unchanged: $t_m = 1$ and $x_m = x$. Taking all of this into account, the trajectory can be calculated as follows:
\begin{equation}\label{eq:closed_form_integration}\tag{4}
\begin{split}
\phi^\theta(x,t) &= \psi^\theta(x=x_m,t=t_m) \\ &= 
\bigg(
    x e^{t a_c} + \Big(e^{t a_c}-1\Big) \frac{b_c}{a_c}
\bigg)_{\substack{x = x_m \\ t = t_m}}
\end{split}
\end{equation}
\addtocounter{equation}{2}
Therefore, the derivative can be calculated by going backwards in the integration direction. We are interested in the derivative of the trajectory w.r.t. the parameters 
$\boldsymbol{\theta} = \begin{bmatrix} \theta_1 & \theta_2 & \cdots & \theta_d \end{bmatrix} \in 
\mathbb{R}^{d}$. \cref{eq:derivative_complete} shows the expression for the partial derivative w.r.t. one of the coefficients of $\boldsymbol{\theta}$, i.e., $\theta_k$. We refer the reader to \cref{apx:closed_form_derivatives,apx:closed_form_slope} for the corresponding detailed mathematical derivations.

\subsection{Scaling-and-Squaring}\label{sec:scaling_squaring}

We use the scaling-and-squaring method \cite{Moler2003,Higham2009} to approximate the numerical or closed-form integration of the velocity field. This method uses the following property of diffeomorphic transformations to accelerate the computation of the integral: $\phi(x,t+s) = \phi(x,t) \,\circ\,\phi(x,s)$. Namely, computing the transformation $\phi$ at time $t+s$ is equivalent to composing the transformations at time $t$ and $s$. The scaling-and-squaring method imposes $t=s$, so it only needs to compute one transformation and self-compose it: 
$\phi(x,2t) = \phi(x,t)\,\circ\,\phi(x,t)$. Repeating this procedure multiple times (N) we can efficiently approximate the integration:
\begin{equation}\label{eq:scaling_squaring}
\phi(x,t^{2N}) = \phi(x,t) \; \underbrace{\circ \; \cdots \; \circ}_{N} \; \phi(x,t)
\end{equation}
This procedure relates to the log-euclidean framework \cite{Arsigny2006,Arsigny2006a} that uses (inverse) scaling and squaring for exponential (logarithm) operation on velocity fields. 

\subsection{Temporal Transformer Network (TTN) Model}

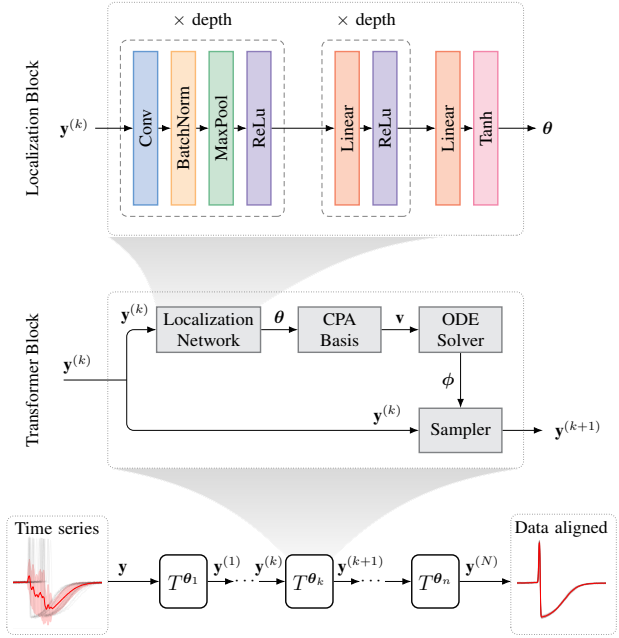
\begin{figure}[!t]
    \vskip 0.2in
    \begin{center}
    \scalebox{0.67}{
    \begin{tikzpicture}[scale=1]
        \draw[draw=white, fill, top color=black!15, bottom color=white] (5,0) to[bend right=20] (1, 2.25) -- (9.25, 2.25) to[bend right=20] cycle;
        \draw[draw=white, fill, top color=black!15, bottom color=white] (3,4.75) -- (1.95,5.5) to[bend right=20] (1, 6.85) -- (9.25, 6.85) to[bend right=18] (4.05,5.5) -- cycle;

        \node[draw=none, text centered, rotate=90] at (-0.5,4) {Transformer Block};
        \node[draw=none, text centered, rotate=90] at (-0.5, 9) {Localization Block};
    
        \begin{scope}[shift={(0,0)}]
            \tikzstyle{r}=[draw, text width=2em, minimum height=3em, text centered, thick, rounded corners]
            
            \node[draw=none,fill=none] (original) at (0,0){\includegraphics[width=5em]{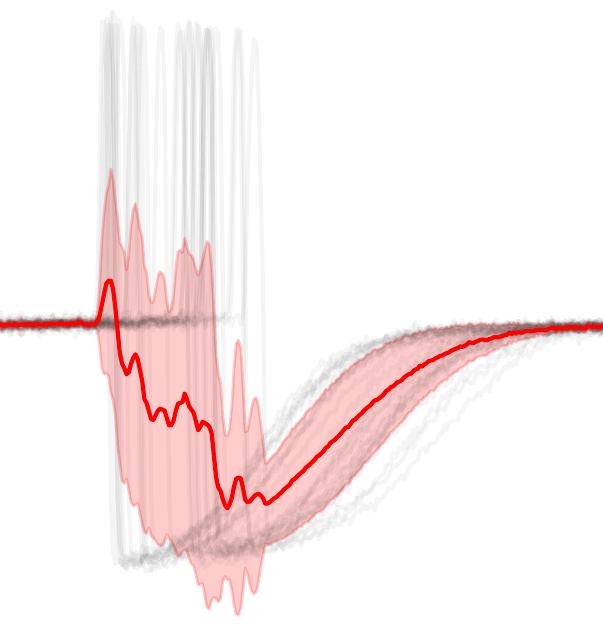}};
            \node[draw=none,fill=none] (aligned) at (10,0){\includegraphics[width=5em]{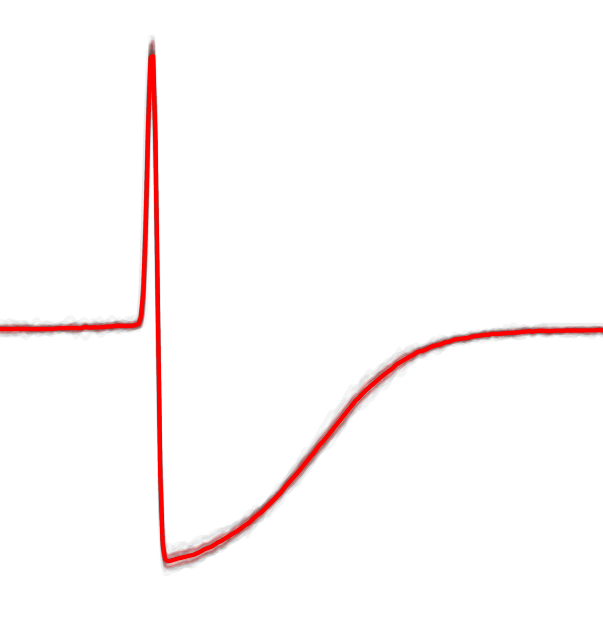}};
            
            \node[text width=5em, text centered] (A) at (original.north) {Time series};
            \node[text width=5.5em, text centered] (B) at (aligned.north) {Data aligned};
    
            \draw[draw=black!50, fill=none, style=densely dotted, rounded corners] (original.south west) rectangle (A.north east);
            \draw[draw=black!50, fill=none, style=densely dotted, rounded corners] (aligned.south west) rectangle (B.north east);
    
            \node[r, fill=white] (C) at (2.5,0) {\large $T^{\boldsymbol{\theta}_1}$};
            \node[r, fill=white] (D) at (5,0) {\large $T^{\boldsymbol{\theta}_k}$};
            \node[r, fill=white] (E) at (7.5,0) {\large $T^{\boldsymbol{\theta}_n}$};
            
            \draw[-Latex, color=black] (original.east) -- node[anchor=south east] {$\textbf{y}$} (C.west);
            \draw[-Latex, color=black] (C.east) node[anchor=south west] {$\textbf{y}^{(1)}$}  -- (3.5,0);
            \node[] at (3.75,0) {$\cdots$};
            \draw[-Latex, color=black] (4,0) node[anchor=south west, xshift=-0.2cm] {$\textbf{y}^{(k)}$} -- (D.west);
    
            \draw[-Latex, color=black] (D.east) -- node[anchor=south west, xshift=-0.3cm] {$\textbf{y}^{(k+1)}$} (6,0);
            \draw[-Latex, color=black] (6.5, 0) -- (E.west);
            \node[] at (6.25,0) {$\cdots$};
            \draw[-Latex, color=black] (E.east) node[anchor=south west] {$\textbf{y}^{(N)}$} -- (aligned.west);
        \end{scope}
    
        \begin{scope}[shift={(3,5)}]
            \tikzstyle{r}=[text width=4em, minimum height=2.5em, text centered, thick]
            \node[r, text width=5.2em, draw=Black!60, fill=Black!10, text=black] (A) at (0,0) {Localization Network};
            \node[r, draw=Black!60, fill=Black!10, text=black] (B) at (2.6,0) {CPA Basis};
            \node[r, draw=Black!60, fill=Black!10, text=black] (C) at (5,0) {ODE Solver};
            \node[r, draw=Black!60, fill=Black!10, text=black] (D) at (5,-2) {Sampler};
            \node[] (E) at (-1.5, -1) {};
            \node[] (F) at (-3, -1) {};
            \node[] (G) at (6.75, -2) {};
    
            \draw[] (F) node[anchor=south west] {$\textbf{y}^{(k)}$} -- (E);
            \draw[-Latex, color=black, rounded corners] (E.west) |- (A.west) node[anchor=south east] {$\textbf{y}^{(k)}$};
            \draw[-Latex, color=black] (A.east) -- (B.west) node[midway, anchor=south] {$\boldsymbol{\theta}$};
            \draw[-Latex, color=black] (B.east) -- (C.west) node[midway, anchor=south] {$\textbf{v}$};
            \draw[-Latex, color=black] (C.south) -- (D.north) node[midway, anchor=east] {$\boldsymbol{\phi}$};
            \draw[-Latex, color=black, rounded corners] (E.west) |- (D.west) node[anchor=south east, xshift=-0.2cm] {$\textbf{y}^{(k)}$};
            \draw[-Latex, color=black] (D.east) -- (G) node[anchor=west] {$\textbf{y}^{(k+1)}$};
            \draw[draw=black!50, fill=none, style=densely dotted, rounded corners] (-2, -2.75) rectangle (6.25,0.75);
        \end{scope}

        \begin{scope}[scale=0.5, shift={(17,18)}, rotate=90, nodes={rotate=90}]
            \begin{scope}[shift={(0,9)}, nodes={draw, text width=8em, align=center, thick}]
                \node[draw=RoyalBlue!60, fill=RoyalBlue!20, text=black] (D3) at (0,4.5) {Conv};
                \node[draw=BurntOrange!60, fill=BurntOrange!20, text=black] (C3) at (0,3) {BatchNorm};
                \node[draw=ForestGreen!60, fill=ForestGreen!20, text=black] (B3) at (0,1.5) {MaxPool};
                \node[draw=Violet!60, fill=Violet!20, text=black] (A3) at (0,0) {ReLu};
                \draw[-Latex, color=black] (D3) -- (C3);
                \draw[-Latex, color=black] (C3) -- (B3);
                \draw[-Latex, color=black] (B3) -- (A3);
                \draw[draw=black!50, fill=none, style=densely dashed, rounded corners] (-3.5,-1) rectangle (3.5,5.5);
                \node[draw=none, rotate=-90] at (4.25,2.25) {$\times$ depth};
            \end{scope}
    
            \begin{scope}[shift={(0,4)}, nodes={draw, text width=8em, align=center, thick}]
                \node[draw=Red!60, fill=Red!20, text=black] (B2) at (0,1.5) {Linear};
                \node[draw=Violet!60, fill=Violet!20, text=black] (A2) at (0,0) {ReLu};
                \draw[-Latex, color=black] (B2) -- (A2);
                \draw[draw=black!50, fill=none, style=densely dashed, rounded corners] (-3.5,-1) rectangle (3.5,2.5);
                \node[draw=none, rotate=-90] at (4.25,0.75) {$\times$ depth};
            \end{scope}
    
            \begin{scope}[shift={(0,0)}, nodes={draw, text width=8em, align=center, thick}]
                \node[draw=Red!60, fill=Red!20, text=black] (B1) at (0,1.5) {Linear};
                \node[draw=RubineRed!60, fill=RubineRed!20, text=black] (A1) at (0,0) {Tanh};
                \draw[-Latex, color=black] (B1) -- (A1);
            \end{scope}
    
            \draw[-Latex, color=black] (A3) -- (B2);
            \draw[-Latex, color=black] (A2) -- (B1);
            
            \draw[draw=black!50, fill=none, style=densely dotted, rounded corners] (-4.25, -1.5) rectangle (5,15);
            
            \draw[-Latex, color=black]  (0,15.5) node[anchor=east, rotate=-90] {$\textbf{y}^{(k)}$}  -- (D3);
            \draw[-Latex, color=black] (A1) -- (0,-2) node[anchor=west, rotate=-90] {$\boldsymbol{\theta}$};
    
        \end{scope}
    
    \end{tikzpicture}
    }
    \caption{Proposed temporal transformer architecture. \textbf{Bottom}: time series $\textbf{y}$ is aligned by applying a sequence of transformers $T^{\boldsymbol{\theta}}$ that minimize the empirical variance of the warped signals. \textbf{Middle}: each transformer block resembles the STN proposed by \cite{Jaderberg2015}. 
    Transformed data is sampled based on a diffeomorphic flow $\boldsymbol{\phi}$ obtained from the integration of a velocity function $\textbf{v}$ from a first order ordinary differential equation (ODE). \textbf{Top}: The parameters $\boldsymbol{\theta}$ of the velocity function $\textbf{v}$ are computed by the localization network based on each signal $\textbf{y}^{(k)}$.
    }
    \label{fig:architecture}
    \end{center}
    \vskip -0.2in
\end{figure}

The proposed model resembles the Spatial Transformer Network proposed by \cite{Jaderberg2015}, later adapted for time series alignment by \cite{oh2018learning,Lohit2019,Weber2019} (see \cref{fig:architecture}), and can recurrently apply nonlinear time warps to the input signal. Composing warps increases the expressiveness without refining $\Omega$ as it implies non-stationary velocity functions which are CPA in $\Omega$ and piecewise constant in time. Note that unlike \cite{Weber2019}, in our model the localization network parameters are not shared across layers. 
The differentiable sampler uses piecewise linear interpolation to estimate the warped signals. We refer the reader to \cref{apx:linear_interpolation_grid} for details about the sampler estimation and its derivatives. 

\paragraph{Loss function}
Let $\textbf{y}_i$ denote an input signal among $N$ time series samples, and $\boldsymbol{\theta}_i = F_{loc}(\textbf{w}, \textbf{y}_i)$ denote the corresponding output of the localization network $F_{loc}(\textbf{w},\cdot)$, and let 
$\textbf{z}_i = \textbf{y}_i \circ \boldsymbol{\phi}_{\boldsymbol{\theta}_i}$ 
denote the result of warping $\textbf{y}_i$ by $ \boldsymbol{\phi}_{\boldsymbol{\theta}_i}$, where $\boldsymbol{\theta}_i$ depends on $\textbf{w}$ and $\textbf{y}_i$. The variance of the observed $(\textbf{y}_i)_{i=1}^{N}$ is partially explained by the latent warps $(\textbf{y}_i)_{i=1}^{N}$, so we seek to minimize the empirical variance of the warped signals:

\begin{equation}\label{eq:loss_data_single}
\mathcal{L}_{data}^{1} (\textbf{y}_i|_{i=1}^{N}) =
\frac{1}{N} \sum_{i=1}^{N} 
\Big\Vert 
\textbf{y}_i \circ \boldsymbol{\phi}_i - \frac{1}{N} \sum_{j=1}^{N} \textbf{y}_j \circ \boldsymbol{\phi}_j
\Big\Vert_{2}^2
\end{equation}
where $\Vert\cdot \Vert_{2}$ is the $l_2$ norm. For multi-class problems, the expression is the sum of within-class variances:
\begin{equation}\label{eq:loss_data_multi}
\mathcal{L}_{data}^{K} (\textbf{y}_i|_{i=1}^{N})= 
\sum_{k=1}^{K}
\frac{1}{N_k} 
\mathcal{L}_{data}^{1} (\textbf{y}_{r_{i}=k})
\end{equation}
where $K$ is the number of classes, $r_i$ takes values in $\{1,\cdots,K\}$ and is the class label associated with $\textbf{y}_i$ and $N_k$ is the number of examples in class $k$.
Hence, we formulate the joint alignment problem as to simultaneously compute the centroid and align all sequential data within a class, under a semi-supervised schema: A single net learns how to perform joint alignment within each class without knowing the class labels at test time. Thus, while the within-class alignment remains unsupervised (as in the single-class case), for multi-class problems labels are used during training (not on testing) to reduce the variance within each class separately.
Both single- and multi-class cases use a regularization term for the warps:
\begin{equation}\label{eq:loss_reg}
\mathcal{L}_{reg}(\textbf{y}_i|_{i=1}^{N}) = 
\frac{1}{N} \sum_{i=1}^{N}
\boldsymbol{\theta}_i^T \boldsymbol{\Sigma}_{CPA}^{-1} \boldsymbol{\theta}_i
\end{equation}
In the context of joint-alignment using regularization is critical, partly since it is too easy to minimize $\mathcal{L}_{data}$ by unrealistically-large deformations that would cause most of the inter-signal variability to concentrate on a small region of the domain \cite{Weber2019}; the regularization term prevents that. 
Therefore, our loss function, to be minimized over $\textbf{w}$, is $\mathcal{L}=\mathcal{L}_{data} + \mathcal{L}_{reg}$.

\paragraph{Variable length \& multi-channel data} 
As in \cite{Weber2019}, generalization to multichannel signal is trivial. Variable-length signals can be managed with the presented loss function as long as a fixed number of points is set on the linear interpolation sampler. Yet, a neural network that can handle variable-length (a Recurrent Neural Network, for instance) is required for the localization network $F_{loc}$.

\paragraph{Implementation} 
CPA diffeomorphic transformations were implemented on multiple backends for CPU (\textit{NumPy} and \textit{PyTorch} with \textit{C++}) and GPU (\textit{PyTorch} with \textit{CUDA}) and presented under an open-source library called Diffeomorphic Fast Warping \textit{DIFW}\footnote{\url{https://github.com/imartinezl/difw}} (see Supplementary Material). 
The TTN model was implemented in \textit{PyTorch} and can be integrated with a few lines of code with other deep learning architectures for time series classification or prediction.

\section{Experiments and Results}\label{sec:results}

\subsection{Computation Time}\label{sec:computation_time} 

\begin{figure}[!htb]
    \begin{center}
    \centerline{\includegraphics[width=\linewidth]{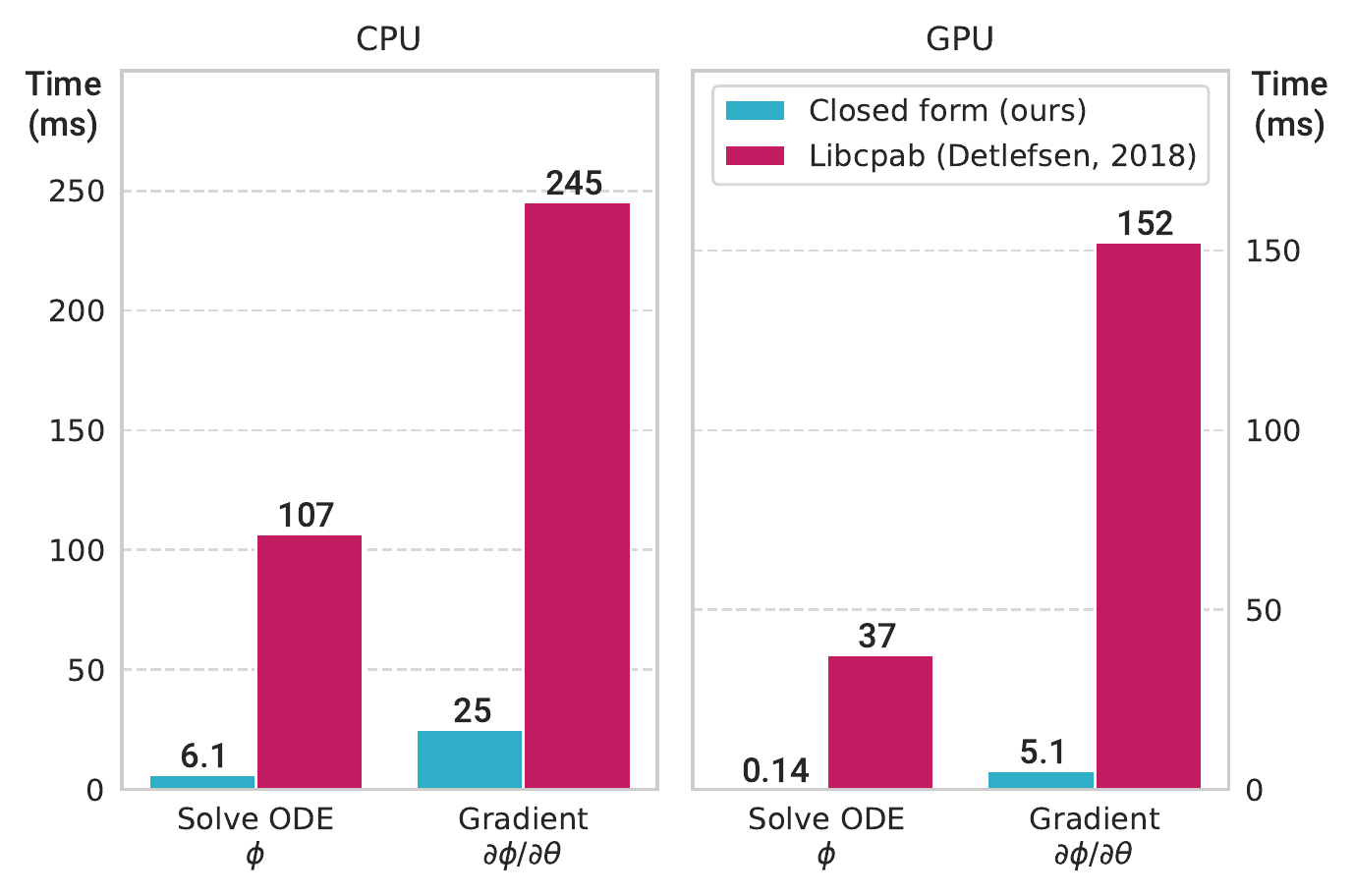}}
    \caption{Computation time ($ms$) for the forward (velocity field integration) and backward (gradient w.r.t. parameters) operations on CPU (left) and GPU (right). Our \textbf{closed form} method is compared to \textbf{libcpab} numeric solution implementation \cite{detlefsen2018LIBCPAB} which is based on \cite{Freifeld2017}. 
    }
    \label{fig:benchmark_times_compiled_linear}
    \end{center}
    \vskip -0.2in
\end{figure}

This section compares the performance of forward and backward operations between the proposed closed-form method and \textit{libcpab} \cite{detlefsen2018LIBCPAB}, which supports CPAB-based diffeomorphic transformations and implements a custom-made ODE solver \cite{Freifeld2015} that alternates between the analytic solution and a generic solver.
Speed tests show an \textbf{x18} and \textbf{x10} improvement on CPU over \textit{libcpab} for forward and backward operations respectively. On GPU the performance gain of \textit{DIFW} reaches \textbf{x260} and \textbf{x30} (see \cref{fig:benchmark_times_compiled_linear}). (Parameters: $10^{3}$ points in $\Omega$, $N_\mathcal{P}=30$, and batch size $40$).

Performance comparisons with other methods such as SoftDTW, and ResNet-TW were avoided due to parametrization and complexity differences. SoftDTW and its gradient have quadratic time \& space complexity, but produce non-diffeomorphic warping functions. ResNet-TW, on the other hand, requires multiple convolutional layers to incrementally compute the warping function and is limited by the recursive computation of the monotonic temporal constraints.

\subsection{Comparison with Numerical Methods}\label{sec:results:comparison_numerical}

We compared the integration (and gradient) error averaged over random CPA velocity fields; i.e. we drew 100 CPA velocity functions from the zero-mean gaussian prior, $\{v^{\theta_i}\}_{i=1}^{100}$, sampled uniformly 1000 points in $\Omega$, $\{x^{\theta_i}\}_{i=1}^{1000}$, and then computed the difference as presented in \cref{eq:error_comparison}. Results show a 5 decimals precision for the integration error but only 3 decimals for the gradient computation using numeric methods by \cite{detlefsen2018LIBCPAB}.

\subsection{Scaling-and-Squaring}\label{sec:results:scaling_squaring}

The scaling-and-squaring method can be applied to approximate the flow and speed up the forward and backward operations. 
This method needs to balance two factors to be competitive: the faster integration computation (scaling step) versus the extra time to self-compose the trajectory multiple times (squaring step). 
The scaled-down integral solution is computed using the closed-form expression (\cref{sec:closed_form_integration}), while adjoint equations are directly applied to compute derivatives. The reduced number of time steps benefits the memory footprint.
We show that this method can indeed boost speed performance for large deformations (see \cref{apx:scaling_squaring} for details on this matter).

\subsection{UCR Time Series Classification Archive}\label{sec:results:ucr}

The UCR \cite{dau2019ucr} time series classification archive contains 85 real-world datasets and we use a subset containing 84 datasets, as in DTAN \cite{Weber2019} and ResNet-TW \cite{Huang2021}. Details about these datasets can be found on \cref{apx:ucr_archive}. 
Experiments were conducted with the provided train and test split. 
Here we report a summary of our results which are fully detailed in \cref{apx:additional_results,apx:ncc_results}.

\paragraph{Nearest Centroid Classification (NCC) experiment} NCC first computes the centroid (average) of each class in the training set by minimizing the loss function for each class. During prediction, NCC assigns the class of the nearest centroid. 
In the lack of ground truth for the latent warps in real data, NCC accuracy rates also provide an indicative metric for the quality of the joint alignment and the average signal. Thus, we perform NCC on the UCR archive, comparing our model to: (1) the sample mean of the misaligned sets (Euclidean); (2) DBA \cite{Petitjean2011-DBA}; (3) SoftDTW \cite{cuturi2017soft}, (4) DTAN \cite{Weber2019} and (5) ResNet-TW \cite{Huang2021}.

\paragraph{Experiment hyperparameters}
For each of the UCR datasets, we train our TTN for joint alignment as in \cite{Weber2019}, where 
$N_{\mathcal{P}} \in \{16,32,64\}$,
$\lambda_{\sigma} \in \{10^{-3},10^{-2}\}$,
$\lambda_{s} \in \{0.1,0.5\}$,
the number of transformer layers $\in \{1,5\}$,
scaling-and-squaring iterations $\in \{0,8\}$
and the option to apply the zero-boundary constraint.
Summarized results of hyperparameters grid-search have been included on \cref{apx:hyperparameter}, and the full table of results is available in the Supplementary Material. 
Regarding the tessellation size $N_{\mathcal{P}}$, an ablation study was conducted to investigate how partition fineness in CPA velocity functions controls the trade-off between expressiveness and computational complexity (see \cref{apx:ablation}).
The network was initialized by Xavier initialization using a normal distribution and was trained for 500 epochs with $10^{-5}$ learning rate, a batch size of $32$ and Adam \cite{kingma2014adam} optimizer with $\beta_{1}=0.9$, $\beta_{2}=0.98$ and $\epsilon=10^{-8}$.

\begin{figure*}[t]
\vskip 0.2in
\begin{center}
    \subfigure[SyntheticControl test set, class 1]{
    \includegraphics[width=0.4\linewidth]{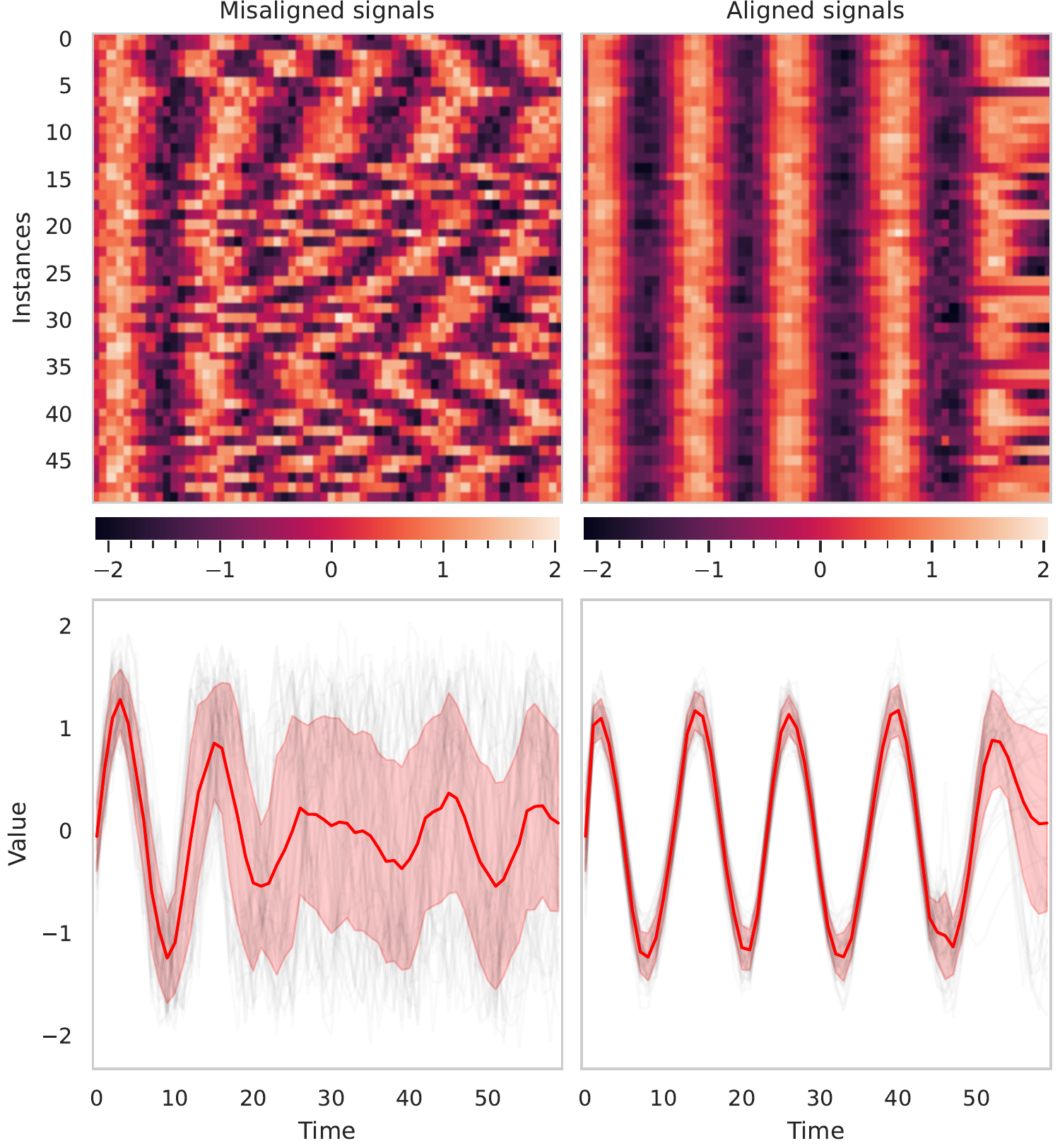}
    }
    \hspace{1.5cm}
    \subfigure[ToeSegmentation1 test set, class 0]{
    \includegraphics[width=0.4\linewidth]{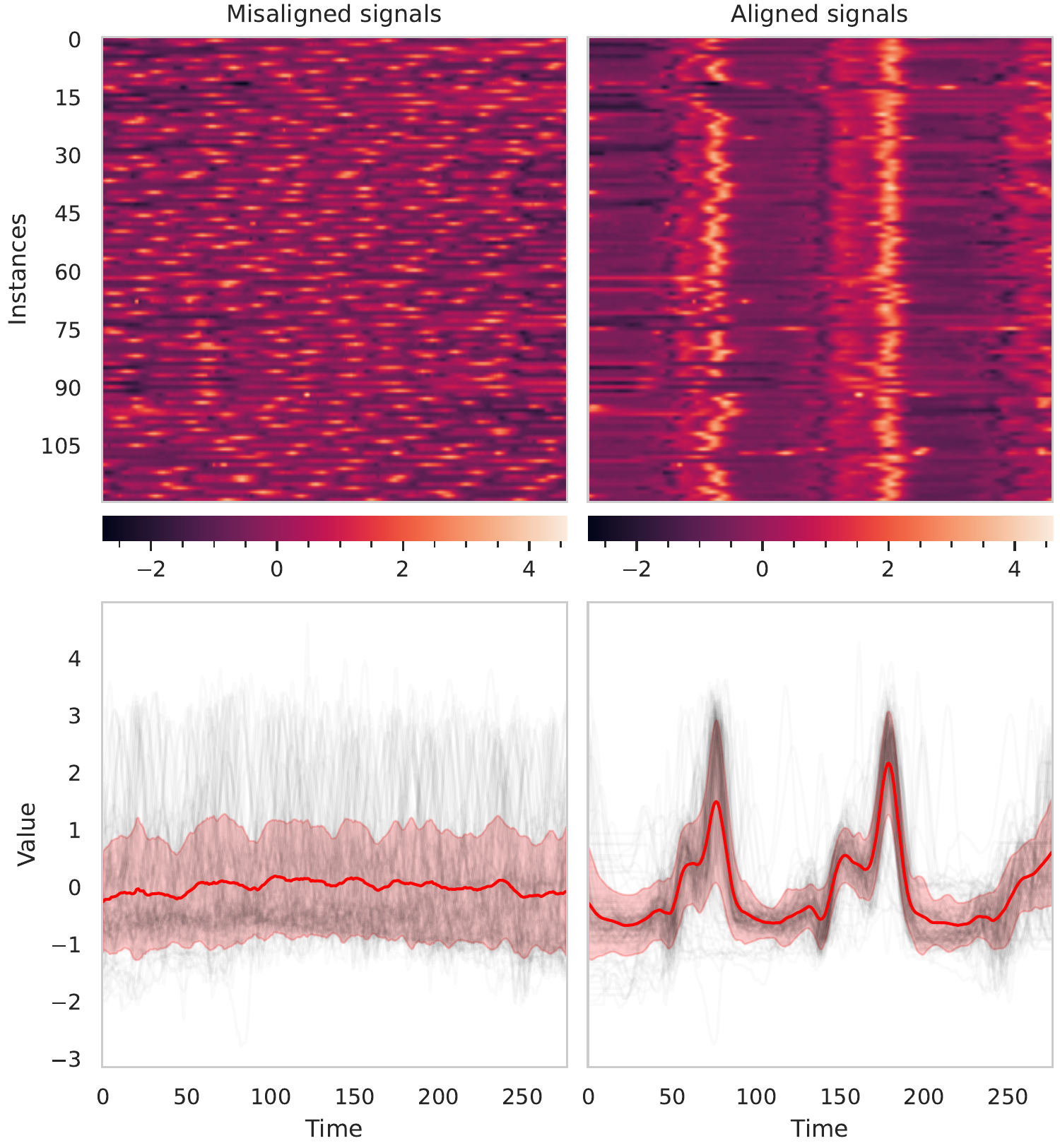}
    }
    \caption{Multi-class time series alignment on multiple test sets. \textbf{Top}: heatmap of each time series sample (row). \textbf{Bottom}: overlapping time series, red line represents Euclidean average. \textbf{Left}: original signals. \textbf{Right}: after alignment. More examples at \cref{apx:additional_results}}
    \label{fig:alignment_example}
\end{center}
\vskip -0.2in
\end{figure*}

\textbf{Results} show that temporal misalignment strongly affects the Euclidean mean, and DBA usually reaches a local minimum. SoftDTW, DTAN and ResNet-TW show similar quantitative results. NCC test accuracy found that our method was better or no worse in 94\% of the datasets compared with Euclidean,  DBA (77\%), SoftDTW (69\%), DTAN (76\%) and ResNet-TW (70\%).
Overall, our proposed TTN using \textit{DIFW} beats all 5 comparing methods. The precise computation of the gradient of the transformation translates to an efficient search in the parameter space, which leads to faster and better solutions at convergence. 

\begin{figure}[!htb]
    \vskip -0.1in
    \begin{center}
    \centerline{\includegraphics[width=\linewidth]{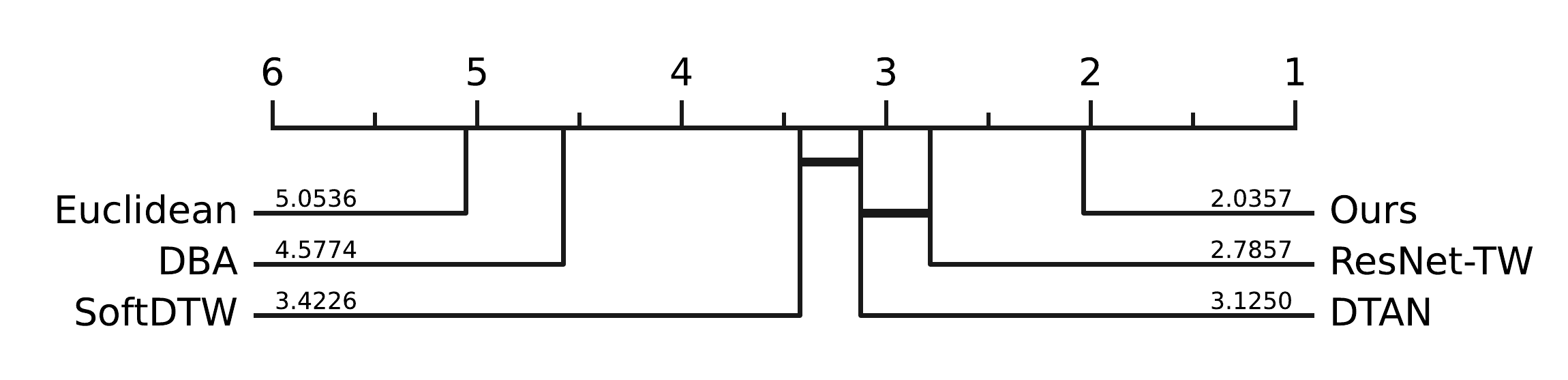}}
    \caption{Critical difference diagram for the proposed method and five competing methods (Euclidean, DBA, SoftDTW, DTAN, ResNet-TW). Solid bars indicate cliques, within which there is no significant difference in rank. Tests are performed with the Wilcoxon sign rank test using the Holm correction. }
    \label{fig:ucr_dataset_cd}
    \end{center}
    \vskip -0.3in
\end{figure}

\section{Conclusions}
\label{sec:conclusions}

In this article a closed-form expression is presented for the gradient of CPA-based one-dimensional diffeomorphic transformations, providing efficiency, speed and precision.  
The proposed method was incorporated into a temporal transformer network that can both align pairwise sequential data and learn representative average sequences for multi-class joint alignment. Experiments performed on 84 datasets from the UCR archive show that our model achieves competitive performance in joint alignment and classification.

\bibliography{main}
\bibliographystyle{icml2022}

\clearpage
\appendix
\onecolumn


\clearpage
\section{Velocity Continuity Constraints}\label{apx:velocity_continuity_constraints}

Let's consider three adjacent cells $U_i, U_j, U_k$ with affine transformations
$A_i = \begin{bmatrix} a_i & b_i \end{bmatrix}$,
$A_j = \begin{bmatrix} a_j & b_j \end{bmatrix}$ and
$A_k = \begin{bmatrix} a_k & b_k \end{bmatrix}$. The velocity field $v(x)$ must be continuous. The velocity field is denoted as $v_A(x)$, since it depends on the affine tranformation $A$. $v_A$ is continuous on every cell, because it is a linear function, but it is discontinuous on cell boundaries.
Continuity of $v_A$ at $x_j$ implies one linear constraint on $A_i$ and $A_j$. In the same way, continuity of $v_A$ at $x_k$ implies another linear constraint on $A_j$ and $A_k$.

\begin{figure}[!ht]
    \vskip -0.25in
    \begin{center}
    \scalebox{0.85}{
\begin{tikzpicture}[scale=1]
    \pgfmathsetmacro{\N}{5};
    \pgfmathsetmacro{\M}{4};
    \pgfmathsetmacro{\P}{\N-1};
    \pgfmathsetmacro{\h}{-0.5};
    \draw[step=1, black!30, thin] (0,0) grid (\N, \M);
    \draw[-latex, very thick] (-0.1,0) -- (\N,0) node[right] {$x$};
    \draw[-latex, very thick] (0,-0.1) -- (0,\M) node[above] {$v(x)$};

    \draw (0,0.25+\h) -- (0,-0.25+\h)  ;
    \draw (1,0.25+\h) -- (1,-0.25+\h)  node[anchor=north,  pos=1.25] {$x_{i}$};
    \draw (2,0.25+\h) -- (2,-0.25+\h)  node[anchor=north,  pos=1.25] {$x_{j}$};
    \draw (3,0.25+\h) -- (3,-0.25+\h)  node[anchor=north,  pos=1.25] {$x_{k}$};
    \draw (4,0.25+\h) -- (4,-0.25+\h)  node[anchor=north,  pos=1.25] {$x_{k+1}$};
    \draw (5,0.25+\h) -- (5,-0.25+\h);

    \draw[-, thin] (0,\h) -- (1,\h) node[midway,fill=white] {$...$};
    \draw[-, thin] (1,\h) -- (2,\h) node[midway,fill=white] {$U_{i}$};
    \draw[-, thin] (2,\h) -- (3,\h) node[midway,fill=white] {$U_{j}$};
    \draw[-, thin] (3,\h) -- (4,\h) node[midway,fill=white] {$U_{k}$};
    \draw[-, thin] (4,\h) -- (5,\h) node[midway,fill=white] {$...$};

    \node[thin, red] at (1.5,1.25) {$A_{i}$};
    \node[thin, red] at (2.5,1.25) {$A_{j}$};
    \node[thin, red] at (3.5,1.25) {$A_{k}$};

    \pgfmathsetseed{2}
    \foreach \i in {1,...,3}{
        \pgfmathparse{3 * random()};
        \coordinate (A) at (\i,1+\pgfmathresult);
        \pgfmathparse{3 * random()};
        \coordinate (B) at (\i+1,1+\pgfmathresult);
        \filldraw[red] (A) circle (2pt);
        \draw[red, thin] (A) -- (B);
        \filldraw[red] (B) circle (2pt);
    }
\end{tikzpicture}
    }
    \caption{Piecewise-affine velocity function. Here three adjacent cells $U_i, U_j, U_k$ are represented. Continuity conditions at the boundary are necessary to comprise a continuous piecewise-affine velocity function}
    \label{fig:velocity_continuity_constraints}
    \end{center}
    \vskip -0.1in
\end{figure}
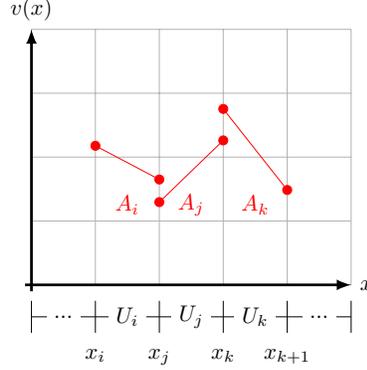

$\forall x \in \{U_i, U_j, U_k\} : v_A \text{ is continuous }$. In order to be continuous on points $x_j$ and $x_k$, two linear constraints must be satisfied:

\begin{equation}
\left\{\begin{matrix}
    A_i \cdot \tilde{x_j} = A_j \cdot \tilde{x_j} \\
    A_j \cdot \tilde{x_k} = A_k \cdot \tilde{x_k}
\end{matrix}\right.
\Rightarrow 
\left\{\begin{matrix}
    a_i \cdot x_j + b_i = a_j \cdot x_j + b_j \\
    a_j \cdot x_k + b_j = a_k \cdot x_k + b_k
\end{matrix}\right.
\Rightarrow 
\left\{\begin{matrix}
    a_i \cdot x_j + b_i - a_j \cdot x_j - b_j = 0\\
    a_j \cdot x_k + b_j - a_k \cdot x_k - b_k = 0
\end{matrix}\right.
\end{equation}

To place the linear constraints in matrix form, let's recall the vectorize operation $vec$ for this case:
\begin{equation}
\begin{split}
vec(A) &= 
\begin{bmatrix} vec(A_i)^T & vec(A_j)^T & vec(A_k)^T  \end{bmatrix}^T \\ &= 
\begin{bmatrix} A_i & A_j & A_k  \end{bmatrix}^T \\ &= 
\begin{bmatrix} a_i & b_i & a_j & b_j & a_k & b_k \end{bmatrix}^T \\ &= 
\begin{bmatrix} a_i \\ b_i \\ a_j \\ b_j \\ a_k \\ b_k \end{bmatrix}
\end{split}
\end{equation}

Therefore,
\begin{equation}
\begin{bmatrix} 
    x_j & 1 & -x_j & -1 & 0 & 0 \\
    0 & 0 & x_k & 1 & -x_k & -1
\end{bmatrix} 
\begin{bmatrix} a_i \\ b_i \\ a_j \\ b_j \\ a_k \\ b_k \end{bmatrix} =
\begin{bmatrix} 0 \\ 0 \end{bmatrix}
\Longrightarrow 
\boxed{L \cdot vec(A) = \vec{0}}
\end{equation}

Extending the continuity constraints to a tessellation with $N_\mathcal{P}$ cells, the constraint matrix $L$ has dimensions $N_{e} \times 2N_\mathcal{P}$, $vec(A)$ is $2N_\mathcal{P} \times 1$ and the null vector $\vec{0}$ is $N_{e} \times 1$. The number of shared vertices is $N_{e} = N_{v}-2 = N_\mathcal{P}-1$.

Any matrix $A$ that satisfies $L \cdot vec(A) = \vec{0}$ will be continuous everywhere. The null space of $L$ coincides with the CPA vector-field space.

\clearpage

\section{Computing Infrastructures}
We used the following computing infrastructure in our experiments: 
Intel(R) Core(TM) i7-6560U CPU @2.20GHz, 4 cores, 16gb RAM with an Nvidia Tesla P100 graphic card.

\section{Null Space of the Constraint Matrix $L$}\label{apx:null_space}

Four different methods have been implemented to obtain the null space of $L$: 
\begin{enumerate}
\vspace{-6pt}
    \setlength{\itemsep}{1pt}
    \setlength{\parskip}{0pt}
    \setlength{\parsep}{0pt}
    \item SVD decomposition
    \item QR decomposition
    \item Reduced Row Echelon Form (RREF)
    \item Sparse form (SPARSE)
\vspace{-6pt}
\end{enumerate}
Given that the RREF and the Sparse basis can be computed efficiently in closed-form, the computation time of these null spaces is several orders of magnitude faster than SVD and QR
Nonetheless, note that the null space only needs to be computed once, and as a result, this operation is not a critical step in the whole process to obtain the diffeomorphic trajectory.

\newcommand\x{0.45}
\begin{figure}[!htb]
    \vskip 0.2in
    \begin{center}
    \subfigure[SVD velocity basis]{
    \includegraphics[width=\x\linewidth]{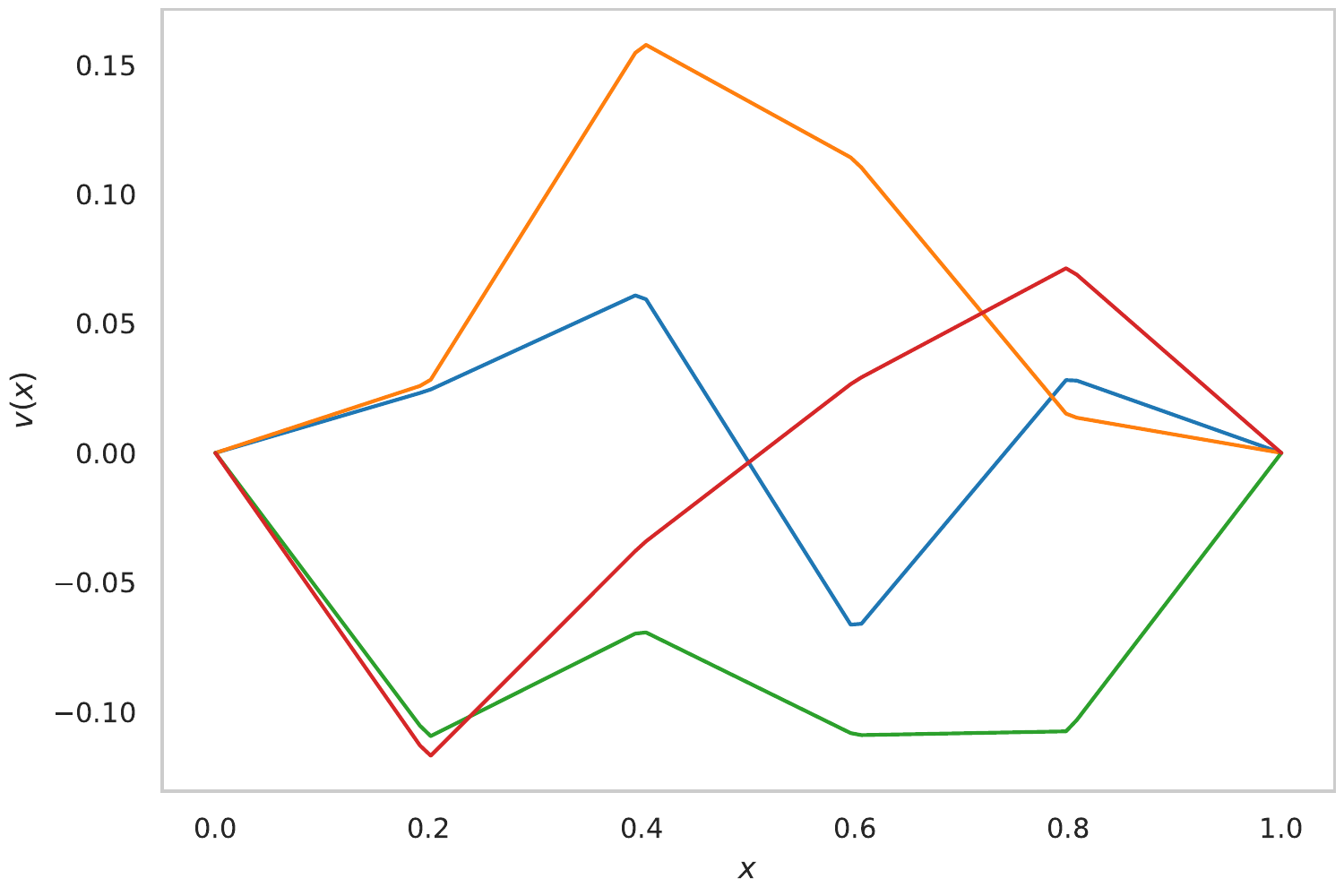}
    \label{fig:basis_svd}}
    \subfigure[QR velocity basis]{
    \includegraphics[width=\x\linewidth]{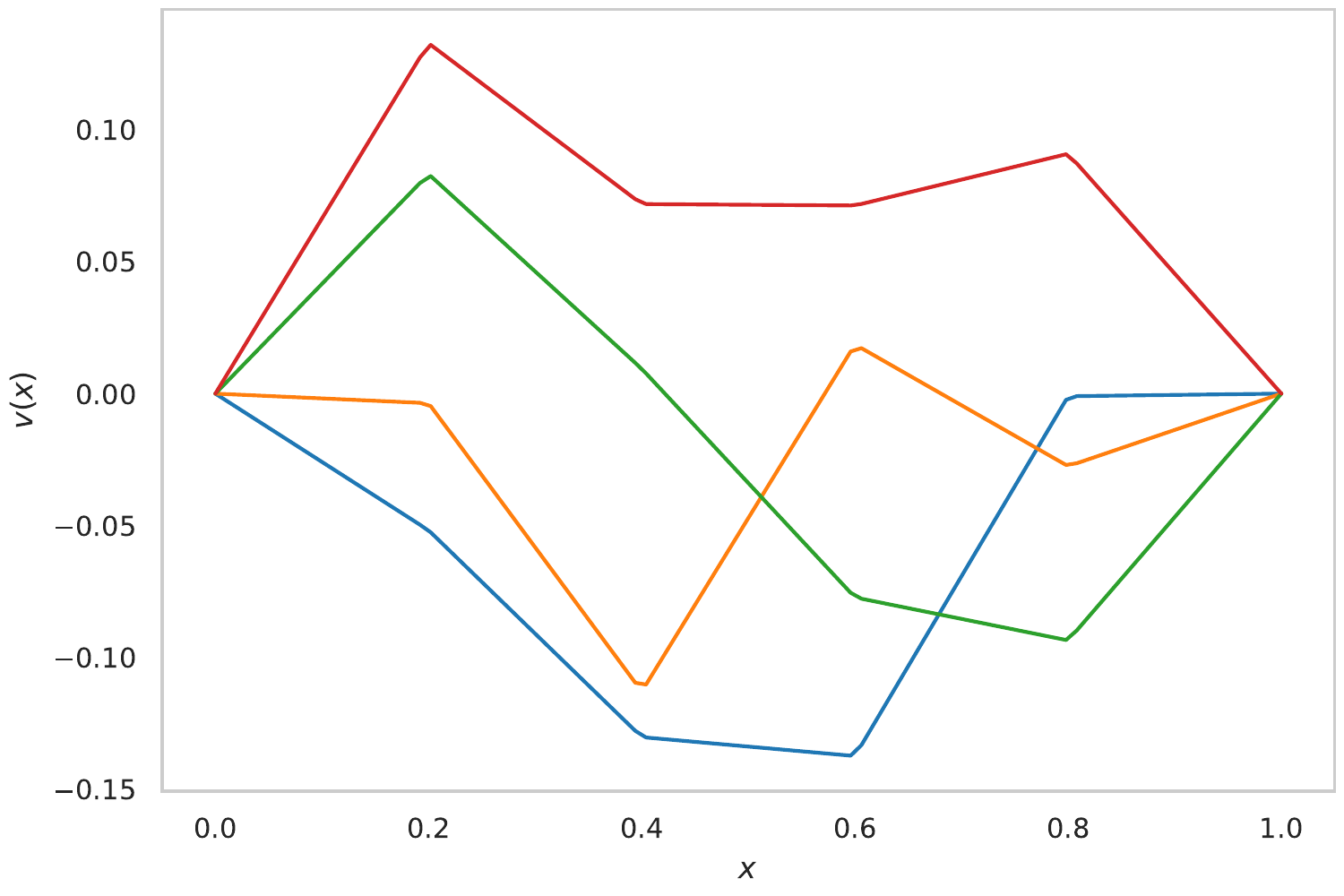}
    \label{fig:basis_qr}}
    \\
    \subfigure[RREF velocity basis]{
    \includegraphics[width=\x\linewidth]{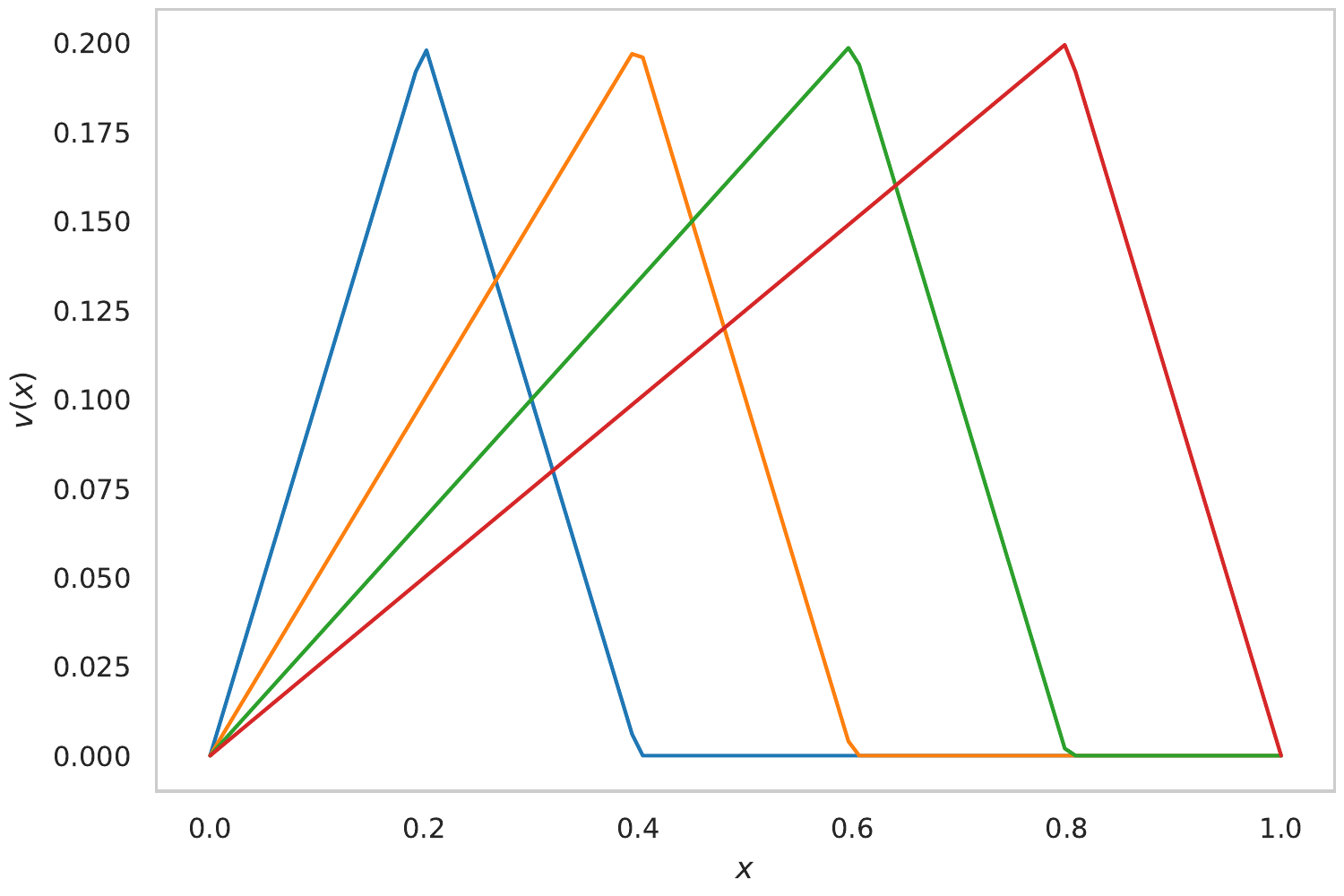}
    \label{fig:basis_rref}}
    \subfigure[SPARSE velocity basis]{
    \includegraphics[width=\x\linewidth]{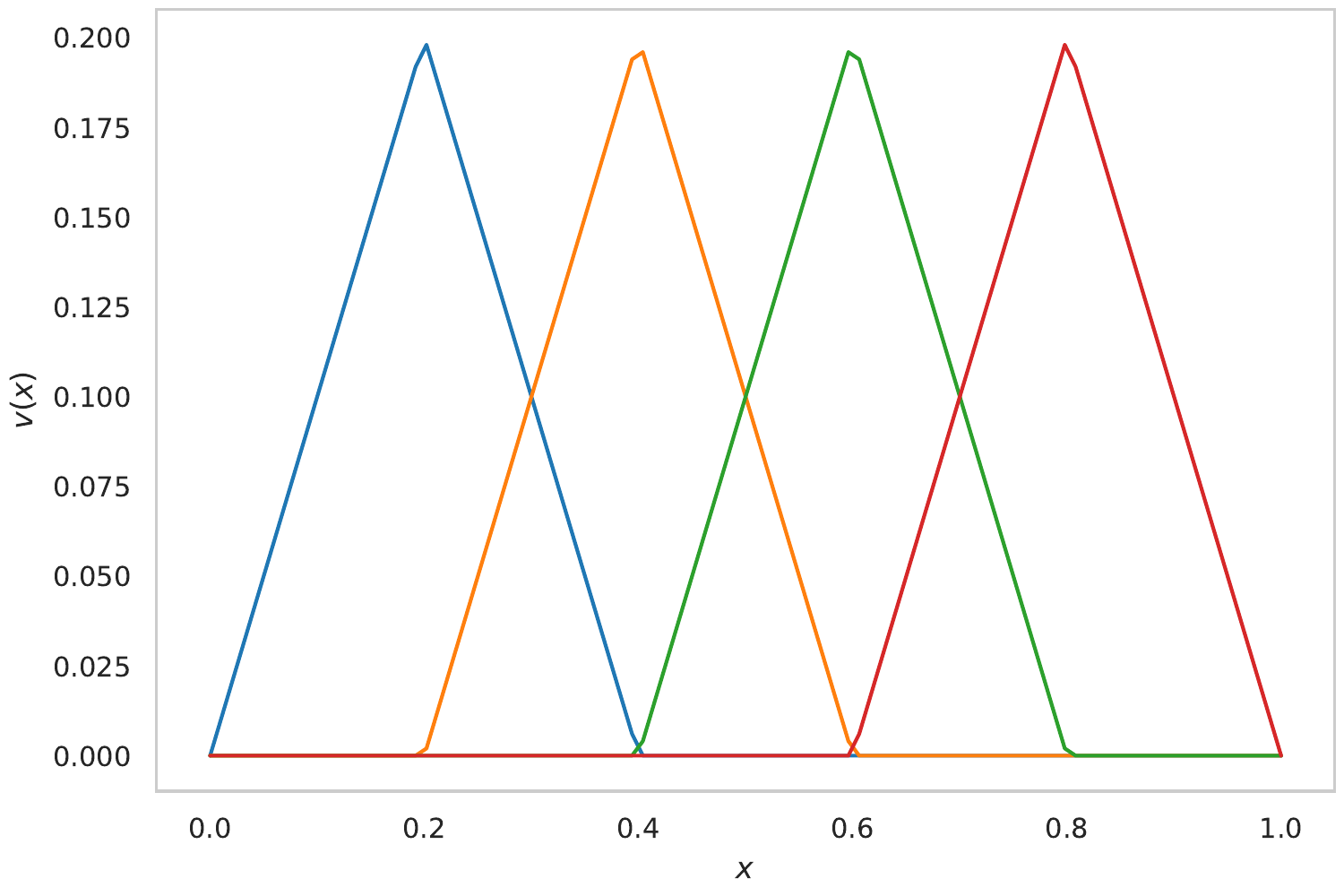}
    \label{fig:basis_sparse}}
    \caption{Comparison of 4 different velocity field basis (SVD, QR, RREF, SPARSE) with 5 cells ($N_\mathcal{P}=5$) and 4 degrees of freedom ($d=4$).}
    \label{fig:basis_velocity}
    \end{center}
    \vskip -0.2in
\end{figure}

\clearpage
\begin{figure}[!htb]
    \vskip 0.2in
    \begin{center}
    \centerline{\includegraphics[width=0.5\linewidth]{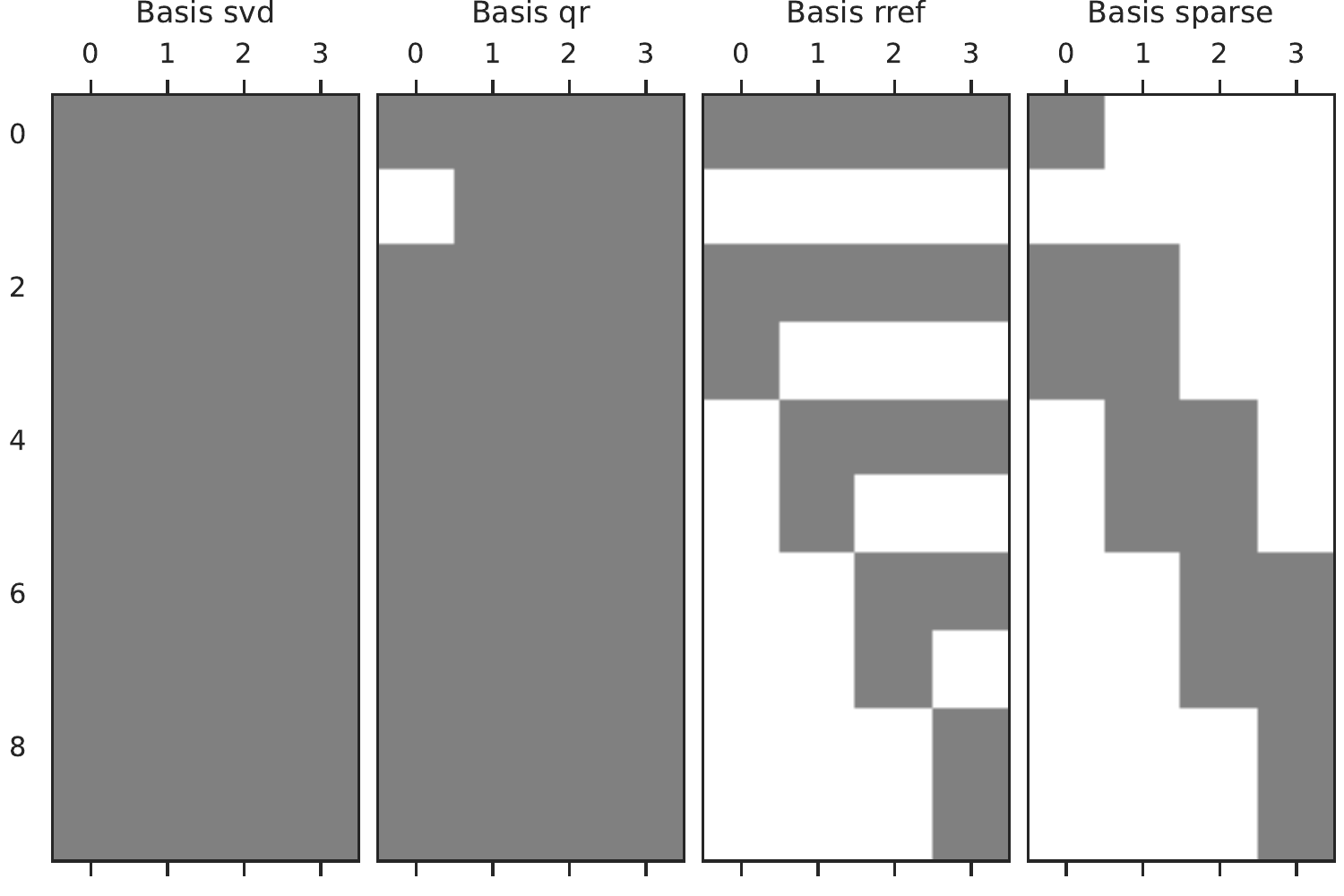}}
    \caption{Sparsity pattern of 4 different velocity field basis (SVD, QR, SPARSE, RREF) with 5 cells ($N_\mathcal{P}=5$) and 4 degrees of freedom ($d=4$). Columns indicate each of the basis vectors and gray cells indicate the non-zero values of the basis.}
    \label{fig:basis_spy}
    \end{center}
    \vskip -0.2in
\end{figure}

\begin{figure}[!htb]
    \vskip 0.2in
    \begin{center}
    \centerline{\includegraphics[width=0.6\linewidth]{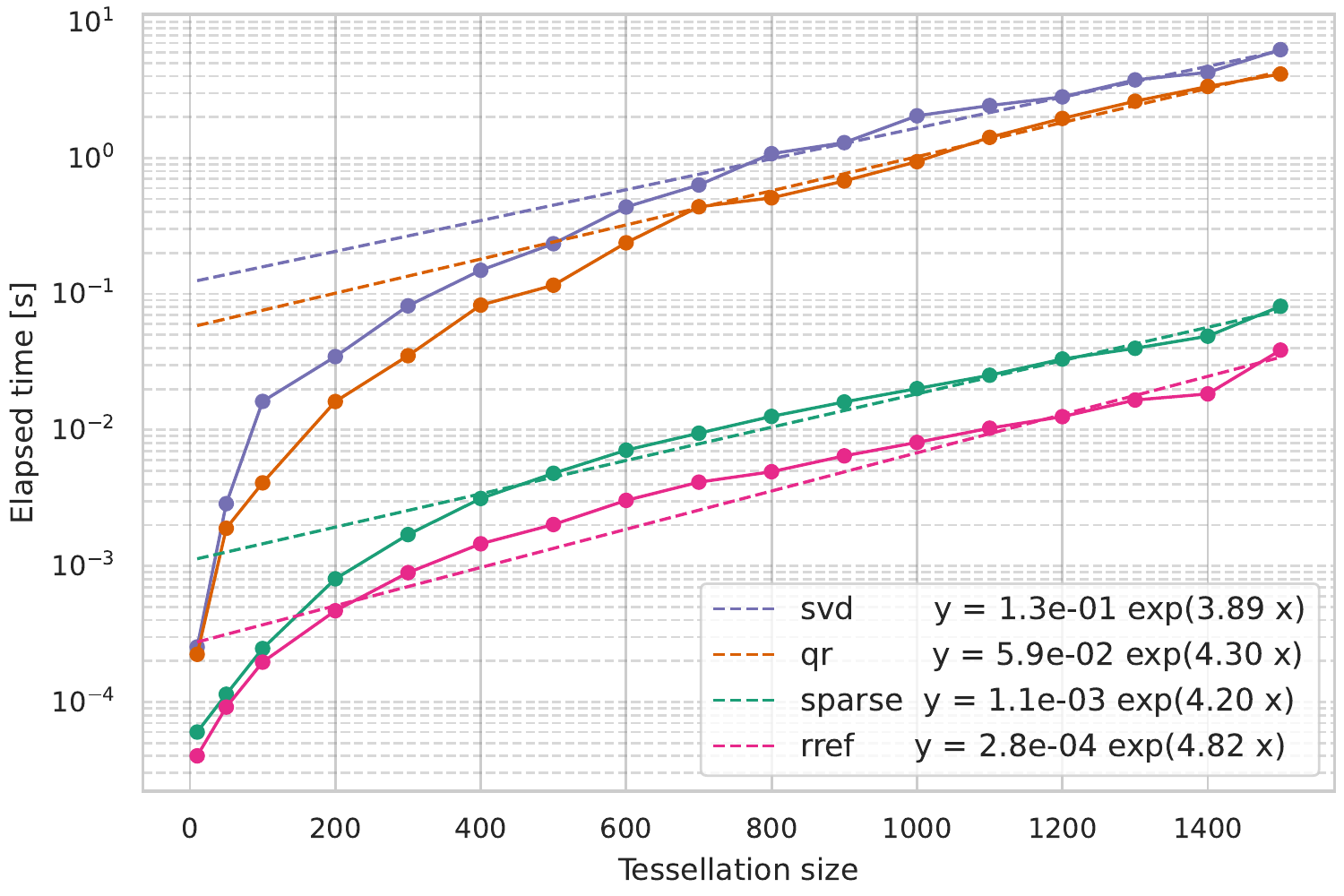}}
    \caption{Computation times (in seconds) for different velocity field basis and tessellation sizes.}
    \label{fig:basis_benchmark_fitted}
    \end{center}
    \vskip -0.2in
\end{figure}

\clearpage
\section{Integration Details}\label{apx:integration_details}

Given the initial value problem (IVT) defined by an ordinary differential equation (ODE) together with an initial condition:
\begin{equation}
\frac{d\psi}{dt}=v^\theta(\psi)=a^\theta \psi + b^\theta \quad; \quad \psi(x,0) = x
\end{equation}

\subsection{Analytical Solution}\label{apx:analytical_solution}

To obtain the analytical solution we follow the separation of variables method, in which algebra allows one to rewrite the equation so that each of two variables occurs on a different side of the equation. We rearrange the ODE and integrate both sides:
\begin{equation}
\frac{d\psi}{dt}=a^\theta \psi + b^\theta \rightarrow 
\frac{d\psi}{a^\theta \psi + b^\theta}=dt \rightarrow
\frac{1}{a^\theta}\int\frac{a^\theta}{a^\theta \psi + b^\theta}d\psi=\int dt 
\end{equation}

\begin{gather}
\frac{1}{a^\theta}\log (a^\theta \psi + b^\theta)=t + C \\
a^\theta \psi + b^\theta = \exp (t a^\theta + C a^\theta ) \\
\psi = \frac{1}{a^\theta} \Big( \exp (t a^\theta + C a^\theta ) -  b^\theta \Big)
\end{gather}

The initial condition is then used to solve the unknown integration constant $C$:
\begin{equation}
\psi(x,0) = x \rightarrow C = \frac{1}{a^\theta} \log (a^\theta x + b^\theta)
\end{equation}

With that, we can obtain the closed-form expression for the IVP:
\begin{equation}
\psi = \frac{1}{a^\theta} \Big( \exp \big(t a^\theta + \frac{a^\theta}{a^\theta} \log (a^\theta x + b^\theta)\big) -  b^\theta \Big)  = 
\frac{1}{a^\theta} \Big( (a^\theta x + b^\theta) e^{t a^\theta} -  b^\theta \Big)
\end{equation}

\begin{equation}
\boxed{\psi = x e^{t a^\theta} + \Big(e^{t a^\theta}-1\Big) \frac{b^\theta}{a^\theta}}
\end{equation}

\subsection{Matrix Form Solution}\label{apx:matrix_form_solution}

As an alternative, we can compute the IVT solution by transforming the system into matrix form and then using the matrix exponential operation. First, let's rearrange the ODE into a $2\times2$ matrix:

\begin{equation}
\frac{d\psi}{dt}=a^\theta \psi + b^\theta \rightarrow 
\begin{bmatrix} \frac{d\psi}{dt} \\ 0 \end{bmatrix} = 
\begin{bmatrix} a^\theta & b^\theta \\ 0 & 0 \end{bmatrix}\begin{bmatrix} \psi \\ 1 \end{bmatrix} \rightarrow 
\frac{d\tilde{\psi}}{dt} = 
\tilde{A} \tilde{\psi} 
\end{equation}

where $A^\theta = \begin{bmatrix} a^\theta & b^\theta \end{bmatrix}$ is the affine vector, $\tilde{A} = 
\begin{bmatrix} A \\ \vec{0} \end{bmatrix} = 
\begin{bmatrix} a^\theta & b^\theta \\ 0 & 0 \end{bmatrix} $ is the augmented affine matrix, and $\tilde{x} = \begin{bmatrix} x \\ 1 \end{bmatrix}$ and $\tilde{\psi} = \begin{bmatrix} \psi \\ 1 \end{bmatrix}$ are augmented vectors for $x$ and $\psi$ respectively.

The solution to the augmented ODE $\dot{\tilde{\psi}} = \tilde{A} \tilde{\psi}$ is given via the exponential matrix action: $\tilde{\psi} = \tilde{x} \cdot e^{t\tilde{A}}$.
The matrix exponential of a matrix $M$ can be mathematically defined by the Taylor series expansion, likewise to the exponential function $e^{x}$:
\begin{equation}
e^{x} = \sum_{k=0}^{\infty} \frac{x^k}{k!} \quad \rightarrow \quad e^{M} = \sum_{k=0}^{\infty} \frac{M^k}{k!}
\end{equation}
In this case, $M = t \tilde{A} = \begin{bmatrix} ta^\theta & tb^\theta \\ 0 & 0 \end{bmatrix}$.
To calculate the matrix exponential of $M$, we need to derive an expression for the infinite powers of $M$. This can be obtained by generalizing the sequence of matrix products, or by using the Cayley Hamilton theorem.

a) If we try to generalize the sequence of matrix product: 
\begin{quote}
\begin{equation}
M^0 = \mathbb{I} = \begin{bmatrix} 1 & 0 \\ 0 & 1 \end{bmatrix}
\end{equation}

\begin{equation}
M^1 = M = \begin{bmatrix} ta^\theta & tb^\theta \\ 0 & 0 \end{bmatrix}
\end{equation}

\begin{equation}
M^2 = M \cdot M = \begin{bmatrix} ta^\theta & tb^\theta \\ 0 & 0 \end{bmatrix}\begin{bmatrix} ta^\theta & tb^\theta \\ 0 & 0 \end{bmatrix}=\begin{bmatrix} t^2(a^\theta)^2 & a^\theta b^\theta t^2 \\ 0 & 0 \end{bmatrix}
\end{equation}

\begin{equation}
M^3 = M^2 \cdot M = \begin{bmatrix} t^2(a^\theta)^2 & a^\theta b^\theta t^2 \\ 0 & 0 \end{bmatrix}\begin{bmatrix} ta^\theta & tb^\theta \\ 0 & 0 \end{bmatrix} = 
\begin{bmatrix} t^3(a^\theta)^3 & (a^\theta)^2 b^\theta t^3 \\ 0 & 0 \end{bmatrix}\\ \vdots
\end{equation}

\begin{equation}
M^k = M^{k-1} \cdot M = 
\begin{bmatrix} t^k(a^\theta)^k & (a^\theta)^{k-1} b^\theta t^k \\ 0 & 0 \end{bmatrix} = 
t^k(a^\theta)^k \begin{bmatrix} 1 & \frac{b^\theta}{a^\theta} \\ 0 & 0 \end{bmatrix}
\end{equation}

Then, 
\begin{equation*}
e^{M} = \sum_{k=0}^{\infty} \frac{M^k}{k!} = \frac{M^0}{0!} + \sum_{k=1}^{\infty} \frac{M^k}{k!} = \mathbb{I} + \sum_{k=1}^{\infty} \frac{M^k}{k!} = 
\end{equation*}

\begin{equation} 
= 
\begin{bmatrix} 1 & 0 \\ 0 & 1 \end{bmatrix} + 
\sum_{k=1}^{\infty} \frac{t^k(a^\theta)^k}{k!} \begin{bmatrix} 1 & \frac{b^\theta}{a^\theta} \\ 0 & 0 \end{bmatrix} = 
\begin{bmatrix} 1 & 0 \\ 0 & 1 \end{bmatrix} + 
(e^{ta^\theta}-1) \begin{bmatrix} 1 & \frac{b^\theta}{a^\theta} \\ 0 & 0 \end{bmatrix} =
\begin{bmatrix} e^{ta^\theta} & (e^{ta^\theta}-1)\frac{b^\theta}{a^\theta} \\ 0 & 1 \end{bmatrix}
\end{equation}

\end{quote}

b) Rather, using the Cayley Hamilton theorem: 
\begin{quote}
The characteristic polynomial of the matrix $M$ is given by:
$p(\lambda) = \text{det}(\lambda I_2 - A) = 
\begin{vmatrix}
\lambda - ta^\theta & tb^\theta\\ 
0 & \lambda 
\end{vmatrix} = 
\lambda^2 - ta^\theta \lambda
$

The Cayley–Hamilton theorem claims that, if we define $p(X) = X^2 - ta^\theta X$ then $p(M) = M^2 - ta^\theta M = 0_{2,2}$. Hence, $M^2 =  ta^\theta M$ and as a consequence, $M^k = ta^\theta M^{k-1} = (ta^\theta)^{k-1} M$.
Now, using the exponential matrix expression we arrive at the same outcome as before: 
\begin{equation*}
e^{M} = 
\mathbb{I} + \sum_{k=1}^{\infty} \frac{M^k}{k!} = 
\mathbb{I} + \sum_{k=1}^{\infty} \frac{(ta^\theta)^{k-1}}{k!} M = 
\mathbb{I} + \frac{1}{ta^\theta} \sum_{k=1}^{\infty} \frac{(ta^\theta)^{k}}{k!} M = 
\mathbb{I} + \frac{1}{ta^\theta} (e^{ta^\theta}-1) M = 
\end{equation*}

\begin{equation}
=\begin{bmatrix} 1 & 0 \\ 0 & 1 \end{bmatrix} + 
\frac{e^{ta^\theta}-1}{ta^\theta} 
\begin{bmatrix} ta^\theta & tb^\theta \\ 0 & 0 \end{bmatrix} = 
\begin{bmatrix} e^{ta^\theta} & (e^{ta^\theta}-1)\frac{b^\theta}{a^\theta} \\ 0 & 1 \end{bmatrix}
\end{equation}

\end{quote}

Finally, from these two identical expressions, we can extract the same result obtained in the analytical solution:
\begin{equation}
e^{M} = e^{t \tilde{A}}=\begin{bmatrix} e^{ta^\theta} & (e^{ta^\theta}-1)\frac{b^\theta}{a^\theta} \\ 0 & 1 \end{bmatrix}
\end{equation}
\begin{equation}
\tilde{\psi} = \tilde{x} \cdot e^{t\tilde{A}} \quad \rightarrow \quad 
\begin{bmatrix} \psi \\ 1 \end{bmatrix} = 
\begin{bmatrix} x \\ 1 \end{bmatrix}
\begin{bmatrix} e^{ta^\theta} & (e^{ta^\theta}-1)\frac{b^\theta}{a^\theta} \\ 0 & 1 \end{bmatrix} \quad \rightarrow \quad 
\boxed{\psi = x e^{t a^\theta} + \Big(e^{t a^\theta}-1\Big) \frac{b^\theta}{a^\theta}}
\end{equation}

\clearpage
\section{Comparing ODE Numerical and Closed-Form Methods for CPA Velocity Functions}\label{apx:comparing_ode_numerical_closed_form}

In this article, we construct diffeomorphic curves by the integration of velocity functions. Specifically we work with continuous piecewise affine (CPA) velocity functions. The integral equation or the equivalent ODE can be solved numerically using any generic ODE solver. 

\begin{figure*}[!htb]
    \vskip 0.2in
    \begin{center}
    \subfigure[Velocity field $v(x)$ is a continuous piecewise affine function of five intervals. The integration of the flow equation with the initial condition $\phi(x, t_{0}) = x$ is shown for different integration times $t=\{0, 1/6, 2/6,\cdots, 1\}$. The gray curve corresponds to the final diffeomorphism at $t=1$.]{%
    \includegraphics[width=0.48\linewidth]{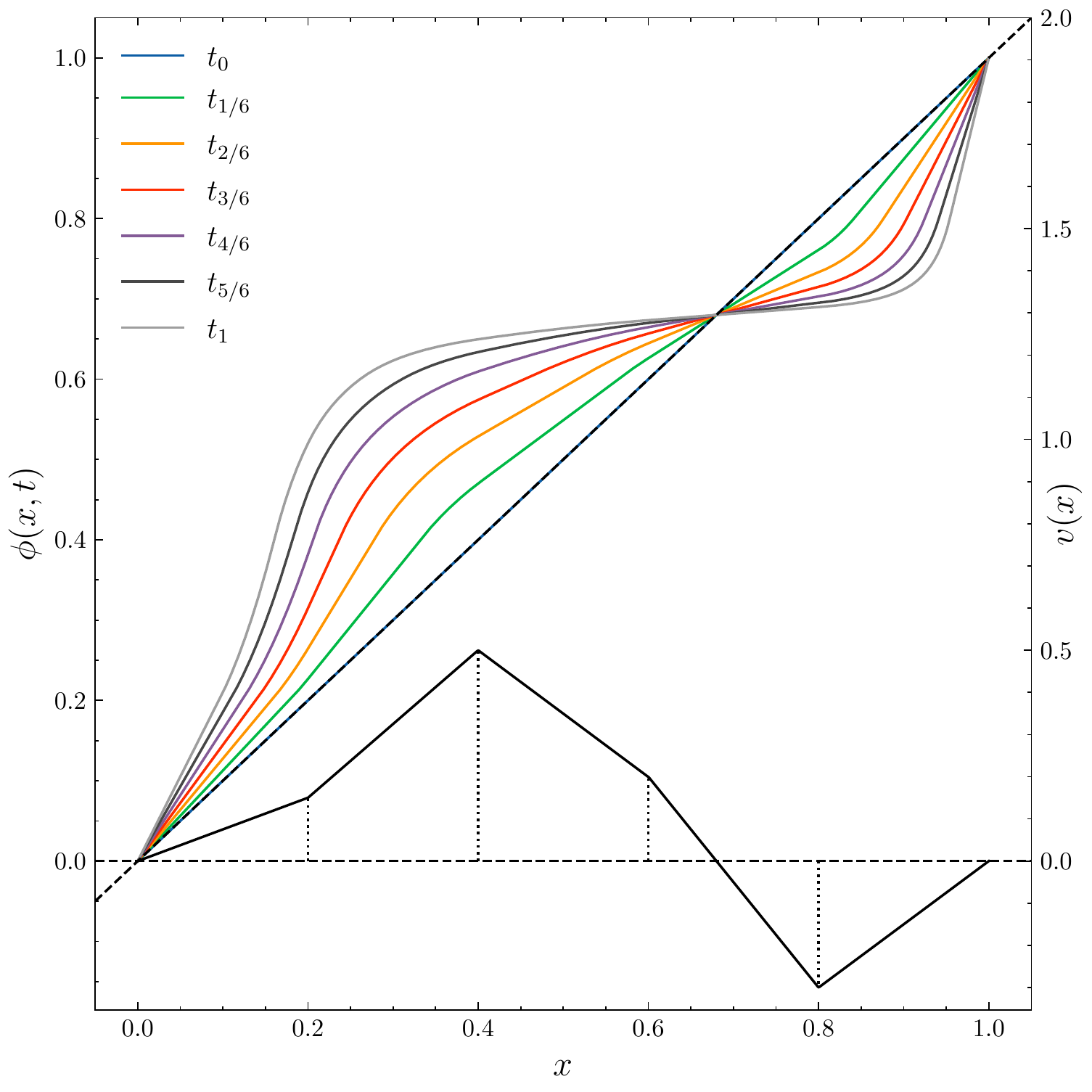}
    \label{fig:integration_1}}
    \hfill
    \subfigure[Illustration of the numerical integration of the flow equation: $\phi(x,t+h)=\phi(x,t) + h \cdot v(x)$. The velocity field $v(x)$ is shown along the y-axis. This scheme produces only increasing functions (invertible one-dimensional functions), when the integration step $h$ is chosen small enough.]{%
    \includegraphics[width=0.48\linewidth]{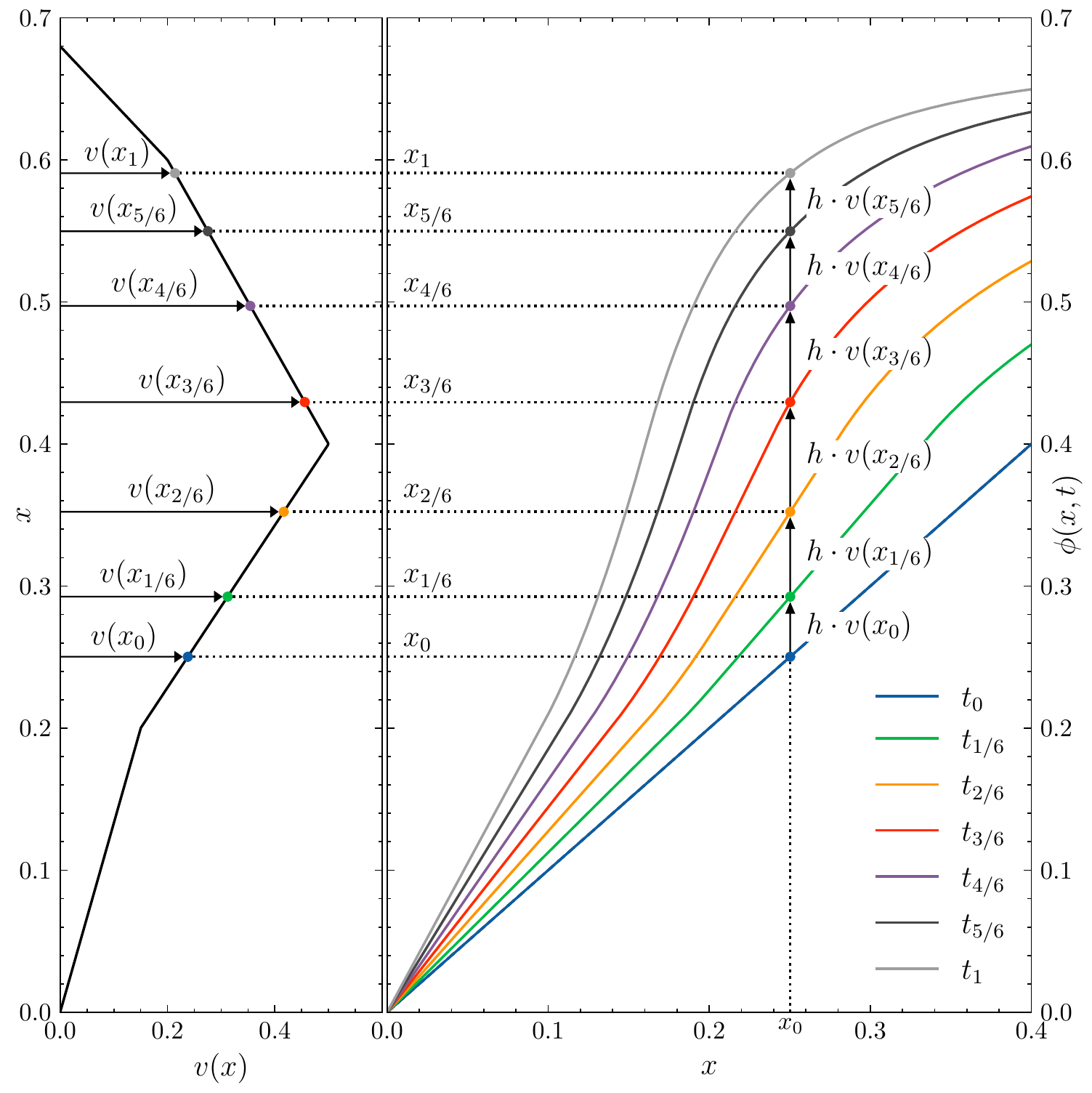}
    \label{fig:integration_2}}
    \caption{Construction of diffeomorphic curves by integration of velocity functions. In this article, we work with continuous piecewise affine (CPA) velocity functions.}
    \label{fig:integration}
    \end{center}
    \vskip -0.2in
\end{figure*}

In this sense, \cite{Freifeld2015,Freifeld2017} proposed a specialized solver for integrating CPA fields which alternates between the analytic solution and a generic solver. This specialized solver is faster and more accurate than a generic solver since most of the trajectory is solved exactly, while the generic solver is only called in small portions. The two parameters used in \cite{Freifeld2017}, $N_{steps}$ and $n_{steps}$ imply two step sizes, one large $\Delta t = t/N_{steps}$ used for exact updates, and one small, $\delta t = \Delta t / n_{steps}$ used for the generic-solver subroutine. By construction, the proposed closed-form solution is more accurate (exact solution) for both the ODE solution and for its derivative with respect to the velocity function parameters. We verified this empirically, comparing the specialized solver presented in \cite{Freifeld2015,Freifeld2017} against the closed-form methods described in \cref{sec:closed_form_integration,sec:closed_form_derivatives}. We compared the integration (and gradient) error averaged over random CPA velocity fields; i.e. we drew 100 CPA velocity functions from the prior (normal distribution of zero mean and unit standard deviation), $\{v^{\theta_i}\}_{i=1}^{100}$, sampled uniformly 1000 points in $\Omega$, $\{x^{\theta_i}\}_{i=1}^{1000}$, and then computed the error:

\begin{equation}\label{eq:error_comparison}
\epsilon = \frac{1}{100}\sum_{i=1}^{100} \; ||\text{abs}(\phi_{exact}^{\theta_i}(x_j) - \phi_{solver}^{\theta_i}(x_j))||_{\infty} \quad \forall j \in [1,1000]
\end{equation}

\begin{figure*}[!htb]
    \vskip 0.2in
    \begin{center}
    \subfigure[Integration error]{%
    \includegraphics[width=0.55\linewidth]{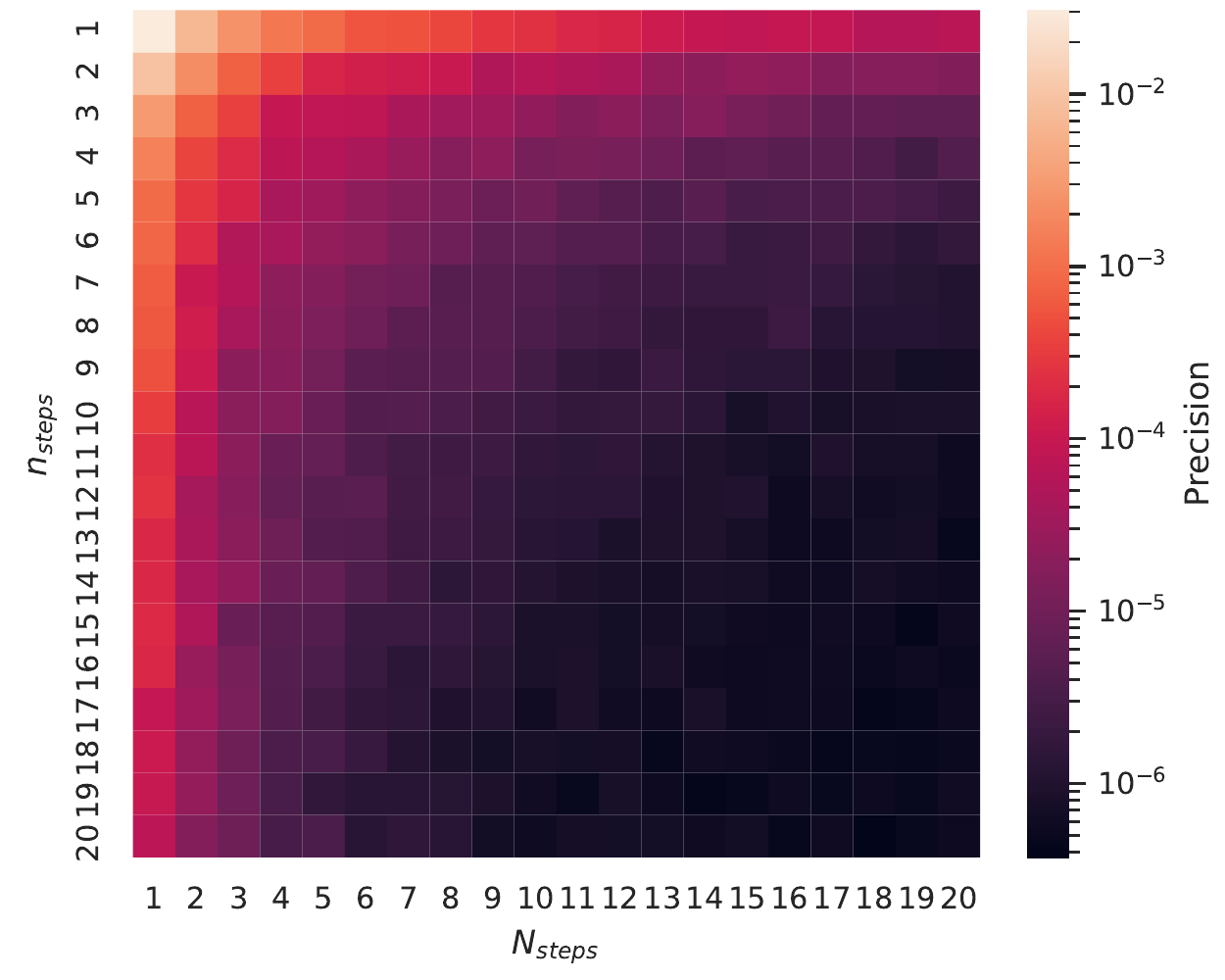}
    \label{fig:error_integration}}
    \hfill
    \subfigure[Gradient error]{%
    \includegraphics[width=0.55\linewidth]{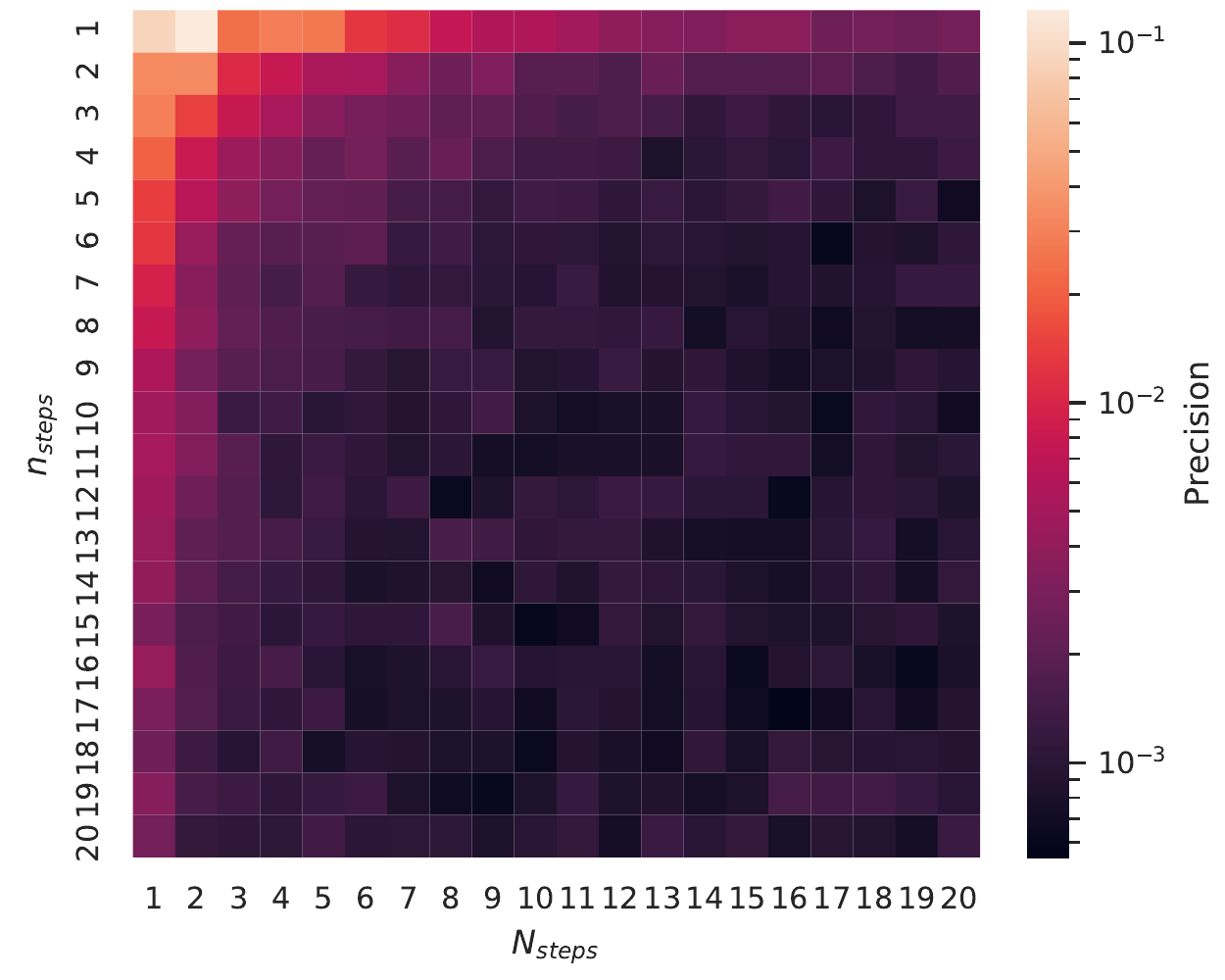}
    \label{fig:error_gradient}}
    \caption{Comparing ODE numerical and closed-form methods for random CPA Velocity Functions: Precision obtained for different values of $N_{steps} \in [1,20]$ and $n_{steps} \in [1,20]$}
    \label{fig:error_integration_numerical}
    \end{center}
    \vskip -0.2in
\end{figure*}

\clearpage
The intention with this analysis was to study the number of numerical steps necessary to guarantee good enough integration and gradient estimations. The warping function is a bijective map from $(0,1)$ to $(0,1)$.
\cref{fig:error_integration_numerical} shows the precision obtained for different values of $N_{steps} \in [1,20]$ and $n_{steps} \in [1,20]$. For instance, using the recommended values for $N_{steps}$ and $n_{steps}]$ by the authors in \cite{Freifeld2017} ($N_{steps}=10$ and $n_{steps}=10$) we get a 5 decimals precision for the integration result but only 3 decimals for the gradient computation. Recall that \cref{eq:error_comparison} computes the maximum absolute difference, and that the total error can be much higher over the entire gradient function. Lack of precision in the computation of transformation gradient translates to inefficient search in the parameter space, which leads to worse training processes and non-optimal solutions. 

\clearpage
\section{Closed-Form Derivatives of $T^{\theta}(x)$ w.r.t. $\theta$}\label{apx:closed_form_derivatives}

Let's recall that the trajectory (the solution for the integral equation \ref{eq:integral}) is a composition of a finite number of solutions $\psi$, as given by:
\begin{equation}
\phi^\theta(x,t) = \Big(\psi_{\theta,c_m}^{t_m} \circ \psi_{\theta,c_{m-1}}^{t_{m-1}} \circ \cdots \circ \psi_{\theta,c_2}^{t_2} \circ \psi_{\theta,c_1}^{t_1} \Big)(x)
\end{equation}
During the iterative process of integration, several cells are crossed, starting from $c_1$ at integration time $t_1=1$, and finishing at $c_m$ at integration time $t_m$. The integration time $t_m$ of the last cell $c_m$ can be calculated by subtracting from the initial integration time the accumulated boundary hitting times $t_{hit}$: $t_m = t_1 - \sum_{i=1}^{m-1} t_{hit}^\theta(c_i, x_i)$. The final integration point $x_m$ is the boundary of the penultimate cell $c_{m-1}$: $x_m = x_{c_{m-1}}$. In case only one cell is visited, both time and space remain unchanged: $t_m = 1$ and $x_m = x$. Taking this into consideration, the trajectory can be calculated as follows:
\begin{equation}
\phi^\theta(x,t) = \psi^\theta(x=x_m,t=t_m) = 
\bigg(
    x e^{t a_c} + \Big(e^{t a_c}-1\Big) \frac{b_c}{a_c}
\bigg)_{\substack{x = x_m \\ t = t_m}}
\end{equation}
In this section both the $\theta$-dependent $a^\theta$ and $b^\theta$ are expressed as $a$ and $b$ to avoid overloading the notation.
Therefore, the derivative can be calculated by going backwards in the integration direction. We are interested in the derivative of the trajectory w.r.t. the parameters 
$\theta = \begin{bmatrix} \theta_1 & \theta_2 & \cdots & \theta_d \end{bmatrix} \in 
\mathbb{R}^{d}$. To obtain such gradient, we focus on the partial derivative w.r.t. one of the coefficients of $\theta$, i.e., $\theta_k$:
\begin{equation}\label{eq:derivative_chain_rule}
\frac{\partial \phi^\theta(x,t)}{\partial \theta_k} = 
\bigg(
\frac{\partial \psi^\theta(x,t)}{\partial \theta_k} + 
\frac{\partial \psi^\theta(x,t)}{\partial t^\theta} \cdot
\frac{\partial t^\theta}{\partial \theta_k} + 
\frac{\partial \psi^\theta(x,t)}{\partial x} \cdot
\frac{\partial x}{\partial \theta_k}
\bigg)_{\substack{x = x_m \\ t = t_m}}
\end{equation}
Let’s derive each of the terms of this derivative:

\begin{center}
$\cfrac{\partial \psi^\theta(x,t)}{\partial \theta_k}\;,\;$
$\cfrac{\partial \psi^\theta(x,t)}{\partial t^\theta}\;,\;$ 
$\cfrac{\partial t^\theta}{\partial \theta_k}\;,\;$
$\cfrac{\partial x}{\partial \theta_k}$.
\end{center}

\begin{quote}
\begin{center}
\rule{0.9\textwidth}{.4pt}
\end{center}
\textbf{Notation.} Recall the expression for $\psi(x,t)$ from \cref{sec:cpa_diffeomorphic_transformations}:
\begin{equation}\label{eq:psi_expression_1}
\psi^\theta(x,t) = x e^{t a_c} + \Big(e^{t a_c}-1\Big) \frac{b_c}{a_c}
\end{equation}
Note that the slope $a_c$ and intercept $b_c$ are a linear combination of the orthogonal basis $B$ of the constraint matrix $L$, with $\theta$ as coefficients.
\begin{equation}\label{eq:vec_A}
vec(\textbf{A}) = \textbf{B} \cdot \boldsymbol{\theta} = \sum_{j=1}^{d} \theta_j \cdot \textbf{B}_j
\end{equation}
If we define one of the components from the orthogonal basis $\textbf{B}_j$ as:
\begin{equation}\label{eq:basis}
\textbf{B}_{j} = \begin{bmatrix} a_1^{(j)} & b_1^{(j)} & \cdots & a_c^{(j)} & b_c^{(j)} & \cdots & a_{N_\mathcal{P}}^{(j)} & b_{N_\mathcal{P}}^{(j)}\end{bmatrix}^T
\end{equation}
Then,
\begin{equation}\label{eq:theta_dot_basis}
\theta_j \cdot \textbf{B}_j = \begin{bmatrix} & \cdots & \theta_j a_c{(j)} & \theta_j b_c{(j)} & \cdots & \end{bmatrix}^T
\end{equation}
And as a result,
\begin{equation}\label{eq:vec_A_complete}
vec(\textbf{A}) = \sum_{j=1}^{d} \theta_j \cdot \textbf{B}_j = \begin{bmatrix} & \cdots & \sum_{j=1}^{d} \theta_j a_c^{(j)} & \sum_{j=1}^{d} \theta_j b_c^{(j)} & \cdots & \end{bmatrix}^T
\end{equation}
Thus, the slope $a_c$ and intercept $b_c$ ( parameters of the affine transformation) are denoted as follows:
\begin{equation}\label{eq:ac_bc}
a_c = \sum_{j=1}^{d} \theta_j a_c^{(j)} \quad\quad\quad
b_c = \sum_{j=1}^{d} \theta_j b_c^{(j)}
\end{equation}
\begin{center}
\rule{0.9\textwidth}{.4pt}
\end{center}
\end{quote}

\subsection{Expression for $\cfrac{\partial \psi^\theta(x,t)}{\partial \theta_k}$}
\label{apx:expression_derivative_1}

On the basis of the expression for $\psi(x,t)$,
\begin{equation}\label{eq:psi_expression_2}
\psi^\theta(x,t) = x e^{t a_c} + \Big(e^{t a_c}-1\Big) \frac{b_c}{a_c}
\end{equation}
apply the chain-rule to obtain the derivative w.r.t $\theta_k$:
\begin{equation}\label{eq:psi_derivative}
\frac{\partial \psi^\theta(x,t)}{\partial \theta_k} = 
\frac{\partial \psi^\theta(x,t)}{\partial a_c} \cdot
\frac{\partial a_c}{\partial \theta_k} +
\frac{\partial \psi^\theta(x,t)}{\partial b_c} \cdot
\frac{\partial b_c}{\partial \theta_k}
\end{equation}
Considering \cref{eq:psi_expression_2} it is immediate to obtain the partial derivatives of $a_c$ and $b_c$ w.r.t $\theta_k$:
\begin{equation}\label{eq:ac_derivative}
a_c = \sum_{j=1}^{d} \theta_j a_c^{(j)} \rightarrow \frac{\partial a_c}{\partial \theta_k} = a_c^{(k)}
\end{equation}
\begin{equation}\label{eq:bc_derivative}
b_c = \sum_{j=1}^{d} \theta_j b_c^{(j)} \rightarrow \frac{\partial b_c}{\partial \theta_k} = b_c^{(k)}
\end{equation}
Next, we find the derivative $\psi(x,t)$ w.r.t $a_c$ and $b_c$:
\begin{equation}\label{eq:psi_derivative_ac}
\frac{\partial \psi^\theta(x,t)}{\partial a_c} = 
x \, t \, e^{t a_c} + \Big(t\,e^{t a_c}\Big)\frac{b_c}{a_c} - \Big(e^{t a_c}-1\Big)\frac{b_c}{a_c^2} =
t \, e^{t a_c} \Big(x + \frac{b_c}{a_c} \Big) - \Big(e^{t a_c}-1\Big)\frac{b_c}{a_c^2}
\end{equation}
\begin{equation}\label{eq:psi_derivative_bc}
\frac{\partial \psi^\theta(x,t)}{\partial b_c} = 
\Big(e^{t a_c}-1\Big)\frac{1}{a_c}
\end{equation}
Finally, these expressions (\cref{eq:ac_derivative,eq:bc_derivative,eq:psi_derivative_ac,eq:psi_derivative_bc}) are joined together into \cref{eq:psi_derivative}:
\begin{equation}\label{eq:psi_derivative_developed}
\begin{split}
\frac{\partial \psi^\theta(x,t)}{\partial \theta_k} &= 
\frac{\partial \psi^\theta(x,t)}{\partial a_c} \cdot
\frac{\partial a_c}{\partial \theta_k} +
\frac{\partial \psi^\theta(x,t)}{\partial b_c} \cdot
\frac{\partial b_c}{\partial \theta_k} = \\
&=
\underbrace{
\Bigg(t \, e^{t a_c} \Big(x + \frac{b_c}{a_c} \Big) - \Big(e^{t a_c}-1\Big)\frac{b_c}{a_c^2}\Bigg) 
}_{\cfrac{\partial \psi^\theta(x,t)}{\partial a_c}}
\cdot
\underbrace{\vphantom{\Bigg(}
a_c^{(k)}
}_{\cfrac{\partial a_c}{\partial \theta_k}}
+
\underbrace{\vphantom{\Bigg(}
\Big(e^{t a_c}-1\Big)\frac{1}{a_c} 
}_{\cfrac{\partial \psi^\theta(x,t)}{\partial b_c}}
\cdot
\underbrace{\vphantom{\Bigg(}
b_c^{(k)}
}_{\cfrac{\partial b_c}{\partial \theta_k}}
= \\
&= 
a_c^{(k)} \, t \, e^{t a_c} \Big(x + \frac{b_c}{a_c} \Big) + \Big(e^{t a_c}-1\Big)\frac{b_c^{(k)} \, a_c - a_c^{(k)} \, b_c}{a_c^2}
\end{split}
\end{equation}

\subsection{Expression for $\cfrac{\partial \psi^\theta(x,t)}{\partial t^\theta}$}
\label{apx:expression_derivative_2}

Starting from the expression for $\psi(x,t)$,
\begin{equation}\label{eq:psi_expression_3}
\psi^\theta(x,t) = x e^{t a_c} + \Big(e^{t a_c}-1\Big) \frac{b_c}{a_c}
\end{equation}
we explicitly get the derivative w.r.t $t^\theta$:
\begin{equation}\label{eq:psi_derivative_t}
\frac{\partial \psi^\theta(x,t)}{\partial t^\theta} = x \, a_c \, e^{t a_c} + a_c \, e^{t a_c} \frac{b_c}{a_c} = 
e^{t a_c} \Big( a_c x + b_c \Big)
\end{equation}

\subsection{Expression for $\cfrac{\partial t^\theta}{\partial \theta_k}$}
\label{apx:expression_derivative_3}

After visiting $m$ cells, the integration time $t^\theta$ can be expressed as:
\begin{equation}\label{eq:time}
t^\theta = t_1 - \sum_{i=1}^{m-1} t_{hit}^\theta(c_i, x_i)
\end{equation}
The derivative w.r.t. $\theta_k$ is obtained as follows:
\begin{equation}\label{eq:time_derivative_theta}
\frac{\partial t^\theta}{\partial \theta_k} = 
-\sum_{i=1}^{m-1} \frac{\partial t_{hit}^\theta(c_i, x_i)}{\partial \theta_k}
\end{equation}
where
\begin{equation}\label{eq:thit}
t_{hit}^\theta(c, x) = \frac{1}{a_c} \log \bigg( \frac{a_c x_c + b_c}{a_c x + b_c} \bigg)
\end{equation}
and $x_c$ is the boundary for cell index $c$. Now, apply the chain rule operation to the hitting time $t_{hit}^\theta(c, x)$ expression:
\begin{equation}\label{eq:thit_derivative_theta}
\frac{\partial t_{hit}^\theta(c, x)}{\partial \theta_k} = 
\frac{\partial t_{hit}^\theta(c, x)}{\partial a_c} \cdot 
\frac{\partial a_c}{\partial \theta_k} +
\frac{\partial t_{hit}^\theta(c, x)}{\partial b_c} \cdot 
\frac{\partial b_c}{\partial \theta_k}
\end{equation}
As it was previously derived in \cref{apx:expression_derivative_2}, considering \cref{eq:psi_expression_2} it is immediate to obtain the partial derivatives of $a_c$ and $b_c$ w.r.t $\theta_k$:
\begin{equation}\label{eq:ac_derivative_theta}
a_c = \sum_{j=1}^{d} \theta_j a_c^{(j)} \rightarrow \frac{\partial a_c}{\partial \theta_k} = a_c^{(k)}
\end{equation}
\begin{equation}\label{eq:bc_derivative_theta}
b_c = \sum_{j=1}^{d} \theta_j b_c^{(j)} \rightarrow \frac{\partial b_c}{\partial \theta_k} = b_c^{(k)}
\end{equation}
Next, we develop the derivatives of $t_{hit}^\theta(c, x)$ w.r.t $a_c$ and $b_c$:
\begin{equation}\label{eq:thit_derivative_ac}
\begin{split}
\frac{\partial t_{hit}^\theta(c, x)}{\partial a_c} &= 
-\frac{1}{a_c^2} \log \bigg( \frac{a_c x_c + b_c}{a_c x + b_c} \bigg) + 
\frac{1}{a_c} \frac{x_c (a_c x + b_c) - x (a_c x_c + b_c)}{(a_c x + b_c)(a_c x_c + b_c)} = \\ 
&= -\frac{1}{a_c^2} \log \bigg( \frac{a_c x_c + b_c}{a_c x + b_c} \bigg) + 
\frac{b_c}{a_c} \frac{x_c - x}{(a_c x + b_c)(a_c x_c + b_c)}
\end{split}
\end{equation}
\begin{equation}\label{eq:thit_derivative_bc}
\frac{\partial t_{hit}^\theta(c, x)}{\partial b_c} = 
\frac{1}{a_c} \frac{(a_c x + b_c) - (a_c x_c + b_c)}{(a_c x + b_c)(a_c x_c + b_c)} = 
\frac{x - x_c}{(a_c x + b_c)(a_c x_c + b_c)}
\end{equation}
Finally, these expressions (\cref{eq:ac_derivative_theta,eq:bc_derivative_theta,eq:thit_derivative_ac,eq:thit_derivative_bc}) are joined together to obtain the derivative of $t_{hit}^\theta(c, x)$ w.r.t $\theta_k$ into \cref{eq:thit_derivative_theta}:
\begin{equation}\label{eq:thit_derivative_theta_complete}
\begin{split}
\frac{\partial t_{hit}^\theta(c, x)}{\partial \theta_k} &= 
\frac{\partial t_{hit}^\theta(c, x)}{\partial a_c} \cdot 
\frac{\partial a_c}{\partial \theta_k} +
\frac{\partial t_{hit}^\theta(c, x)}{\partial b_c} \cdot 
\frac{\partial b_c}{\partial \theta_k} = \\
&= 
\underbrace{\vphantom{\Bigg(}
\Bigg(
    -\frac{1}{a_c^2} \log \bigg( \frac{a_c x_c + b_c}{a_c x + b_c} \bigg) + 
    \frac{b_c}{a_c} \frac{x_c - x}{(a_c x + b_c)(a_c x_c + b_c)}
\Bigg) 
}_{\cfrac{\partial t_{hit}^\theta(c, x)}{\partial a_c}}
\cdot
\underbrace{\vphantom{\Bigg(}
a_c^{(k)}
}_{\cfrac{\partial a_c}{\partial \theta_k}} +
\underbrace{\vphantom{\Bigg(}
\frac{x - x_c}{(a_c x + b_c)(a_c x_c + b_c)} 
}_{\cfrac{\partial t_{hit}^\theta(c, x)}{\partial b_c}}
\cdot
\underbrace{\vphantom{\Bigg(}
b_c^{(k)} 
}_{\cfrac{\partial b_c}{\partial \theta_k}} =\\
&=
-\frac{a_c^{(k)}}{a_c^2} \log \bigg( \frac{a_c x_c + b_c}{a_c x + b_c} \bigg) +
\frac{(x - x_c)(b_c^{(k)} a_c - a_c^{(k)} b_c)}{a_c(a_c x + b_c)(a_c x_c + b_c)}
\end{split}
\end{equation}

\subsection{Expression for $\cfrac{\partial x}{\partial \theta_k}$}
\label{apx:expression_derivative_4}

After visiting $m$ cells, the integration point $x$ is the boundary point of the last visited cell. In case only one cell is visited, the integration point remains unchanged.
In either case, $x$ does not depend on the parameters $\theta$, thus 
\begin{equation}\label{eq:x_derivative_theta}
\frac{\partial x}{\partial \theta_k} = 0
\end{equation}

\subsection{Final Expression for $\cfrac{\partial \phi^\theta(x,t)}{\partial \theta_k}$}
\label{apx:expression_derivative_5}

Joining all the terms together and evaluating the derivative at $x = x_m$ and $t = t_m$ yields:
\begin{equation}\label{eq:derivative_complete_apx}
\boxed{
\begin{aligned}
\frac{\partial \phi^\theta(x,t)}{\partial \theta_k} &= 
\bigg(
\frac{\partial \psi^\theta(x,t)}{\partial \theta_k} + 
\frac{\partial \psi^\theta(x,t)}{\partial t^\theta} \cdot
\frac{\partial t^\theta}{\partial \theta_k} + 
\frac{\partial \psi^\theta(x,t)}{\partial x} \cdot
\frac{\partial x}{\partial \theta_k}
\bigg)_{\substack{x = x_m \\ t = t_m}} = \\
 &= a_{c_m}^{(k)} \, t_m \, e^{t_m a_{c_m}} \Big(x_m + \frac{b_{c_m}}{a_{c_m}} \Big) + 
\Big(e^{t_m a_{c_m}}-1\Big)\frac{b_{c_m}^{(k)} \, a_{c_m} - a_{c_m}^{(k)} \, b_{c_m}}{a_{c_m}^2} - \\ & \quad 
e^{t_m a_{c_m}} \Big( a_{c_m} x_m + b_{c_m} \Big) 
\sum_{i=1}^{m-1} 
\Bigg(
-\frac{a_{c_i}^{(k)}}{a_{c_i}^2} \log \bigg( \frac{a_{c_i} x_{c_i} + b_{c_i}}{a_{c_i} x_i + b_{c_i}} \bigg) +
 \frac{(x_{c_i} - x_i)(b_{c_i}^{(k)} a_{c_i} - a_{c_i}^{(k)} b_{c_i})}{a_{c_i}(a_{c_i} x_i + b_{c_i})(a_{c_i} x_{c_i} + b_{c_i})}
\Bigg)
\end{aligned}
}
\end{equation}

\begin{figure}[!ht]
    \vskip 0.2in
    \begin{center}
    \scalebox{1}{
    \begin{tikzpicture}
    \pgfmathsetmacro{\N}{5};
    \pgfmathsetmacro{\M}{5};
    \pgfmathsetmacro{\P}{\N-1};
    \pgfmathsetmacro{\a}{0.15};
    \pgfmathsetmacro{\h}{0.6};
    \draw[step=1, black!30, thin] (0,0) grid (\N, \M);
    \draw[step=1, black!80, thick] (2,2) grid (3, \M);
    
    \draw[step=1, black!30, thin, shift={(0,5+\h)}] (0,0) grid (\N,2);
    \draw[step=1, black!30, thin, shift={(5+\h,0)}] (0,0) grid (2,\N);
    \draw[step=1, black!30, thin, shift={(5+\h,5+\h)}] (0,0) grid (2,2);
    \draw[step=1, black!80, thick, shift={(2,5+\h)}] (0,0) grid (1,1);

    \draw[thin, black!50, densely dashed] (0,0) -- (\N,\M);
    \draw[thin, black!50, densely dashed] (\N+\h,\M+\h) -- (\N+2+\h,\M+2+\h);

    \draw[-latex, very thick] (0,0) -- (\N+2.1+\h,0) node[right] {$x$};
    \draw[-latex, very thick] (0,0) -- (0,\M+2.1+\h) node[above] {$\phi(x)$};
    
    \draw[red, thick, dashed] (2.5, 0) -- (2.5, 2.5);
    \draw[red, thick, dashed] (0, 2.5) -- (2.5, 2.5);
    \filldraw[red] (2.5, 2.5) circle (2pt);
    \draw (2.5,\a) -- (2.5,0)  node[anchor=north, pos=1.25] {$x$};
    \draw (\a,2.5) -- (-\a,2.5)  node[anchor=east] {$x$};

    \draw[black!70, font=\footnotesize] (0,0) -- (0,-0.25);
    \draw[black!70, font=\footnotesize] (1,0) -- (1,-0.25);
    \draw[black!70, font=\footnotesize] (2,0) -- (2,-0.25)  node[anchor=north] {$x_{c_1}^{min}$};
    \draw[black!70, font=\footnotesize] (3,0) -- (3,-0.25)  node[anchor=north] {$x_{c_1}^{max}$};
    \draw[black!70, font=\footnotesize] (4,0) -- (4,-0.25)  node[anchor=north] {$x_{c_2}^{max}$};
    \draw[black!70, font=\footnotesize] (5,0) -- (5,-0.25)  node[anchor=north] {$x_{c_3}^{max}$};
    \draw[black!70, font=\footnotesize] (5+\h,0) -- (5+\h,-0.25) ;
    \draw[black!70, font=\footnotesize] (6+\h,0) -- (6+\h,-0.25)  node[anchor=north] {$x_{c_m}^{max}$};

    \draw[black!70, font=\footnotesize] (0,0) -- (-0.25,0);
    \draw[black!70, font=\footnotesize] (0,1) -- (-0.25,1);
    \draw[black!70, font=\footnotesize] (0,2) -- (-0.25,2)  node[anchor=east] {$x_{c_1}^{min}$};
    \draw[black!70, font=\footnotesize] (0,3) -- (-0.25,3)  node[anchor=east] {$x_{c_1}^{max}$};
    \draw[black!70, font=\footnotesize] (0,4) -- (-0.25,4)  node[anchor=east] {$x_{c_2}^{max}$};
    \draw[black!70, font=\footnotesize] (0,5) -- (-0.25,5)  node[anchor=east] {$x_{c_3}^{max}$};
    \draw[black!70, font=\footnotesize] (0,5+\h) -- (-0.25,5+\h)  node[anchor=east] {$x_{c_m}^{min}$};
    \draw[black!70, font=\footnotesize] (0,6+\h) -- (-0.25,6+\h)  node[anchor=east] {$x_{c_m}^{max}$};

    \filldraw[black] (2.5, 3) circle (2pt);
    \filldraw[black] (2.5, 4) circle (2pt);
    \filldraw[black] (2.5, 5) circle (2pt);
    \filldraw[black] (2.5, 5+\h) circle (2pt);

    \draw[thick] (3,2.5) -- (3.25,2.5)  node[anchor=west,fill=white] {$c_1 , t_1$};
    \draw[thick] (3,3.5) -- (3.25,3.5)  node[anchor=west,fill=white] {$c_2 , t_2$};
    \draw[thick] (3,4.5) -- (3.25,4.5)  node[anchor=west,fill=white] {$c_3 , t_3$};
    \draw[thick] (3,5.5+\h) -- (3.25,5.5+\h)  node[anchor=west,fill=white] {$c_m , t_m$};
    \foreach \i in {0,...,4}{
        \node at (\i+.5,5.4) {$\vdots$};
        \node at (4.7+\h,\i+0.5) {$\cdots$};
    }
    \node at (4.7+\h,5.5+\h) {$\cdots$};
    \node at (4.7+\h,6.5+\h) {$\cdots$};
    \node at (5.5+\h,5.4) {$\vdots$};
    \node at (6.5+\h,5.4) {$\vdots$};

    \foreach \i in {4,...,6}{
        \draw[-Triangle, shift={(1.9,\i-1.3)}, thick] (0, 0) .. controls(-0.35,0) and (-0.35,0.25) .. (0, 0.5);
    }
    \draw[-Triangle, shift={(1.9,5.5)}, thick] (0, 0) .. controls(-0.35,0) and (-0.35,0.25) .. (0, 0.5);
\end{tikzpicture}
    }
    \caption{Iterative process of integration: starting at the initial cell $c_1$ at integration time $t_1=1$, several cells are crossed and the process finishes at cell $c_m$ at integration time $t_m$.}
    \label{fig:integration_process_grid}
    \end{center}
    \vskip -0.2in
\end{figure}
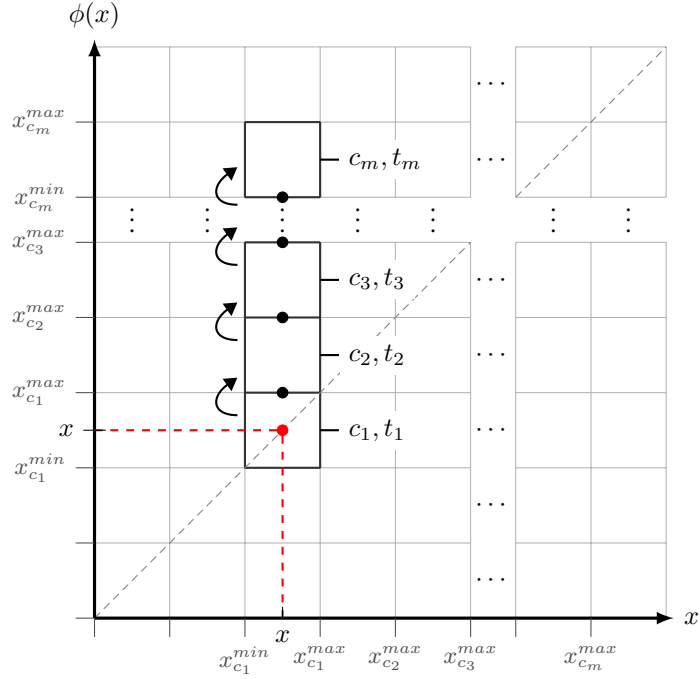

\clearpage
\section{Closed-Form Transformation when Slope $a_c=0$}\label{apx:closed_form_slope}

This special case ($a_{c}=0$ leads to some indeterminations in the previous expressions for the integral (\cref{eq:closed_form_integration}) and derivative (\cref{eq:derivative_complete}), consequently in this section we analyze and provide specific solutions.
\textbf{Note:} infinitesimal equivalents of the exponential and natural logarithm functions: $e^x - 1 \sim x$, if  $x \rightarrow 0$, and $\log x \sim x-1$, if  $x \rightarrow 1$.

\subsection{Integration when Slope $a_c=0$}\label{apx:closed_form_slope_1}
Recall the expression for $\psi(x,t)$ from \cref{sec:cpa_diffeomorphic_transformations}:
\begin{equation}\label{eq:psi_init_azero}
\psi^\theta(x,t) = x e^{t a_c} + \Big(e^{t a_c}-1\Big) \frac{b_c}{a_c}
\end{equation}
Estimate  the limit of this expression when $a_c$ tends to zero:
\begin{equation}\label{eq:psi_limit_azero}
\lim_{a_c \to 0} \psi^\theta(x,t) = 
\lim_{a_c \to 0} x e^{t a_c} + \Big(e^{t a_c}-1\Big) \frac{b_c}{a_c} \simeq  
\lim_{a_c \to 0} x e^{0} + \lim_{a_c \to 0} \Big(t \, a_c\Big) \frac{b_c}{a_c} = x + t \, b_c
\end{equation}
\begin{equation}\label{eq:psi_final_azero}
\boxed{\psi^\theta(x,t) \Big|_{a_c \to 0} = x + t \, b_c}
\end{equation}
Also, for the gradient computation it will be convenient to derive the expression for the hitting time of the boundary of $U_c$:
\begin{equation}\label{eq:thit_init_azero}
t_{hit}^\theta = \frac{1}{a_c} \log \bigg( \frac{a_c x_c + b_c}{a_c x + b_c} \bigg)
\end{equation}
\begin{equation}\label{eq:thit_limit_azero}
\lim_{a_c \to 0} t_{hit}^\theta = 
\lim_{a_c \to 0} \frac{1}{a_c} \log \bigg( \frac{a_c x_c + b_c}{a_c x + b_c} \bigg)  \simeq 
\lim_{a_c \to 0} \frac{1}{a_c}  \bigg( \frac{a_c x_c + b_c}{a_c x + b_c} - 1 \bigg) = \frac{x_c - x}{b_c}
\end{equation}
\begin{equation}\label{eq:thit_final_azero}
\boxed{t_{hit}^\theta \Big|_{a_c \to 0} = \frac{x_c - x}{b_c}}
\end{equation}

\subsection{Derivative when Slope $a_c=0$}\label{apx:closed_form_slope_2}

Note that the partial derivative w.r.t. one of the coefficients of $\theta$, i.e., $\theta_k$ is expressed as:
\begin{equation}\label{eq:psi_derivative_init_azero}
\frac{\partial \phi^\theta(x,t)}{\partial \theta_k} = 
\bigg(
\frac{\partial \psi^\theta(x,t)}{\partial \theta_k} + 
\frac{\partial \psi^\theta(x,t)}{\partial t^\theta} \cdot
\frac{\partial t^\theta}{\partial \theta_k} + 
\frac{\partial \psi^\theta(x,t)}{\partial x} \cdot
\frac{\partial x}{\partial \theta_k}
\bigg)_{\substack{x = x_m \\ t = t_m}}
\end{equation}

To particularize this derivative to the special case $a_c=0$, we first introduce the expressions obtained in \cref{apx:closed_form_derivatives} and then apply the limit when $a_c$ tends to zero. 

\subsubsection{Expression for $\cfrac{\partial \psi^\theta(x,t)}{\partial \theta_k}$ when $a_c=0$}
\label{apx:closed_form_slope_2a}

\begin{equation}\label{eq:psi_derivative_theta_init_azero}
\frac{\partial \psi^\theta(x,t)}{\partial \theta_k} = 
a_c^{(k)} \, t \, e^{t a_c} \Big(x + \frac{b_c}{a_c} \Big) + \Big(e^{t a_c}-1\Big)\frac{b_c^{(k)} \, a_c - a_c^{(k)} \, b_c}{a_c^2}
\end{equation}
\begin{equation}\label{eq:psi_derivative_theta_limit_azero}
\begin{split}
\lim_{a_c \to 0} \frac{\partial \psi^\theta(x,t)}{\partial \theta_k} &= 
\lim_{a_c \to 0} \Bigg(a_c^{(k)} \, t \, e^{t a_c} \Big(x + \frac{b_c}{a_c} \Big) + \Big(e^{t a_c}-1\Big)\frac{b_c^{(k)} \, a_c - a_c^{(k)} \, b_c}{a_c^2}
\Bigg) = \\
&=
a_c^{(k)} \, t \, x + 
\lim_{a_c \to 0} \bigg(a_c^{(k)} \, t \, \frac{b_c}{a_c} \bigg) +
\lim_{a_c \to 0} \bigg(t a_c \frac{b_c^{(k)} \, a_c - a_c^{(k)} \, b_c}{a_c^2}\bigg) = \\
&= 
a_c^{(k)} \, t \, x + 
\lim_{a_c \to 0} \bigg(
a_c^{(k)} \, t \, \frac{b_c}{a_c} +
t b_c^{(k)} - 
a_c^{(k)} \, t \, \frac{b_c}{a_c}
\bigg) = \\
&=
t \Big( a_c^{(k)} x + b_c^{(k)} \Big)
\end{split}
\end{equation}

\subsubsection{Expression for $\cfrac{\partial \psi^\theta(x,t)}{\partial t^\theta}$ when $a_c=0$}
\label{apx:closed_form_slope_2b}

If the partial derivative of $\psi^\theta(x,t)$ w.r.t $t^\theta$ is given by:
\begin{equation}\label{eq:psi_derivative_t_init_azero}
\frac{\partial \psi^\theta(x,t)}{\partial t^\theta} = 
e^{t a_c} \Big( a_c x + b_c \Big)
\end{equation}
then,
\begin{equation}\label{eq:psi_derivative_t_limit_azero}
\lim_{a_c \to 0} \frac{\partial \psi^\theta(x,t)}{\partial t^\theta} = 
\lim_{a_c \to 0} e^{t a_c} \Big( a_c x + b_c \Big) = b_c
\end{equation}

\subsubsection{Expression for $\cfrac{\partial t^\theta}{\partial \theta_k}$ when $a_c=0$}
\label{apx:closed_form_slope_2c}

Recall that the partial derivative of the $t^\theta$ w.r.t $\theta_k$ depends on the previously hit boundaries:
\begin{equation}\label{eq:t_derivative_theta_init_azero}
\frac{\partial t^\theta}{\partial \theta_k} = 
-\sum_{i=1}^{m-1} \frac{\partial t_{hit}^\theta(c_i, x_i)}{\partial \theta_k}
\end{equation}
\begin{equation}\label{eq:thit_derivative_theta_init_azero}
\frac{\partial t_{hit}^\theta(c, x)}{\partial \theta_k} = 
-\frac{a_c^{(k)}}{a_c^2} \log \bigg( \frac{a_c x_c + b_c}{a_c x + b_c} \bigg) +
 \frac{(x - x_c)(b_c^{(k)} a_c - a_c^{(k)} b_c)}{a_c(a_c x + b_c)(a_c x_c + b_c)}
\end{equation}
Therefore,
\begin{equation}\label{eq:thit_derivative_theta_limit_azero}
\begin{split}
\lim_{a_c \to 0} \frac{\partial t_{hit}^\theta(c, x)}{\partial \theta_k} &= 
\lim_{a_c \to 0} \Bigg(
    -\frac{a_c^{(k)}}{a_c^2} \log \bigg( \frac{a_c x_c + b_c}{a_c x + b_c} \bigg) +
    \frac{(x - x_c)(b_c^{(k)} a_c - a_c^{(k)} b_c)}{a_c(a_c x + b_c)(a_c x_c + b_c)}
\Bigg) =\\
&=
\lim_{a_c \to 0} \Bigg(
    -\frac{a_c^{(k)}}{a_c^2} \bigg( \frac{a_c x_c + b_c}{a_c x + b_c} - 1 \bigg) +
    \frac{(x - x_c)(b_c^{(k)} a_c - a_c^{(k)} b_c)}{a_c(a_c x + b_c)(a_c x_c + b_c)}
\Bigg) = \\
&=
\lim_{a_c \to 0} \Bigg(
    -\frac{a_c^{(k)}}{a_c} \bigg( \frac{x_c - x}{\underbrace{a_c x + b_c}_{\to b_c}} \bigg) +
    \frac{(x - x_c)(b_c^{(k)} a_c - a_c^{(k)} b_c)}{a_c(\underbrace{a_c x + b_c}_{\to b_c})(\underbrace{a_c x_c + b_c}_{\to b_c})}
\Bigg) = \\
&=
\lim_{a_c \to 0} \Bigg(
    -\frac{a_c^{(k)} (x_c - x) }{a_c \, b_c} +
    \frac{(x - x_c)(b_c^{(k)} a_c - a_c^{(k)} b_c)}{a_c \, b_c^2}
\Bigg) = \\
&=
\lim_{a_c \to 0} \Bigg(
    -\frac{a_c^{(k)} (x_c - x) }{a_c \, b_c} +
    \frac{b_c^{(k)}(x - x_c)  }{b_c^2} -
    \frac{a_c^{(k)}(x - x_c)  }{a_c \, b_c} 
\Bigg) = \\
&=
\frac{b_c^{(k)}(x - x_c)  }{b_c^2}
\end{split}
\end{equation}
\begin{equation}
\frac{\partial t^\theta}{\partial \theta_k} \Bigg|_{a_c \to 0} = 
-\sum_{i=1}^{m-1} \frac{\partial t_{hit}^\theta(c_i, x_i)}{\partial \theta_k} \Bigg|_{a_c \to 0} = 
-\sum_{i=1}^{m-1} \frac{b_{c_i}^{(k)}(x_i - x_{c_i})  }{b_{c_i}^2}
\end{equation}

\subsubsection{Final Expression for $\cfrac{\partial \phi^\theta(x,t)}{\partial \theta_k}$ when $a_c=0$}\label{apx:closed_form_slope_2d}

Joining all the terms together and evaluating the derivative at $x = x_m$ and $t = t_m$ yields:
\begin{equation}\label{eq:phi_derivative_theta_final_azero}
\boxed{
\begin{aligned}
\frac{\partial \phi^\theta(x,t)}{\partial \theta_k} \Bigg|_{a_c \to 0} &= 
\Bigg(
    t \Big( a_c^{(k)} x + b_c^{(k)} \Big) -
    b_c \sum_{i=1}^{m-1} \frac{b_{c_i}^{(k)}(x_i - x_{c_i})  }{b_{c_i}^2}
\Bigg)_{\substack{x = x_m \\ t = t_m}} = \\
&=
t_m \Big( a_{c_m}^{(k)} x_m + b_{c_m}^{(k)} \Big) -
b_{c_m} \sum_{i=1}^{m-1} \frac{b_{c_i}^{(k)}(x_i - x_{c_i})  }{b_{c_i}^2}
\end{aligned}
}
\end{equation}

\clearpage
\section{Linear Interpolation Grid}\label{apx:linear_interpolation_grid}

Let $f$ be a discretized one-dimensional function defined by two arrays, i.e the input
$x = \{x_{1}, \cdots, x_{n}\}$, and the output $y = \{y_{1}, \cdots, y_{n}\}$, such that $x_{1} < x_{2} < \cdots < x_{n}$. The differentiable time series sampler presented in the Temporal Transformer Network (TTN) is in charge of interpolating output values based on input values, i.e. get an estimate of the function value at $x = \hat{x}$ using piecewise linear interpolation $\hat{y} = f(\hat{x})$. 

For a general irregular grid $x$, $\hat{y}$ is defined as follows:
\begin{equation}\label{eq:grid_y}
\hat{y} = \sum_{i=1}^{n} y_i \cdot \phi_{i}(x)
\end{equation}
where $\phi_{i}(x)$ are known as the basis (hat functions), denoted as:
\begin{equation}\label{eq:grid_basis}
\phi_{i}(x) = 
\left\{\begin{matrix}
    \cfrac{\hat{x} - x_{i-1}}{x_{i} - x_{i-1}} & & x_{i-1} < \hat{x} < x_{i}
    \\
    \cfrac{x_{i+1} - \hat{x}}{x_{i+1} - x_{i}}  & & x_{i} < \hat{x} < x_{i+1}
    \\
    0 & &  \text{otherwise}
\end{matrix}\right.
\end{equation}
for $i=2,\cdots,n-1$, with the boundary "half-hat" basis $\phi_{1}(x)$ and $\phi_{n}(x)$.

In a regular grid $x$ every pair of values are separated by the same distance $h=x_{i+1}-x_{i}=x_{i}-x_{i-1}$. In such case, the above expression for $\phi_{i}(x)$ simplifies to: 
\begin{equation}\label{eq:grid_basis_regular}
\phi_{i}(x) = 
\left\{\begin{matrix}
    (\hat{x} - x_{i-1})/h & & x_{i-1} < \hat{x} < x_{i}
    \\
    (x_{i+1} - \hat{x})/h  & & x_{i} < \hat{x} < x_{i+1}
    \\
    0 & &  \text{otherwise}
\end{matrix}\right.
\end{equation}

The piecewise linear interpolation method is differentiable, which is a necessary condition to propagate the loss to the rest of the model. Thus, let's compute the derivative of the piecewise linear interpolation function with respect to $y_{i}$, $\hat{x}$ and $x_{i}$:

\begin{equation}\label{eq:grid_y_derivative_yi}
\cfrac{\partial \hat{y}}{\partial y_{i}}=\phi_{i}(x)
\end{equation}
\begin{equation}\label{eq:grid_y_derivative_x}
\cfrac{\partial \hat{y}}{\partial \hat{x}}=\sum_{i=1}^{n} y_{i}(x)
\left\{\begin{matrix}
    1/(x_{i} - x_{i-1}) & & x_{i-1} < \hat{x} < x_{i}
    \\
    -1/(x_{i+1} - x_{i})  & & x_{i} < \hat{x} < x_{i+1}
    \\
    0 & &  \text{otherwise}
\end{matrix}\right.
\end{equation}
\begin{equation}\label{eq:grid_y_derivative_xi}
\cfrac{\partial \hat{y}}{\partial x_{i}}=\sum_{i=1}^{n}
\left\{\begin{matrix}
    (y_{i-1}(x) - y_{i}(x))\cfrac{\hat{x} - x_{i-1}}{(x_{i} - x_{i-1})^2}  & & x_{i-1} < \hat{x} < x_{i}
    \\
    (y_{i}(x) - y_{i+1}(x))\cfrac{x_{i+1} - \hat{x}}{(x_{i+1} - x_{i})^2}  & & x_{i} < \hat{x} < x_{i+1}
    \\
    0 & & \text{otherwise}
\end{matrix}\right.
\end{equation}

\paragraph{Self-composition} In order to apply the scaling-squaring approximate method we are interested in a special case of the piecewise linear interpolation in which the input $\hat{x}$ is one of the output values $y_k$, i.e. $\hat(x) = y_{k}$. In such case the derivative with respect to $y_k$ is extended as follows:

\begin{equation}\label{eq:grid_y_derivative_yk}
\cfrac{\partial \hat{y}}{\partial y_{k}}=
\cfrac{\partial \hat{y}}{\partial y_{i}}+
\cfrac{\partial \hat{y}}{\partial \hat{x}}=
\phi_{i}(x) + 
\sum_{i=1}^{n} y_{i}(x)
\left\{\begin{matrix}
    1/(x_{i} - x_{i-1}) & & x_{i-1} < \hat{x} < x_{i}
    \\
    -1/(x_{i+1} - x_{i})  & & x_{i} < \hat{x} < x_{i+1}
    \\
    0 & & \text{otherwise}
\end{matrix}\right.
\end{equation}

\clearpage
\section{Scaling-and-Squaring Integration Method}\label{apx:scaling_squaring}

\begin{figure}[!htb]
    \vskip 0.2in
    \begin{center}
    \subfigure[Forward operation (ODE integration)]{
    \includegraphics[width=0.4\linewidth]{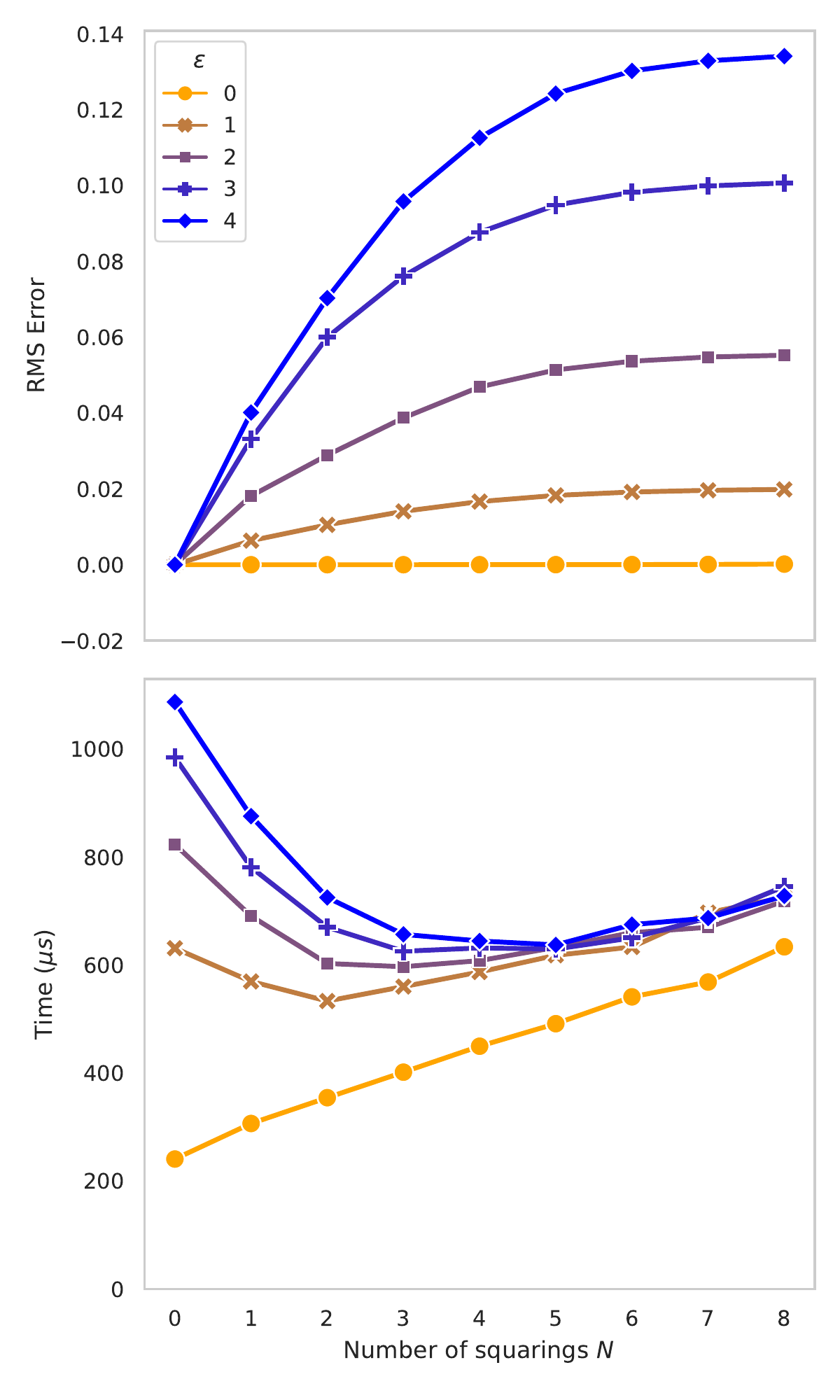}
    \label{fig:scaling_squaring_integration}}
    \hspace{1cm}
    \subfigure[Backward operation (derivative of ODE solution)]{
    \includegraphics[width=0.4\linewidth]{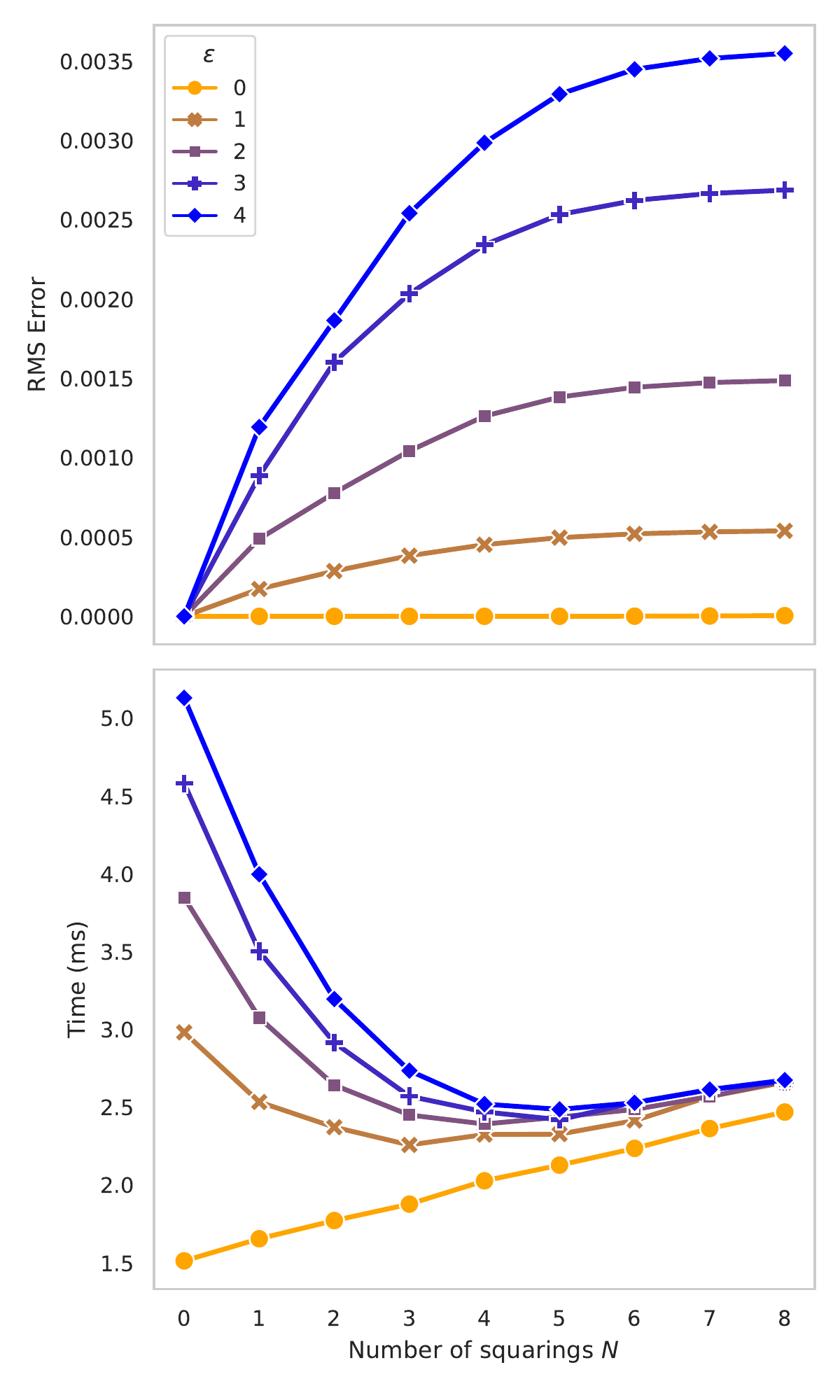}
    \label{fig:scaling_squaring_gradient}}
    \caption{Scaling-and-squaring method: \textbf{Left}: Forward operation (ODE integration).
    \textbf{Right}: Backward operation (derivative of ODE solution). Top row illustrates the RMS error, while the bottom row show the computation time for each calculation.  
    }
    \label{fig:scaling_squaring}
    \end{center}
    \vskip -0.2in
\end{figure}

The scaling-and-squaring method needs to balance two factors to be competitive: the faster integration computation (scaling step) versus the extra time to self-compose the trajectory multiple times (squaring step). \cref{fig:scaling_squaring} shows (\textbf{top}) the integration error (RMS) and (bottom) the computation time for different squaring iterations $N \in [1,8]$. 
Similarly, \cref{fig:scaling_squaring} shows the error and the time for computing the derivative of the ODE solution for different scaling values $N \in [0,8]$. Under this setting, $\epsilon$ represents the parameter vector $\boldsymbol{\theta}=\textbf{1}_{d} \cdot \epsilon$. Thus, a large $\epsilon$ value generates large deformations. 

Executing more squaring iterations (large $N$) yields larger errors (both in the forward and backward operations). 
Even though the computation time increases for small deformations (positive slope for $\epsilon=0$) as the number of squaring iterations increases, experiments show that for large deformations computation speed can be gained at the expense of small integration errors. For example, in the backward operation with $\epsilon=4$, takes around $5$ ms to compute if no scaling-and-squaring method is used. On the contrary, applying the scaling-and-squaring method 8 times (a reduction factor of $2^8=256$) reduces the computation time to around 2.5 ms (50\% improvement), at the expense of 0.0035 RMS error in the gradient value. Therefore, this study shows that the scaling-and-squaring method can boost the speed performance for large deformations.

\clearpage
\section{UCR Archive \cite{dau2019ucr} Dataset List}\label{apx:ucr_archive}

\begin{longtable}{lrrrrl}
\caption{UCR dataset \cite{dau2019ucr} information}\label{tab:ucr_dataset_table}\\
\toprule
                       Dataset &  Train &  Test &  Length &  Classes &      Type \\
\midrule\endfirsthead
\caption{UCR dataset \cite{dau2019ucr} information}\\
\toprule
                       Dataset &  Train &  Test &  Length &  Classes &      Type \\
\midrule \endhead 
                         Adiac &    390 &   391 &     176 &       37 &     Image \\
                     ArrowHead &     36 &   175 &     251 &        3 &     Image \\
                          Beef &     30 &    30 &     470 &        5 &   Spectro \\
                     BeetleFly &     20 &    20 &     512 &        2 &     Image \\
                   BirdChicken &     20 &    20 &     512 &        2 &     Image \\
                           Car &     60 &    60 &     577 &        4 &    Sensor \\
                           CBF &     30 &   900 &     128 &        3 & Simulated \\
         ChlorineConcentration &    467 &  3840 &     166 &        3 & Simulated \\
                  CinCECGTorso &     40 &  1380 &    1639 &        4 &       ECG \\
                        Coffee &     28 &    28 &     286 &        2 &   Spectro \\
                     Computers &    250 &   250 &     720 &        2 &    Device \\
                      CricketX &    390 &   390 &     300 &       12 &    Motion \\
                      CricketY &    390 &   390 &     300 &       12 &    Motion \\
                      CricketZ &    390 &   390 &     300 &       12 &    Motion \\
           DiatomSizeReduction &     16 &   306 &     345 &        4 &     Image \\
  DistalPhalanxOutlineAgeGroup &    400 &   139 &      80 &        3 &     Image \\
   DistalPhalanxOutlineCorrect &    600 &   276 &      80 &        2 &     Image \\
               DistalPhalanxTW &    400 &   139 &      80 &        6 &     Image \\
                   Earthquakes &    322 &   139 &     512 &        2 &    Sensor \\
                        ECG200 &    100 &   100 &      96 &        2 &       ECG \\
                       ECG5000 &    500 &  4500 &     140 &        5 &       ECG \\
                   ECGFiveDays &     23 &   861 &     136 &        2 &       ECG \\
               ElectricDevices &   8926 &  7711 &      96 &        7 &    Device \\
                       FaceAll &    560 &  1690 &     131 &       14 &     Image \\
                      FaceFour &     24 &    88 &     350 &        4 &     Image \\
                      FacesUCR &    200 &  2050 &     131 &       14 &     Image \\
                    FiftyWords &    450 &   455 &     270 &       50 &     Image \\
                          Fish &    175 &   175 &     463 &        7 &     Image \\
                         FordA &   3601 &  1320 &     500 &        2 &    Sensor \\
                         FordB &   3636 &   810 &     500 &        2 &    Sensor \\
                      GunPoint &     50 &   150 &     150 &        2 &    Motion \\
                           Ham &    109 &   105 &     431 &        2 &   Spectro \\
                  HandOutlines &   1000 &   370 &    2709 &        2 &     Image \\
                       Haptics &    155 &   308 &    1092 &        5 &    Motion \\
                       Herring &     64 &    64 &     512 &        2 &     Image \\
                   InlineSkate &    100 &   550 &    1882 &        7 &    Motion \\
           InsectWingbeatSound &    220 &  1980 &     256 &       11 &    Sensor \\
              ItalyPowerDemand &     67 &  1029 &      24 &        2 &    Sensor \\
        LargeKitchenAppliances &    375 &   375 &     720 &        3 &    Device \\
                    Lightning2 &     60 &    61 &     637 &        2 &    Sensor \\
                    Lightning7 &     70 &    73 &     319 &        7 &    Sensor \\
                        Mallat &     55 &  2345 &    1024 &        8 & Simulated \\
                          Meat &     60 &    60 &     448 &        3 &   Spectro \\
                 MedicalImages &    381 &   760 &      99 &       10 &     Image \\
  MiddlePhalanxOutlineAgeGroup &    400 &   154 &      80 &        3 &     Image \\
   MiddlePhalanxOutlineCorrect &    600 &   291 &      80 &        2 &     Image \\
               MiddlePhalanxTW &    399 &   154 &      80 &        6 &     Image \\
                    MoteStrain &     20 &  1252 &      84 &        2 &    Sensor \\
    NonInvasiveFetalECGThorax1 &   1800 &  1965 &     750 &       42 &       ECG \\
    NonInvasiveFetalECGThorax2 &   1800 &  1965 &     750 &       42 &       ECG \\
                      OliveOil &     30 &    30 &     570 &        4 &   Spectro \\
                       OSULeaf &    200 &   242 &     427 &        6 &     Image \\
      PhalangesOutlinesCorrect &   1800 &   858 &      80 &        2 &     Image \\
                       Phoneme &    214 &  1896 &    1024 &       39 &    Sensor \\
                         Plane &    105 &   105 &     144 &        7 &    Sensor \\
ProximalPhalanxOutlineAgeGroup &    400 &   205 &      80 &        3 &     Image \\
 ProximalPhalanxOutlineCorrect &    600 &   291 &      80 &        2 &     Image \\
             ProximalPhalanxTW &    400 &   205 &      80 &        6 &     Image \\
          RefrigerationDevices &    375 &   375 &     720 &        3 &    Device \\
                    ScreenType &    375 &   375 &     720 &        3 &    Device \\
                   ShapeletSim &     20 &   180 &     500 &        2 & Simulated \\
                     ShapesAll &    600 &   600 &     512 &       60 &     Image \\
        SmallKitchenAppliances &    375 &   375 &     720 &        3 &    Device \\
         SonyAIBORobotSurface1 &     20 &   601 &      70 &        2 &    Sensor \\
         SonyAIBORobotSurface2 &     27 &   953 &      65 &        2 &    Sensor \\
                    Strawberry &    613 &   370 &     235 &        2 &   Spectro \\
                   SwedishLeaf &    500 &   625 &     128 &       15 &     Image \\
                       Symbols &     25 &   995 &     398 &        6 &     Image \\
              SyntheticControl &    300 &   300 &      60 &        6 & Simulated \\
              ToeSegmentation1 &     40 &   228 &     277 &        2 &    Motion \\
              ToeSegmentation2 &     36 &   130 &     343 &        2 &    Motion \\
                         Trace &    100 &   100 &     275 &        4 &    Sensor \\
                    TwoLeadECG &     23 &  1139 &      82 &        2 &       ECG \\
                   TwoPatterns &   1000 &  4000 &     128 &        4 & Simulated \\
        UWaveGestureLibraryAll &    896 &  3582 &     945 &        8 &    Motion \\
          UWaveGestureLibraryX &    896 &  3582 &     315 &        8 &    Motion \\
          UWaveGestureLibraryY &    896 &  3582 &     315 &        8 &    Motion \\
          UWaveGestureLibraryZ &    896 &  3582 &     315 &        8 &    Motion \\
                         Wafer &   1000 &  6164 &     152 &        2 &    Sensor \\
                          Wine &     57 &    54 &     234 &        2 &   Spectro \\
                  WordSynonyms &    267 &   638 &     270 &       25 &     Image \\
                         Worms &    181 &    77 &     900 &        5 &    Motion \\
                 WormsTwoClass &    181 &    77 &     900 &        2 &    Motion \\
                          Yoga &    300 &  3000 &     426 &        2 &     Image \\
\bottomrule
\end{longtable}

\clearpage
\section{Additional Alignment Results of Test Data}\label{apx:additional_results}

In this section, we provide more qualitative results of joint alignment of test data in different datasets from the UCR archive \cite{dau2019ucr}.

\begin{figure*}[!htb]
\vskip 0.2in
\begin{center}
    \subfigure[SyntheticControl test set, class 1]{
    \includegraphics[width=0.48\linewidth]{figures/ucr_dataset/alignment/SyntheticControl_aligned_heatmap_test_1.pdf}
    }
    \hfill
    \subfigure[ToeSegmentation1 test set, class 0]{
    \includegraphics[width=0.48\linewidth]{figures/ucr_dataset/alignment/ToeSegmentation1_aligned_heatmap_test_0.pdf}
    }\\
    \subfigure[FacesUCR test set, class 6]{
    \includegraphics[width=0.48\linewidth]{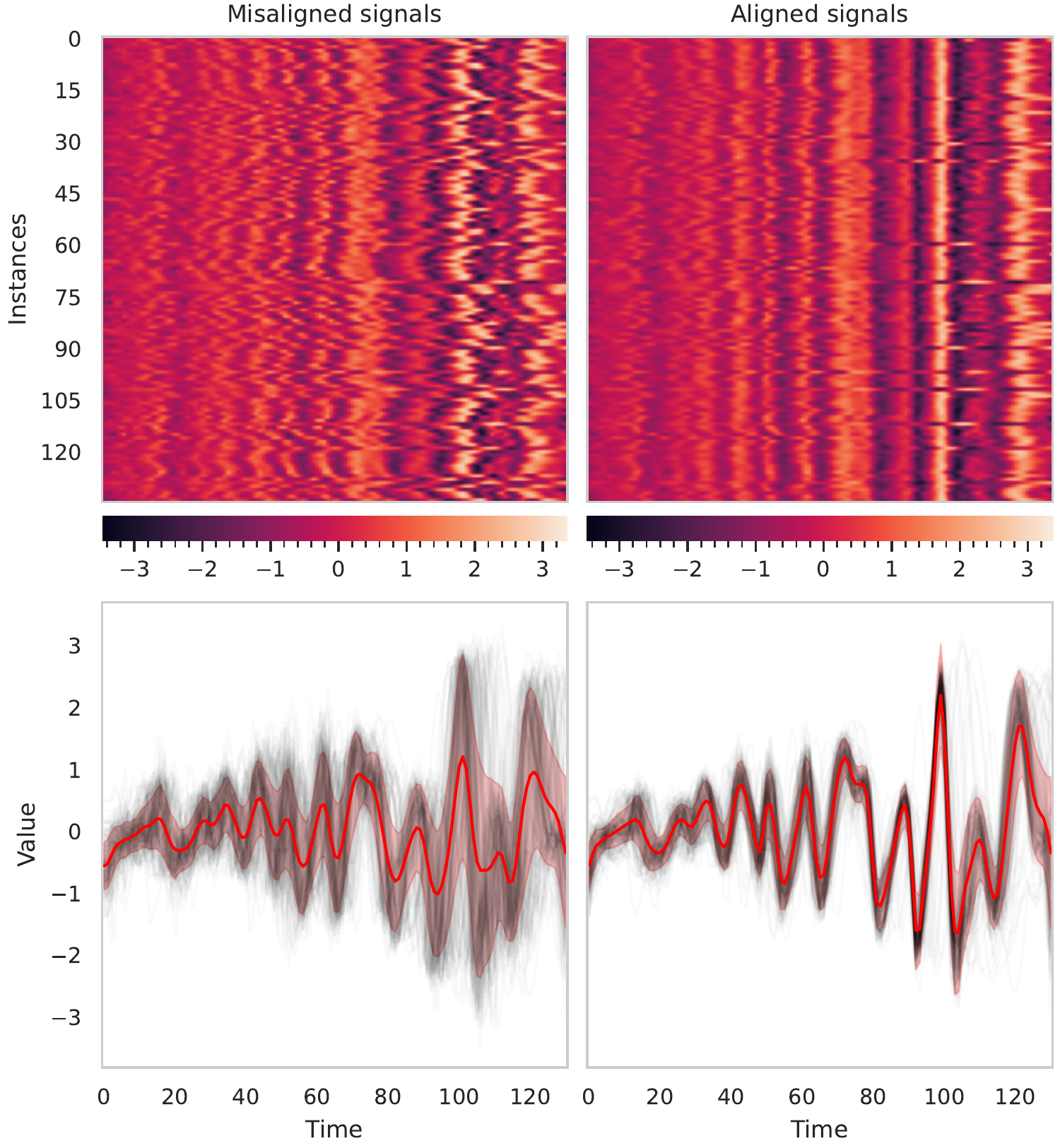}
    }
    \hfill
    \subfigure[SwedishLeaf test set, class 1]{
    \includegraphics[width=0.48\linewidth]{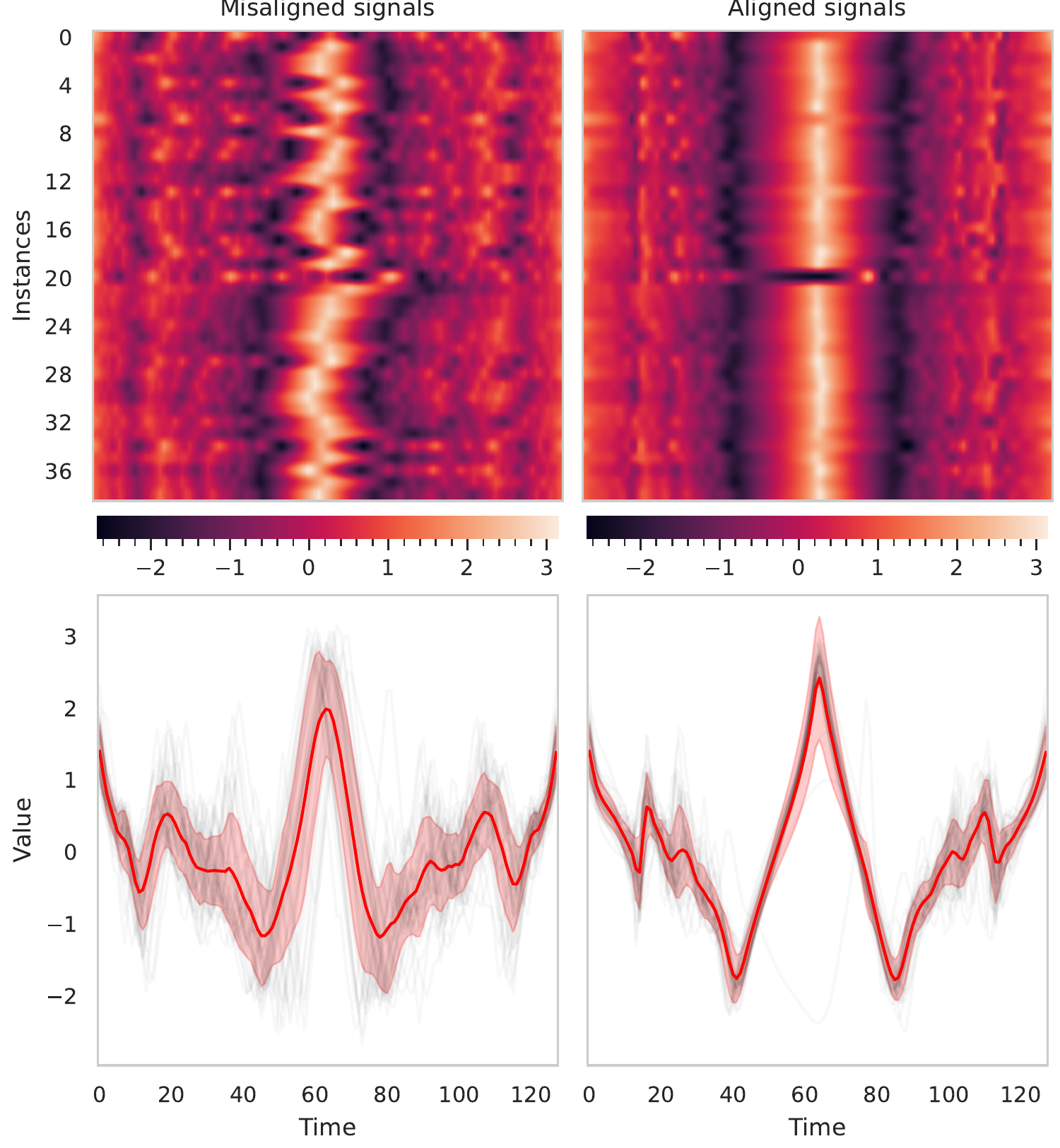}
    }
    \caption{Multi-class time series alignment on multiple datasets. Top: heatmap of each time series sample (row). Bottom: overlapping time series, red line represents Euclidean average. Left: original signals. Right: after alignment.}
    \label{fig:alignment_example_1}
\end{center}
\vskip -0.2in
\end{figure*}

\begin{figure*}[!htb]
\vskip 0.2in
\begin{center}
    \subfigure[Trace test set, class 0]{
    \includegraphics[width=0.48\linewidth]{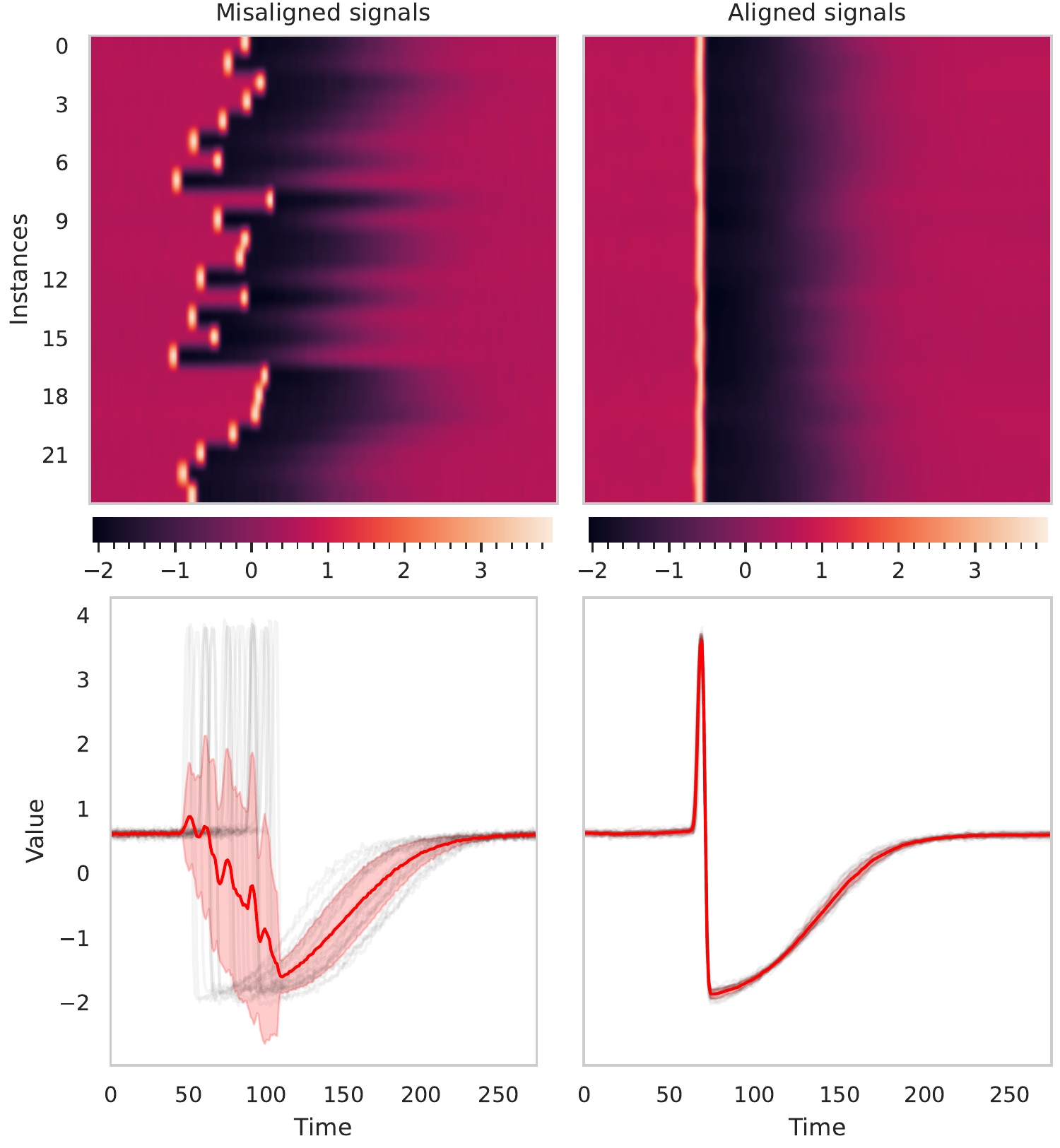}
    }
    \hfill
    \subfigure[Trace test set, class 1]{
    \includegraphics[width=0.48\linewidth]{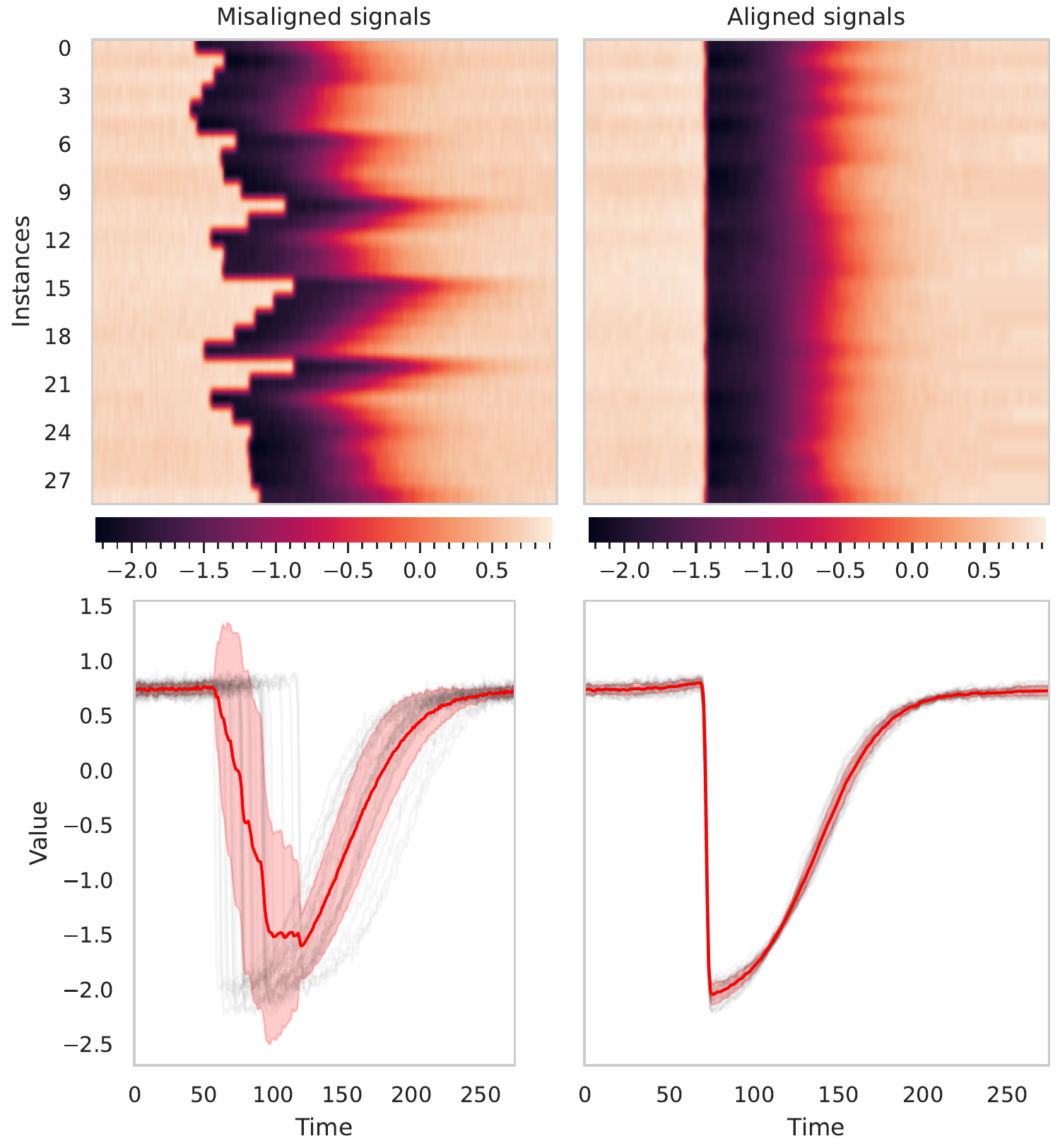}
    }
    \caption{More examples. Multi-class time series alignment on multiple datasets. Top: heatmap of each time series sample (row). Bottom: overlapping time series, red line represents Euclidean average. Left: original signals. Right: after alignment.}
    \label{fig:alignment_example_2}
\end{center}
\vskip -0.2in
\end{figure*}

\begin{figure}[!htb]
    \vskip 0.2in
    \begin{center}
    \includegraphics[width=0.48\linewidth]{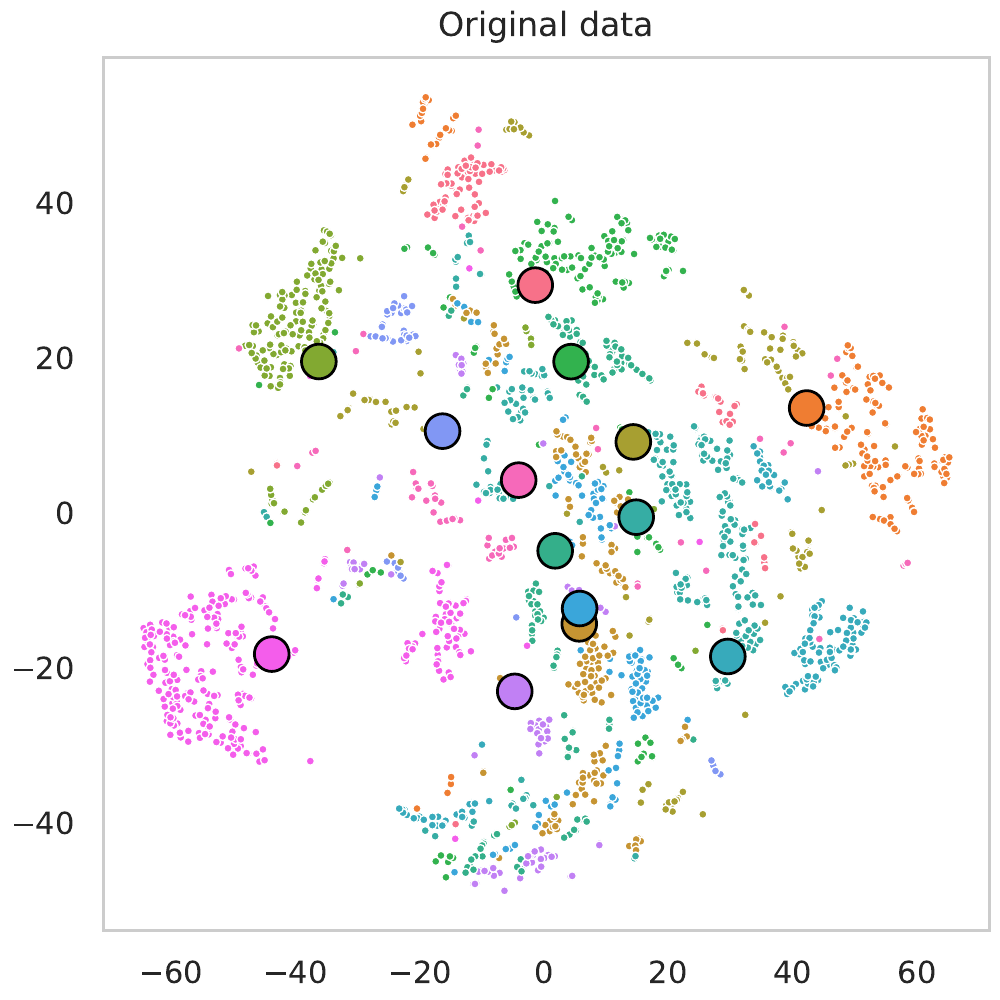}
    \includegraphics[width=0.48\linewidth]{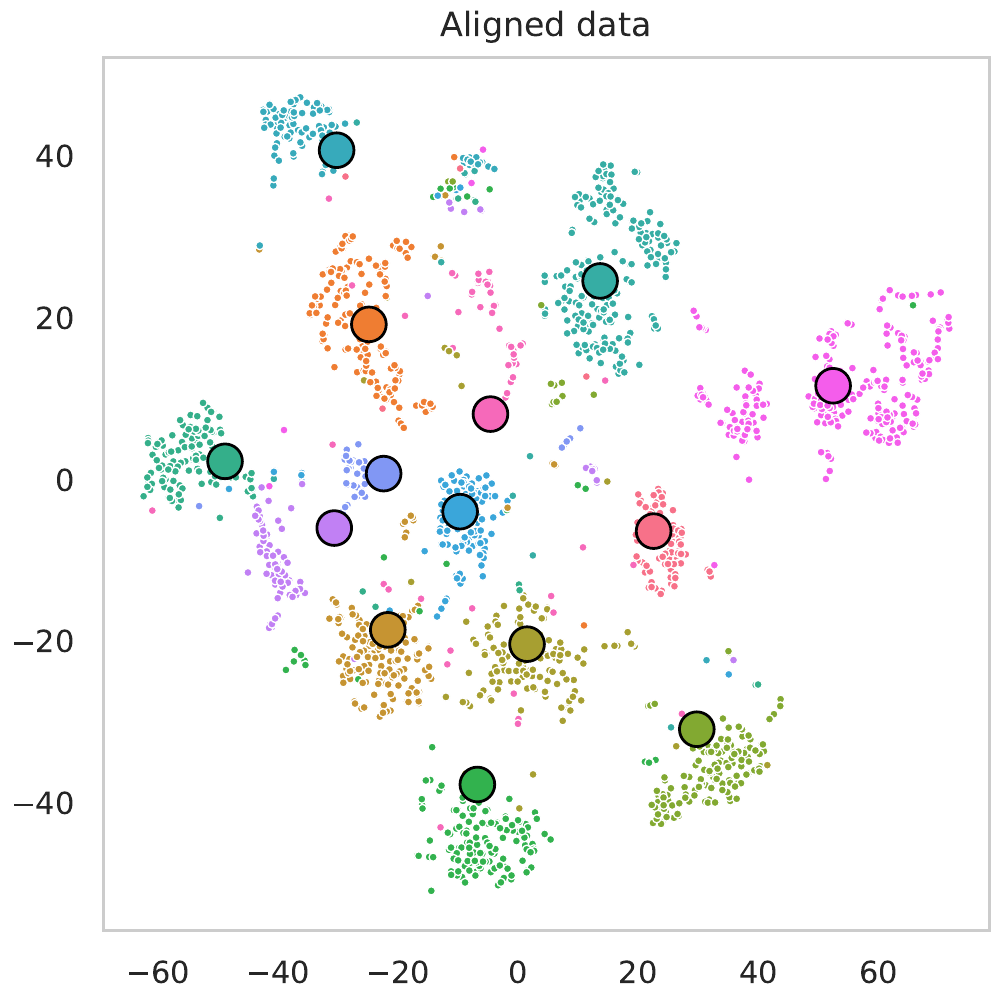}
    \caption{A t-SNE visualization of the FacesUCR dataset from the UCR archive before and after alignment.}
    \label{fig:alignment_tsne}
    \end{center}
    \vskip -0.2in
\end{figure}

\cref{fig:alignment_tsne} shows a t-SNE \cite{van2008visualizing} visualization of the original and aligned data of the 11-class FacesUCR dataset. This illustrates how our TTN indeed decreases intra-class variance while increasing inter-class one, thus improving the performance of classification.

\clearpage

\begin{figure*}[t]
\vskip 0.2in
\begin{center}
    \subfigure[SyntheticControl dataset]{
    \includegraphics[height=0.37\linewidth]{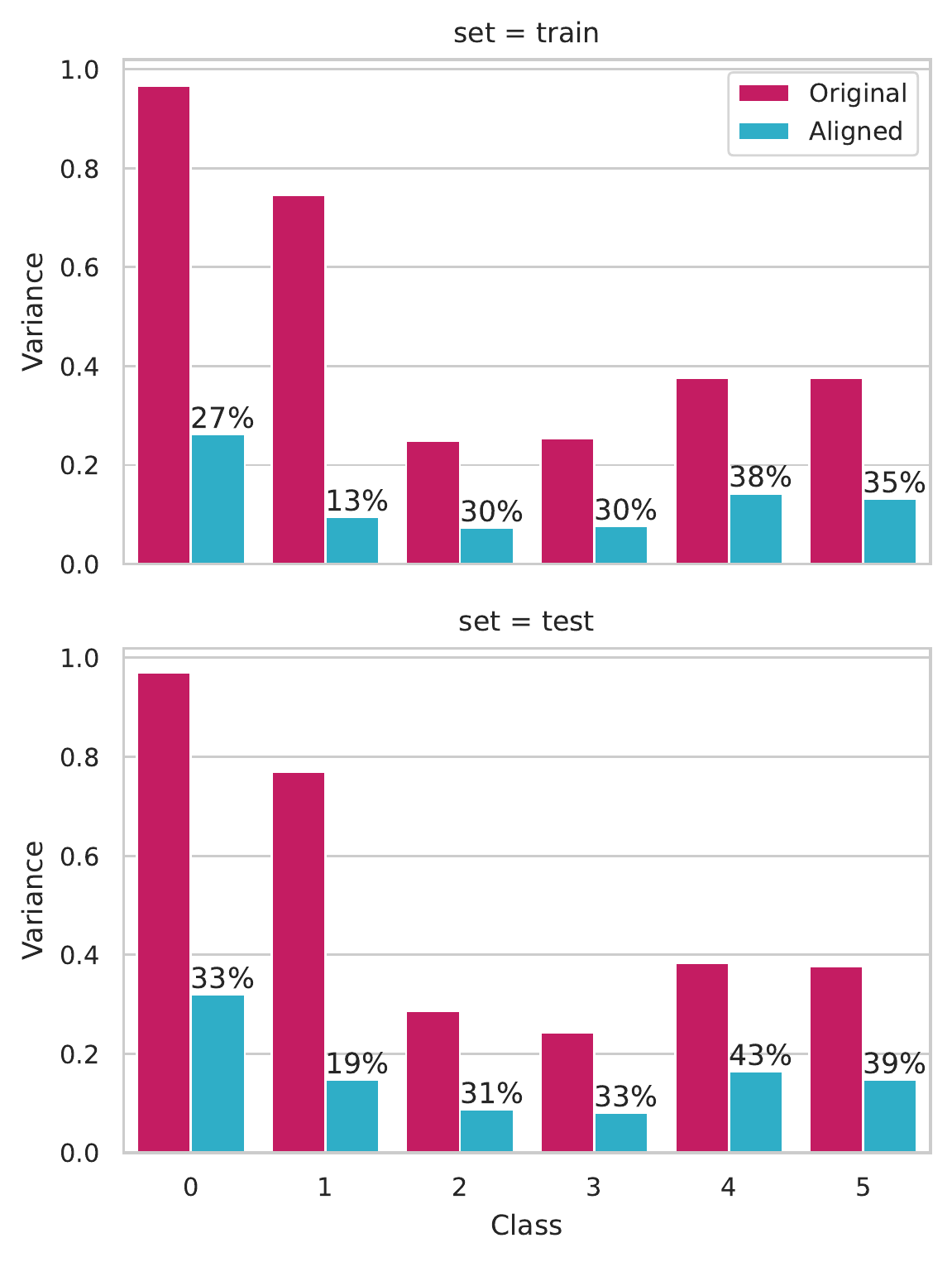}
    }
    \hfill
    \subfigure[FacesUCR dataset]{
    \includegraphics[height=0.37\linewidth]{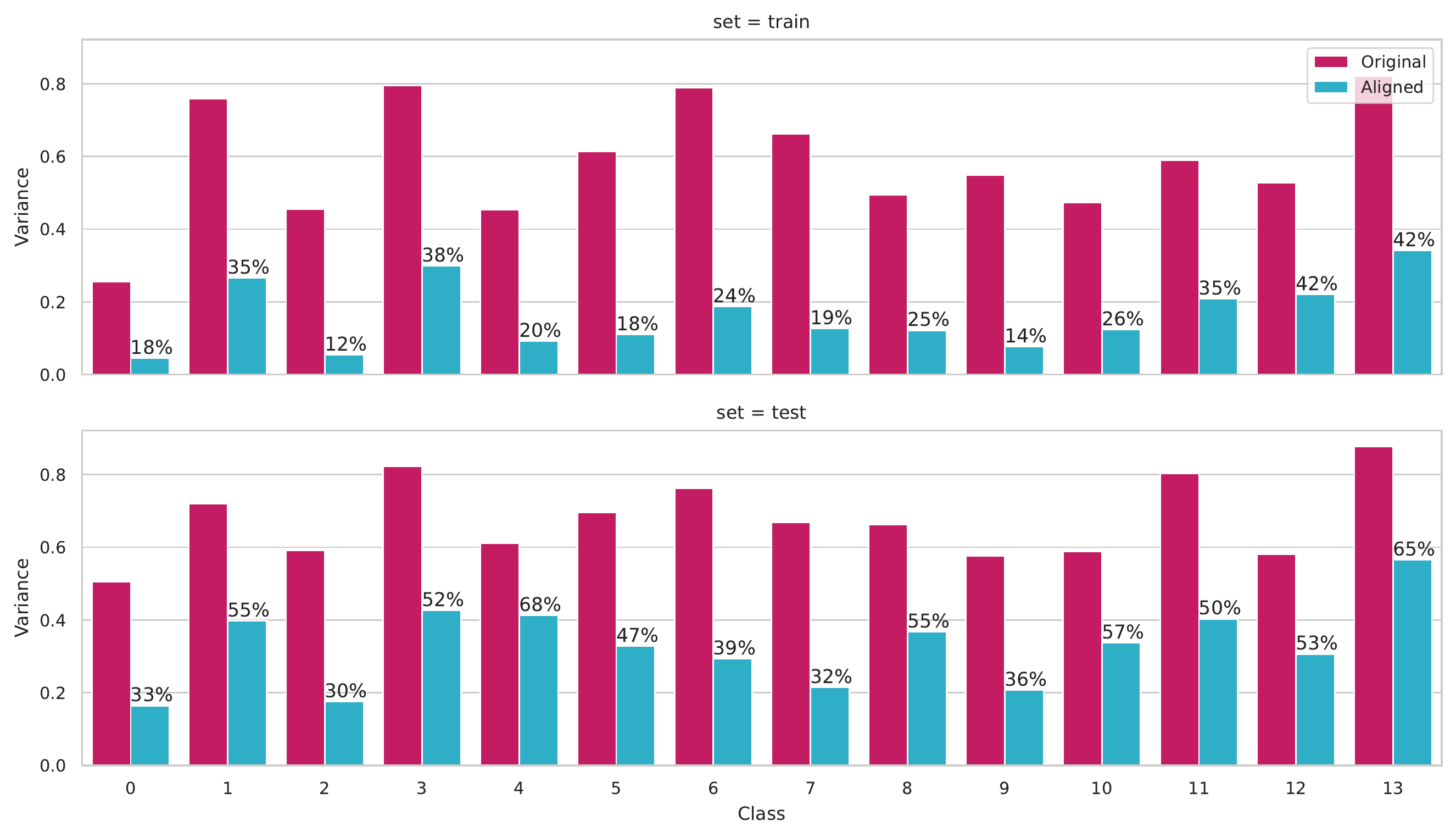}
    }\\
    \subfigure[SyntheticControl dataset]{
    \includegraphics[height=0.32\linewidth]{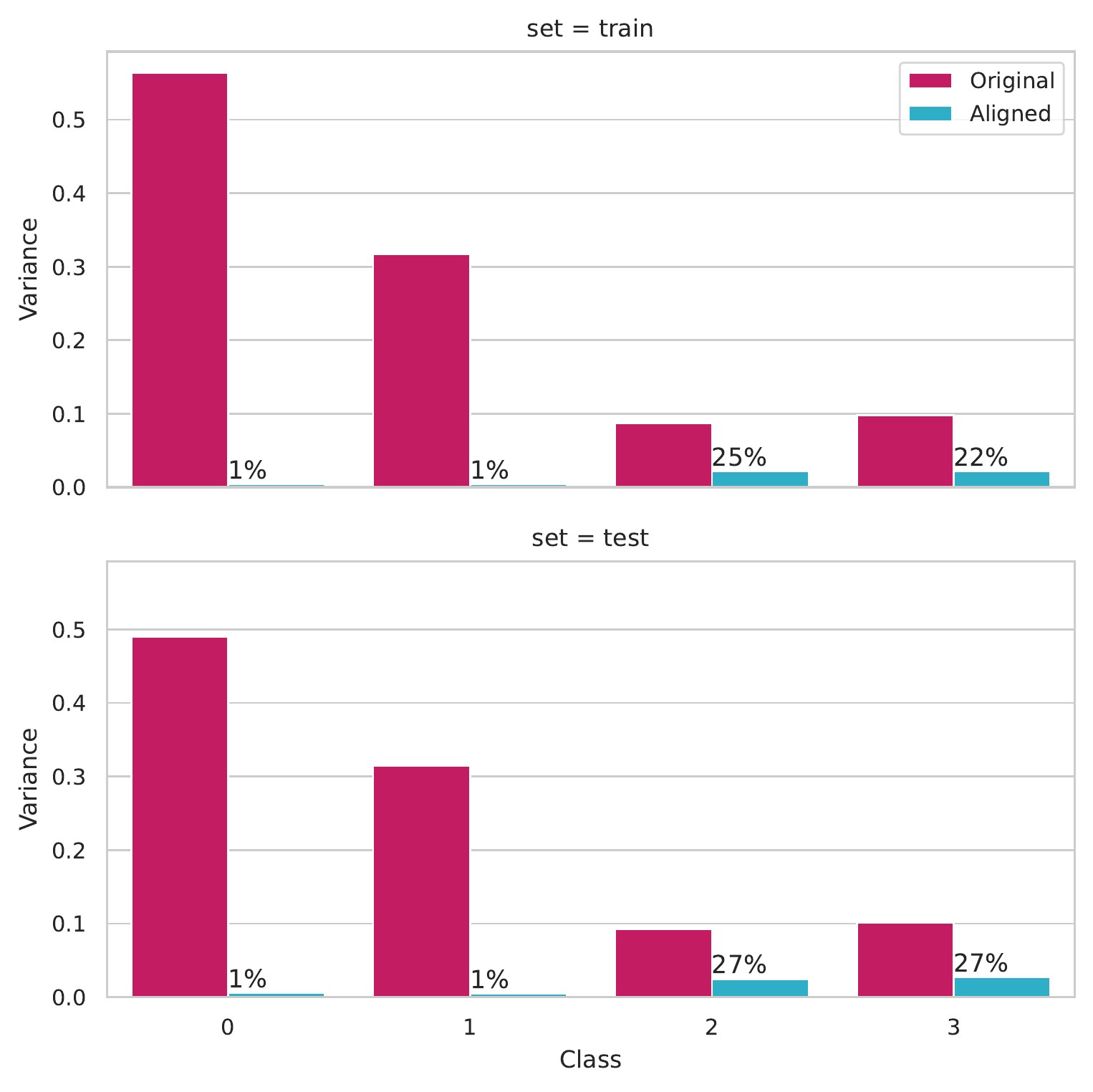}
    }
    \hfill
    \subfigure[SyntheticControl dataset]{
    \includegraphics[height=0.32\linewidth]{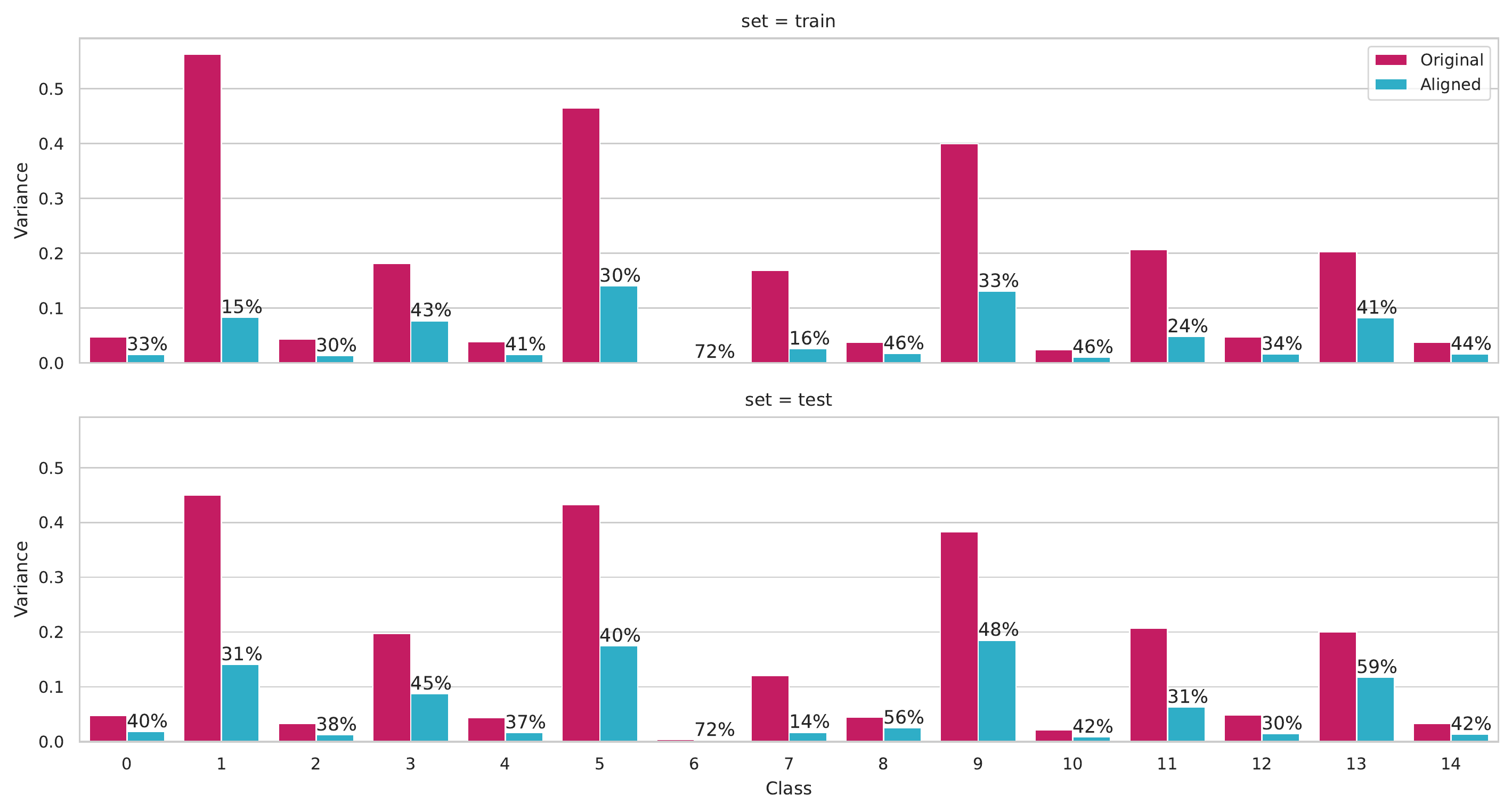}
    }
    \caption{Within-class variances on several UCR archive datasets before and after alignment.}
    \label{fig:alignment_variance}
\end{center}
\vskip -0.2in
\end{figure*}

\clearpage
\section{Nearest Centroid Classification (NCC) Results}\label{apx:ncc_results}

\begin{figure*}[!htb]
    \vskip 0.2in
    \begin{center}
    \centerline{\includegraphics[width=\linewidth]{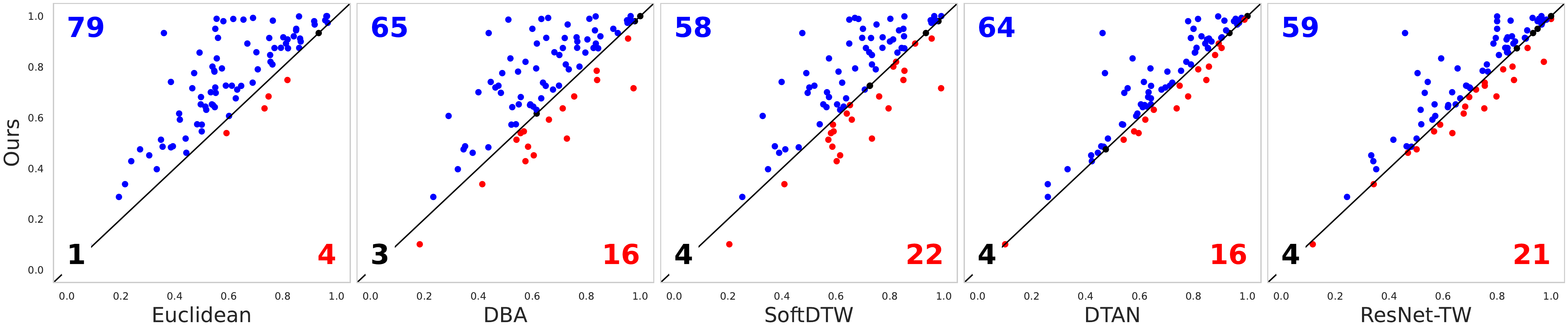}}
    \caption{Correct classification rates using NCC. Each point above the diagonal indicates an entire UCR archive dataset where our model achieves better results than the comparing method. From left to right, our test accuracy compared with: Euclidean (ours wins in 94\% of the datasets), DBA (77\%), SoftDTW (69\%), DTAN (76\%) and ResNet-TW (70\%).}
    \label{fig:ucr_dataset_rates}
    \end{center}
    \vskip -0.2in
\end{figure*}

\begin{figure*}[!htb]
\vskip 0.2in
\begin{center}
    \subfigure[Percentage of change in accuracy]{
    \includegraphics[width=0.48\linewidth]{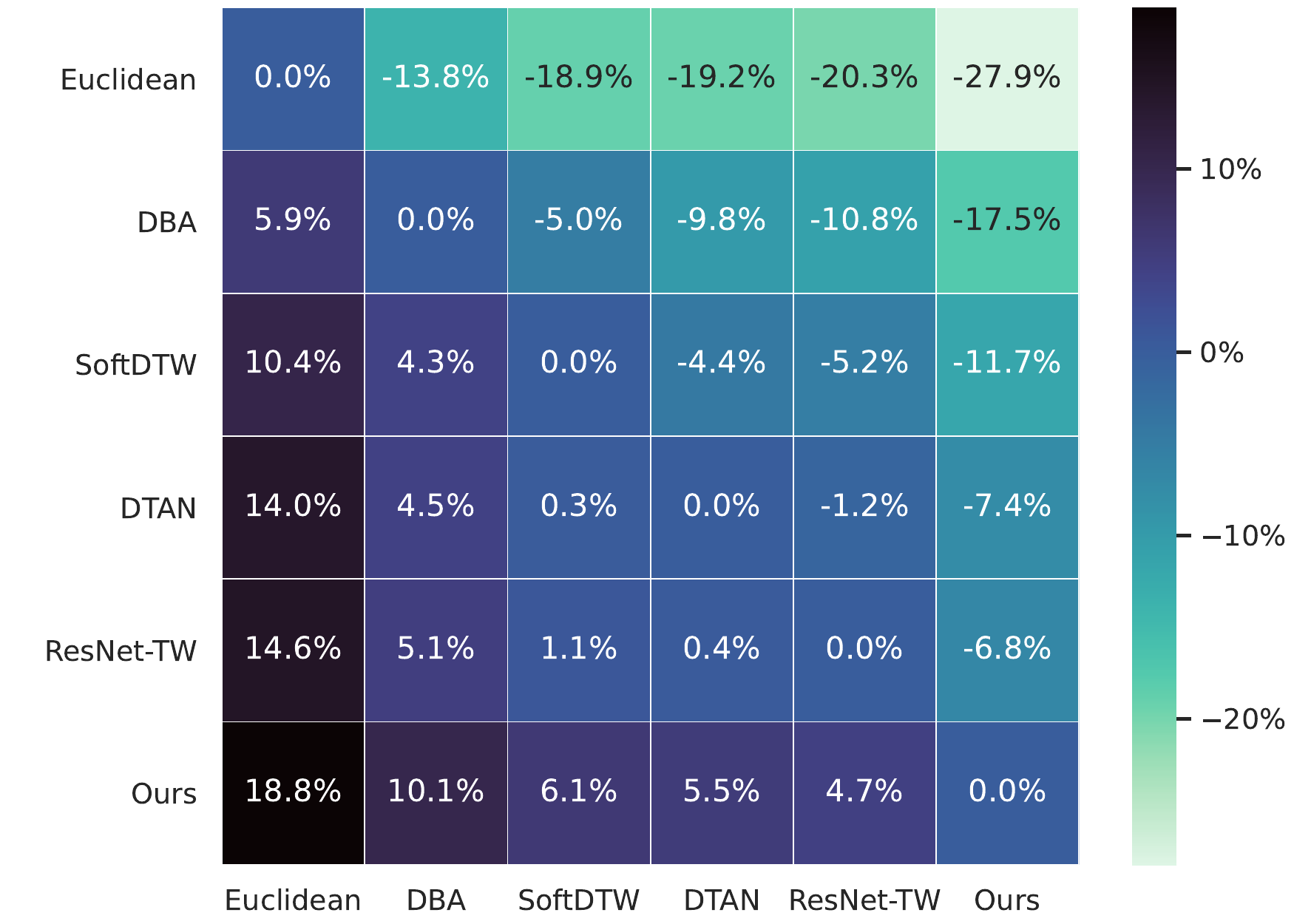}
    \label{fig:ucr_dataset_diff}}
    \hfill
    \subfigure[Percentage of wins]{
    \includegraphics[width=0.48\linewidth]{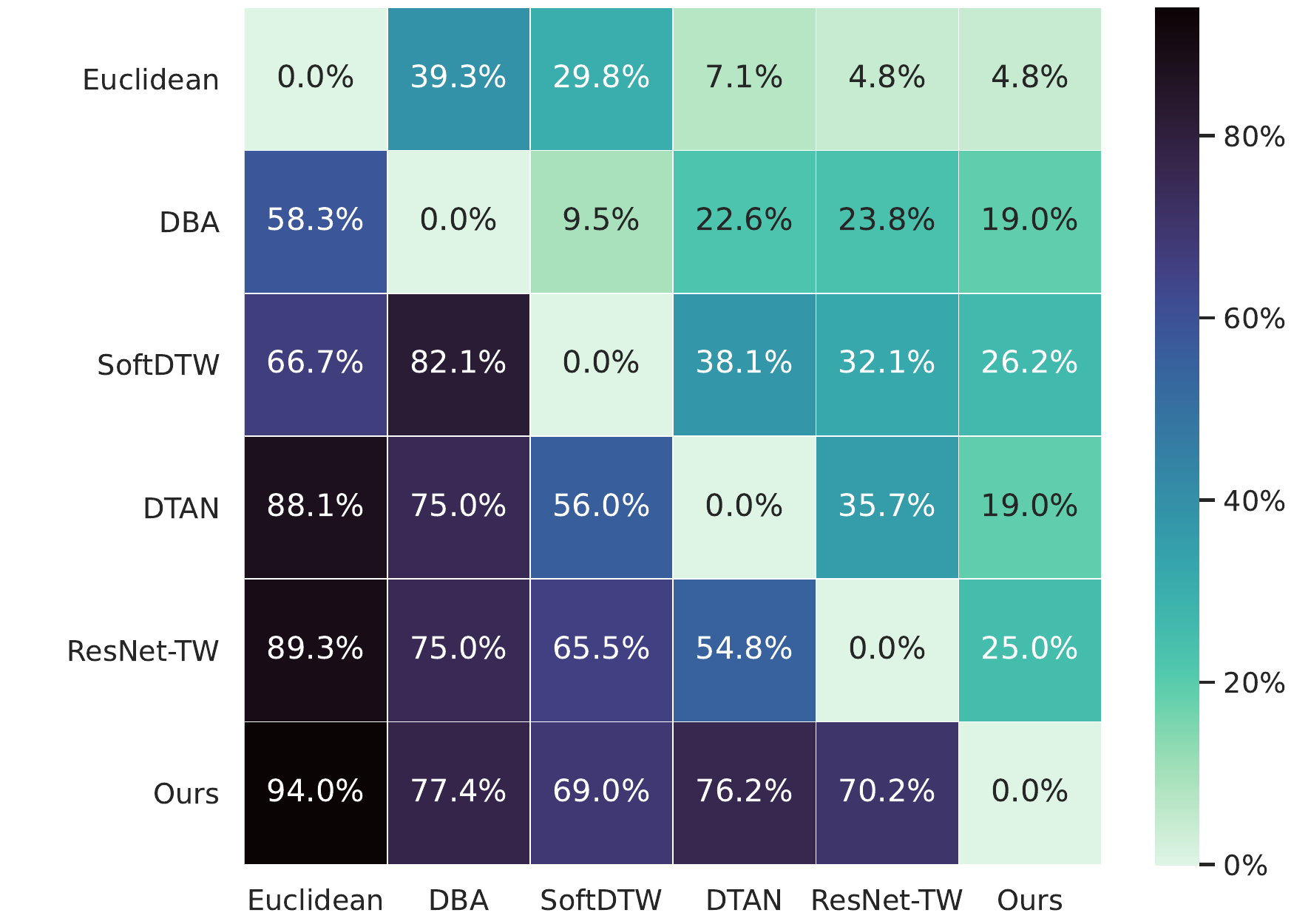}
    \label{fig:ucr_dataset_perc}}
    \caption{Percentage of wins (left) (row against column) and average percentage of change in accuracy (right) over UCR archive.}
    \label{fig:ucr_dataset_percentages}
\end{center}
\vskip -0.2in
\end{figure*}

In this section, we show detailed quantitative results of the NCC experiment on UCR archive \cite{dau2019ucr} in \cref{fig:ucr_dataset_rates,fig:ucr_dataset_percentages,tab:ucr_dataset_by_type,tab:ucr_dataset_accuracy}.
We compare our Temporal Transformer Network with Euclidean average (as a baseline), DTW Barycenter Averaging (DBA) \cite{Petitjean2011-DBA}, SoftDTW \cite{cuturi2017soft}, DTAN \cite{Weber2019} and ResNet-TW \cite{Huang2021}.

\begin{table*}[!htb]
    \caption{Accuracy of nearest centroid classification on UCR dataset \cite{dau2019ucr}, grouped by time series type. Highest values per dataset (row) are shown in bold.}
    \label{tab:ucr_dataset_by_type}
    \vskip 0.15in
    \begin{center}
    \begin{small}
    \begin{tabular}{lcccccc}
        \toprule
         Type (count) &   Euclidean &        DBA &    SoftDTW &       DTAN &  ResNet-TW &       Ours \\
        \midrule
       Device (6) &   0.425781 &   0.584495 &   \textbf{0.591291} &   0.523803 &   0.534700 &   0.540954 \\
          ECG (7) &   0.687460 &   0.706706 &   0.705433 &   0.855727 &   0.814560 &   \textbf{0.918076} \\
       Image (29) &   0.625407 &   0.660084 &   0.686681 &   0.719903 &   0.739120 &   \textbf{0.767902} \\
      Motion (14) &   0.476806 &   0.577721 &   0.603160 &   0.579338 &   0.561593 &   \textbf{0.609039} \\
      Sensor (15) &   0.660346 &   0.651705 &   0.693455 &   0.749147 &   0.772336 &   \textbf{0.783508} \\
    Simulated (6) &   0.657427 &   0.786523 &   \textbf{0.805312} &   0.710171 &   0.734658 &   0.770062 \\
      Spectro (7) &   0.754846 &   0.700941 &   0.745851 &   0.805897 &   0.791203 &   \textbf{0.866869} \\
          \midrule
       Total (84) &   0.610865 &   0.655783 &   0.682124 &   0.705480 &   0.711170 &   \textbf{0.748917} \\
        \bottomrule
    \end{tabular}
    \end{small}
    \end{center}
    \vskip -0.1in
\end{table*}

\clearpage
\begin{longtable}{lcccccc}
\caption{Accuracy of nearest centroid classification on UCR dataset \cite{dau2019ucr}. Highest values per dataset (row) are shown in bold.}\label{tab:ucr_dataset_accuracy}\\
\toprule
Dataset &   Euclidean &        DBA &    SoftDTW &       DTAN &  ResNet-TW &       Ours \\ 
\midrule 
\endfirsthead 
\caption{Accuracy of nearest centroid classification on UCR dataset \cite{dau2019ucr}. Highest values per dataset (row) are shown in bold.}\\
\toprule
Dataset &   Euclidean &        DBA &    SoftDTW &       DTAN &  ResNet-TW &       Ours \\ 
\midrule \endhead 
                         Adiac &   0.549872 &   0.462916 &   0.501279 &   0.695652 &   0.698210 &   \textbf{0.718670} \\ 
                     ArrowHead &   0.611429 &   0.474286 &   0.520000 &   0.748571 &   \textbf{0.754286} &   0.725714 \\ 
                          Beef &   0.533333 &   0.400000 &   0.566667 &   0.633333 &   0.633333 &   \textbf{0.700000} \\ 
                     BeetleFly &   0.850000 &   0.900000 &   0.850000 &   0.800000 &   0.800000 &   \textbf{0.950000} \\ 
                   BirdChicken &   0.550000 &   0.600000 &   0.700000 &   0.800000 &   \textbf{0.950000} &   \textbf{0.950000} \\ 
                           Car &   0.616667 &   0.633333 &   0.683333 &   0.816667 &   \textbf{1.000000} &   0.988889 \\ 
                           CBF &   0.763333 &   0.965556 &   0.971111 &   0.914444 &   0.850000 &   \textbf{0.982222} \\ 
         ChlorineConcentration &   0.333073 &   0.323698 &   0.348177 &   0.333073 &   0.351823 &   \textbf{0.396615} \\ 
                  CinCECGTorso &   0.385507 &   0.445652 &   0.398551 &   0.615942 &   0.542754 &   \textbf{0.740580} \\ 
                        Coffee &   0.964286 &   0.964286 &   0.964286 &   \textbf{1.000000} &   0.964290 &   \textbf{1.000000} \\ 
                     Computers &   0.416000 &   0.616000 &   0.640000 &   0.592000 &   \textbf{0.676000} &   0.616000 \\ 
                      CricketX &   0.238462 &   0.574359 &   \textbf{0.602564} &   0.423077 &   0.341026 &   0.428205 \\ 
                      CricketY &   0.348718 &   0.541026 &   \textbf{0.571795} &   0.541026 &   0.415385 &   0.512821 \\ 
                      CricketZ &   0.305128 &   0.605128 &   \textbf{0.615385} &   0.420513 &   0.333333 &   0.451282 \\ 
           DiatomSizeReduction &   0.957516 &   0.950980 &   0.950980 &   0.970588 &   0.973856 &   \textbf{0.983660} \\ 
  DistalPhalanxOutlineAgeGroup &   0.817500 &   0.840000 &   0.850000 &   0.847500 &   \textbf{0.862500} &   0.748201 \\ 
   DistalPhalanxOutlineCorrect &   0.471667 &   0.488333 &   0.490000 &   0.471667 &   0.505000 &   \textbf{0.775362} \\ 
               DistalPhalanxTW &   0.747500 &   0.755000 &   0.760000 &   0.780000 &   \textbf{0.797500} &   0.683453 \\ 
                   Earthquakes &   0.754658 &   0.574534 &   0.822981 &   0.773292 &   \textbf{0.973287} &   0.820000 \\ 
                        ECG200 &   0.750000 &   0.720000 &   0.730000 &   0.790000 &   0.795031 &   \textbf{0.913778} \\ 
                       ECG5000 &   0.860444 &   0.834667 &   0.853778 &   0.891333 &   0.800000 &   \textbf{0.998839} \\ 
                   ECGFiveDays &   0.689895 &   0.658537 &   0.670151 &   0.977933 &   0.931556 &   \textbf{0.993031} \\ 
               ElectricDevices &   0.482687 &   0.538970 &   0.539748 &   0.534820 &   0.518869 &   \textbf{0.573726} \\ 
                       FaceAll &   0.491716 &   0.796450 &   0.827811 &   0.804734 &   0.840909 &   \textbf{0.856213} \\ 
                      FaceFour &   0.840909 &   0.852273 &   0.852273 &   0.829545 &   0.855122 &   \textbf{0.920455} \\ 
                      FacesUCR &   0.539512 &   0.774634 &   0.812683 &   0.857073 &   \textbf{0.857143} &   0.800976 \\ 
                    FiftyWords &   0.516484 &   0.615385 &   0.615385 &   \textbf{0.652747} &   0.516484 &   0.630769 \\ 
                          Fish &   0.560000 &   0.651429 &   0.697143 &   0.902857 &   0.902857 &   \textbf{0.914286} \\ 
                         FordA &   0.495973 &   0.549570 &   0.552902 &   0.604832 &   0.568176 &   \textbf{0.652273} \\ 
                         FordB &   0.499725 &   0.568482 &   \textbf{0.591309} &   0.579758 &   0.566282 &   0.545679 \\ 
                      GunPoint &   0.753333 &   0.700000 &   0.733333 &   \textbf{0.880000} &   0.806667 &   0.846667 \\ 
                           Ham &   0.761905 &   0.723810 &   0.733333 &   0.790476 &   0.761905 &   \textbf{0.809524} \\ 
                  HandOutlines &   0.818000 &   0.804000 &   0.812000 &   0.850000 &   0.835000 &   \textbf{0.908108} \\ 
                       Haptics &   0.392857 &   0.350649 &   0.373377 &   0.457792 &   0.464286 &   \textbf{0.487013} \\ 
                       Herring &   0.546875 &   0.546875 &   0.609375 &   0.703125 &   0.765625 &   \textbf{0.781250} \\ 
                   InlineSkate &   0.192727 &   0.232727 &   0.252727 &   0.260000 &   0.243636 &   \textbf{0.287273} \\ 
           InsectWingbeatSound &   0.601010 &   0.289394 &   0.328283 &   0.587374 &   0.570707 &   \textbf{0.606566} \\ 
              ItalyPowerDemand &   0.918367 &   0.730807 &   0.750243 &   0.962099 &   0.965015 &   \textbf{0.966958} \\ 
        LargeKitchenAppliances &   0.440000 &   0.728000 &   \textbf{0.733333} &   0.482667 &   0.501333 &   0.517333 \\ 
                    Lightning2 &   0.688525 &   0.639344 &   0.622951 &   0.721311 &   \textbf{0.754098} &   0.737705 \\ 
                    Lightning7 &   0.589041 &   0.698630 &   \textbf{0.726027} &   0.712329 &   0.684932 &   \textbf{0.726027} \\
                        Mallat &   0.966738 &   0.952665 &   0.953945 &   0.968870 &   0.966738 &   \textbf{0.973561} \\ 
                          Meat &   \textbf{0.933333} &   0.916667 &   \textbf{0.933333} &   \textbf{0.933333} &   \textbf{0.933333} &   \textbf{0.933333} \\ 
                 MedicalImages &   0.385526 &   0.436842 &   0.461842 &   0.468421 &   0.473684 &   \textbf{0.482895} \\ 
  MiddlePhalanxOutlineAgeGroup &   0.732500 &   0.712500 &   \textbf{0.795000} &   0.737500 &   0.752500 &   0.636364 \\ 
   MiddlePhalanxOutlineCorrect &   0.551667 &   0.483333 &   0.495000 &   0.543333 &   0.531667 &   \textbf{0.697595} \\ 
               MiddlePhalanxTW &   0.591479 &   0.556391 &   0.581454 &   0.596491 &   \textbf{0.634085} &   0.538961 \\ 
                    MoteStrain &   0.861022 &   0.826677 &   0.843450 &   0.904153 &   \textbf{0.912939} &   0.874601 \\ 
    NonInvasiveFetalECGThorax1 &   0.769466 &   0.712977 &   0.710941 &   0.853435 &   0.838677 &   \textbf{0.874300} \\ 
    NonInvasiveFetalECGThorax2 &   0.802036 &   0.763868 &   0.773028 &   0.905344 &   0.838680 &   \textbf{0.916539} \\ 
                      OliveOil &   0.866667 &   0.766667 &   0.800000 &   0.866667 &   0.866667 &   \textbf{0.900000} \\ 
                       OSULeaf &   0.359504 &   0.438017 &   0.475207 &   0.462810 &   0.458678 &   \textbf{0.933333} \\ 
      PhalangesOutlinesCorrect &   0.625874 &   0.632867 &   0.637529 &   0.642191 &   0.663170 &   \textbf{0.675991} \\ 
                       Phoneme &   0.078586 &   0.182489 &   \textbf{0.204641} &   0.101793 &   0.116561 &   0.100738 \\ 
                         Plane &   0.961905 &   \textbf{1.000000} &   0.990476 &   \textbf{1.000000} &   \textbf{1.000000} &   \textbf{1.000000} \\ 
ProximalPhalanxOutlineAgeGroup &   0.819512 &   0.843902 &   0.853659 &   0.853659 &   \textbf{0.873171} &   \textbf{0.873171} \\ 
 ProximalPhalanxOutlineCorrect &   0.646048 &   0.649485 &   \textbf{0.725086} &   0.642612 &   0.687285 &   \textbf{0.725086} \\
             ProximalPhalanxTW &   0.707500 &   0.735000 &   0.747500 &   0.817500 &   \textbf{0.822500} &   0.790244 \\ 
          RefrigerationDevices &   0.354667 &   0.584000 &   \textbf{0.586667} &   0.466667 &   0.482667 &   0.485333 \\ 
                    ScreenType &   0.442667 &   0.378667 &   0.389333 &   0.445333 &   \textbf{0.469333} &   0.461333 \\ 
                   ShapeletSim &   0.500000 &   0.522222 &   \textbf{0.588889} &   0.538889 &   \textbf{0.588889} &   0.572222 \\ 
                     ShapesAll &   0.513333 &   0.603333 &   0.628333 &   0.628333 &   \textbf{0.681667} & 0.643333 \\
        SmallKitchenAppliances &   0.418667 &   \textbf{0.661333} &   0.658667 &   0.621333 &   0.560000 &   0.592000 \\ 
         SonyAIBORobotSurface1 &   0.811980 &   0.835275 &   \textbf{0.893511} &   \textbf{0.893511} &   0.860233 &   0.891847 \\
         SonyAIBORobotSurface2 &   0.793284 &   0.766002 &   0.772298 &   0.811123 &   0.830010 &   \textbf{0.875131} \\ 
                    Strawberry &   0.668842 &   0.616639 &   0.649266 &   0.843393 &   0.786297 &   \textbf{0.891892} \\ 
                   SwedishLeaf &   0.702400 &   0.681600 &   0.723200 &   0.806400 &   0.836800 &   \textbf{0.857600} \\ 
                       Symbols &   0.864322 &   \textbf{0.95477}4 &   \textbf{0.954774} &   0.857286 &   0.906533 &   0.911558 \\
              SyntheticControl &   0.916667 &   \textbf{0.980000} &   \textbf{0.980000} &   0.950000 &   0.950000 &   \textbf{0.980000} \\ 
              ToeSegmentation1 &   0.574561 &   0.614035 &   0.671053 &   0.640351 &   0.653509 &   \textbf{0.793860} \\ 
              ToeSegmentation2 &   0.546154 &   0.838462 &   \textbf{0.853846} &   0.753846 &   0.746154 &   0.784615 \\
                         Trace &   0.580000 &   0.970000 &   0.970000 &   0.780000 &   0.800000 &   \textbf{0.980000} \\ 
                    TwoLeadECG &   0.554873 &   0.811238 &   0.801580 &   0.956102 &   0.955224 &   \textbf{0.989464} \\ 
                   TwoPatterns &   0.464750 &   0.975000 &   \textbf{0.989750} &   0.555750 &   0.700500 &   0.715750 \\
        UWaveGestureLibraryAll &   0.849525 &   0.831937 &   0.833613 &   0.920715 &   0.911502 &   \textbf{0.943886} \\ 
          UWaveGestureLibraryX &   0.631212 &   0.676438 &   0.706868 &   0.681184 &   \textbf{0.721943} &   0.710497 \\ 
          UWaveGestureLibraryY &   0.548297 &   0.525405 &   0.564768 &   0.611669 &   0.617253 &   \textbf{0.641262} \\ 
          UWaveGestureLibraryZ &   0.537409 &   0.592406 &   0.604132 &   0.642099 &   0.646287 &   \textbf{0.652150} \\ 
                         Wafer &   0.654445 &   0.511032 &   0.649416 &   \textbf{0.988968} &   0.982803 &   0.986210 \\ 
                          Wine &   0.555556 &   0.518519 &   0.574074 &   0.574074 &   0.592593 &   \textbf{0.833333} \\ 
                  WordSynonyms &   0.271160 &   0.344828 &   0.412226 &   0.474922 &   \textbf{0.501567} &   0.474922 \\ 
                         Worms &   0.215470 &   \textbf{0.414365} &   0.408840 &   0.259669 &   0.342541 &   0.337662 \\
                 WormsTwoClass &   0.541436 &   0.591160 &   \textbf{0.651934} &   0.618785 &   0.618785 &   0.649351 \\
                          Yoga &   0.497000 &   0.557000 &   0.574000 &   0.631667 &   \textbf{0.696667} &   0.681000 \\
\midrule
Average accuracy   &   0.610865 &   0.655783 &   0.682124 &   0.705480 &   0.711170 &   \textbf{0.748917} \\
Average arithmetic ranking &   4.964  &  4.488 &   3.297 &   2.988 &   2.666 &   \textbf{1.940} \\
Average geometric ranking  &  4.760 &   4.077 &   2.816 &   2.730 &   2.299 &   \textbf{1.589} \\
Winning times      &   1 &   5 &  18 &   7 &  21 &  \textbf{49} \\
\bottomrule
\bottomrule
\end{longtable}

\clearpage
\subsection{Hyperparameter Grid Search}\label{apx:hyperparameter}

For each of the UCR datasets, we trained our TTN for joint alignment as in \cite{Weber2019}, where 
$N_{\mathcal{P}} \in \{16,32,64\}$,
$\lambda_{\sigma} \in \{10^{-3},10^{-2}\}$,
$\lambda_{s} \in \{0.1,0.5\}$,
the number of transformer layers $\in \{1,5\}$,
scaling-and-squaring iterations $\in \{0,8\}$
and the option to apply the zero-boundary constraint.
The full table of results is available in the Supplementary Materials, and was not included in the manuscript due to the large number of experiments for the 84 UCR datasets. 

\begin{figure*}[!htb]
    \vskip 0.2in
    \begin{center}
    \centerline{\includegraphics[width=0.85\linewidth]{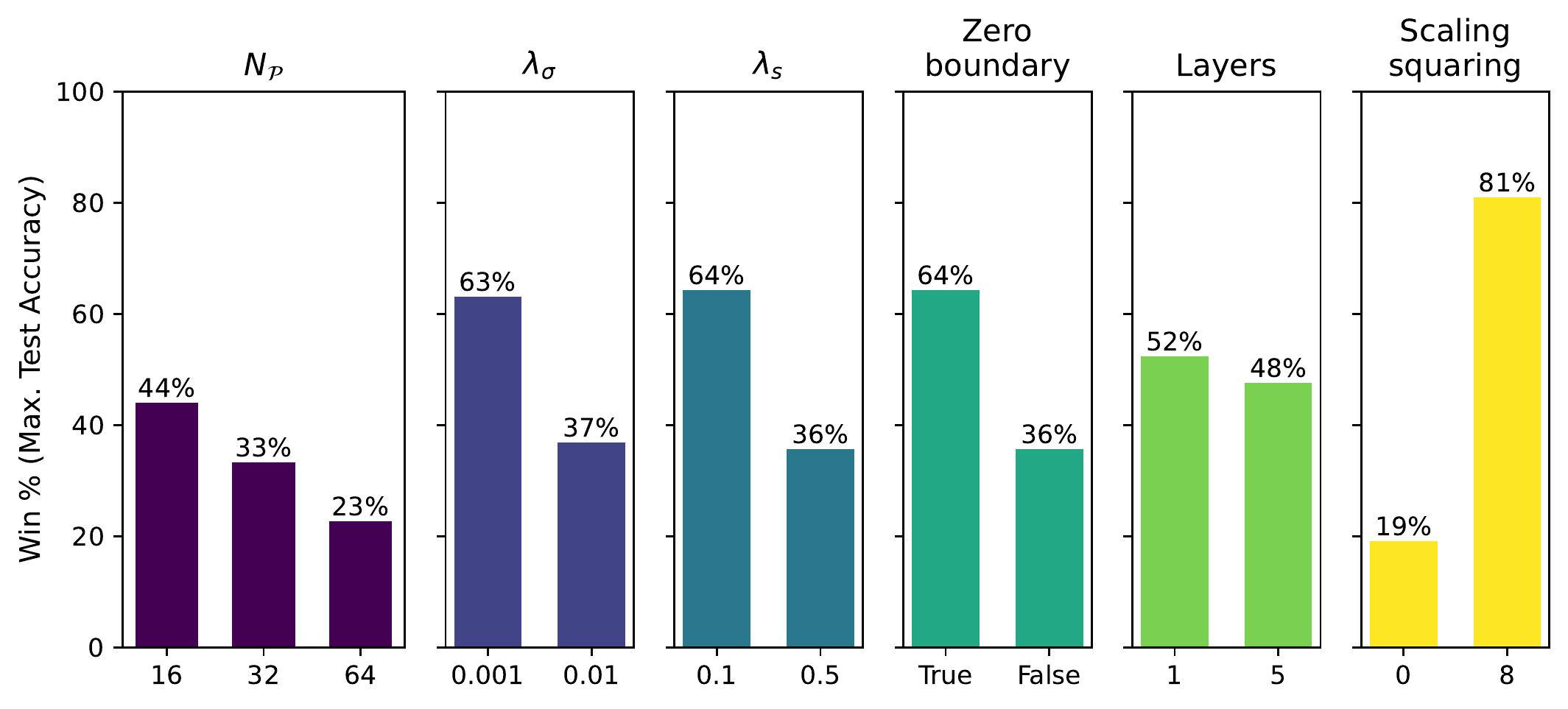}}
    \caption{Winning percentage over 84 UCR datasets for each hyperparameter: 
    a) tessellation size $N_{\mathcal{P}} \in \{16,32,64\}$, 
    b) regularization kernel's overall variance $\lambda_{\sigma} \in \{10^{-3},10^{-2}\}$,
    c) regularization kernel's length-scale $\lambda_{s} \in \{0.1,0.5\}$,
    d) the number of transformer layers $\in \{1,5\}$,
    e) scaling-and-squaring iterations $\in \{0,8\}$
    and f) the option to apply the zero-boundary constraint. A hyperparameter is considered to win in a dataset if it yields maximum NCC test accuracy.
}
    \label{fig:hyperparameter}
    \end{center}
    \vskip -0.2in
\end{figure*}

In total, $84\times3\times2^{5}=8064$ different hyperparameter configurations were tested. To provide a summary of these experiments, we first gathered the hyperparameters that yield optimal (maximum) NCC test accuracy for each UCR dataset. Then we counted how many times each parameter value appears in the optimal NCC accuracy configuration for each of the 84 UCR datasets. \cref{fig:hyperparameter} shows the winning percentage of each hyperparameter value. 
For example, a tessellation size $N_{\mathcal{P}}=16$ appears in $44\%$ of the winning parameter combinations among 84 datasets, while $N_{\mathcal{P}}=32$ and $N_{\mathcal{P}}=64$ get reduced to $33\%$ in $23\%$ respectively. The scaling-and-squaring method appears to be preferable based on the results, with $80\%$ of the datasets using 8 iterations of the method.

\clearpage
\subsection{Ablation Study on Model Expressivity}\label{apx:ablation}

Regarding the expressivity of the proposed model, it is worth noting that the fineness of the partition in CPA velocity functions controls the trade-off between expressiveness and computational complexity. An ablation study is presented in this section to investigate how partition fineness control affects such balance.
We computed the training time and the model accuracy for different values of the tessellation size $N_{\mathcal{P}} \in \{4,16,32,64,128,256\}$.

\begin{figure*}[!htb]
    \vskip 0.2in
    \begin{center}
    \centerline{\includegraphics[width=0.35\linewidth]{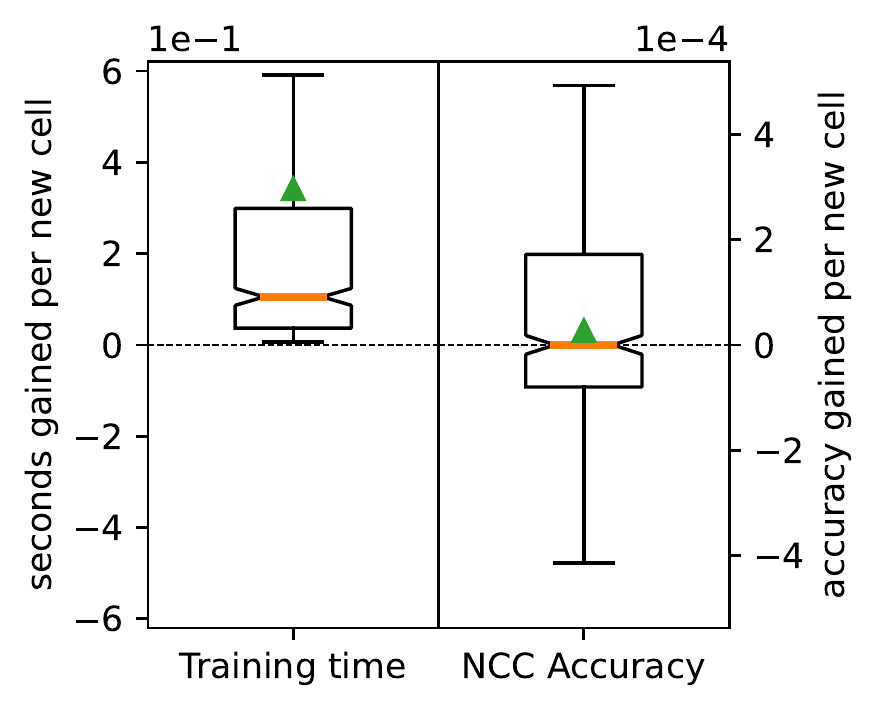}}
    \caption{Ablation study on transformation expressiveness. Distribution of training time (\textbf{left}) and NCC accuracy (\textbf{right}) gained per new cell added to the piecewise transformation.}
    \label{fig:ablation}
    \end{center}
    \vskip -0.2in
\end{figure*}

\cref{fig:ablation} and \cref{tab:ablation} show the results from the ablation study in terms of model accuracy and compute time for all 84 UCR datasets and for different values of $N_\mathcal{P}$. Results show that increasing $N_\mathcal{P}$ lead to longer training times, and large $N_\mathcal{P}$ values can hinder accuracy performance. For instance, on average, going from 4 to 16 cells increments the training time by 1.5\% and increments the NCC accuracy by 0.6962\%. However, going from 16 to 32 cells increments the training time by 1.8\% but reduces the NCC accuracy by 0.0905\%.

\begin{table}[!htb]
    \centering
\begin{tabular}{lll}
	\toprule
	$N_\mathcal{P}$ & \begin{tabular}[c]{@{}l@{}}Training\\Time\end{tabular} & \begin{tabular}[c]{@{}l@{}}NCC\\Accuracy\end{tabular} \\
	\midrule
	4\hfill$\rightarrow$\hfill16 & \trianbox1{cgreen} 1.5\% & \trianbox1{cgreen} 0.69\% \\
	16\hfill$\rightarrow$\hfill32 & \trianbox1{cgreen} 1.8\% & \uptrianbox1{cred} 0.09\% \\
	32\hfill$\rightarrow$\hfill64 & \trianbox1{cgreen} 5.5\% & \uptrianbox1{cred} 0.20\% \\
	64\hfill$\rightarrow$\hfill128 & \trianbox1{cgreen} 16.3\% & \uptrianbox1{cred} 0.24\% \\
	128\hfill$\rightarrow$\hfill256 & \trianbox1{cgreen} 25.0\% & \uptrianbox1{cred} 0.48\% \\ \bottomrule
\end{tabular}
    \caption{Impact of higher expressive CPA functions (more cells) on training time and NCC accuracy.}
    \label{tab:ablation}
\end{table}

\subsection{The Texas Sharpshooter Fallacy}\label{apx:texas}

The Texas Sharpshooter Fallacy is a common logic error that occurs when comparing methods across multiple datasets. In fact, it is not sufficient to have a method that can be more accurate on some datasets unless one can predict on which datasets it will be more accurate.
Therefore, in this section we test whether the \textbf{expected} accuracy gain over another competing method coincides with the \textbf{actual} accuracy gain. The expected accuracy gain and the actual accuracy gain are defined by \cref{eq:texas_expected} and \cref{eq:texas_actual}, respectively.
An expected accuracy gain larger than one indicates that we predict our method will perform better, while an actual accuracy gain larger than one indicates that our method indeed performs better.

\begin{equation}\label{eq:texas_expected}
expected\,gain = \frac{accuracy_{train}(our\,method)}{accuracy_{train}(other\,method)}
\end{equation}
\begin{equation}\label{eq:texas_actual}
actual\,gain = \frac{accuracy_{test}(our\,method)}{accuracy_{test}(other\,method)}
\end{equation}

\begin{figure*}[!htb]
    \vskip 0.2in
    \begin{center}
    \centerline{\includegraphics[width=\linewidth]{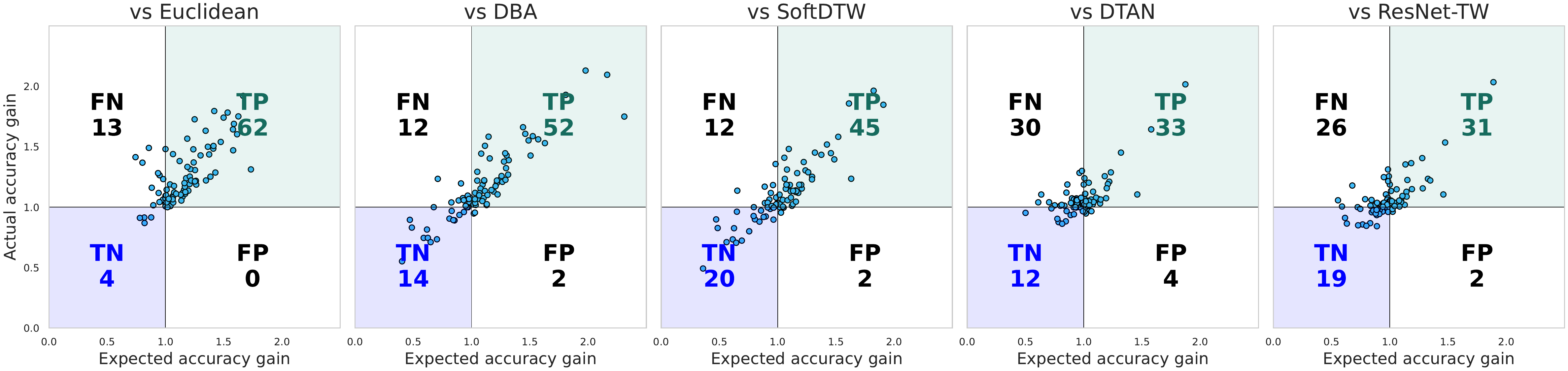}}
    \caption{Texas sharpshooter plot of the proposed method against the other five competing methods respectively. Each point represents an entire dataset.}
    \label{fig:ucr_dataset_texas}
    \end{center}
    \vskip -0.2in
\end{figure*}

As shown in \cref{fig:ucr_dataset_texas}, Texas sharpshooter plot is a convenient tool to visualize the comparison between the expected accuracy gain and the actual accuracy gain on multiple datasets.
Each point represents a dataset which falls into one of the following four possibilities:

\begin{itemize}
    \item \textbf{TP} (True Positive): we predicted that our method would increase accuracy, and it did. This is the most beneficial situation and the majority of points fall into this region.
    \item \textbf{TN} (True Negative): we predicted that our method would decrease accuracy, and it did. This is not truly a bad region since knowing ahead of time that a method will do worse allows choosing another method to avoid the loss of accuracy.
    \item \textbf{FN} (False Negative): we predicted that our method would decrease accuracy but it actually increased. This is also not a bad case even though we might miss the opportunity to improve.
    \item \textbf{FP} (False Positive): we predicted our method would increase accuracy and it did not. Truly the painful region to be, but fortunately not many points fall into it.
\end{itemize}

\subsection{Correlation With Dataset Size}\label{apx:correlation}

\begin{figure*}[!htb]
    \vskip 0.2in
    \begin{center}
    \centerline{\includegraphics[width=\linewidth]{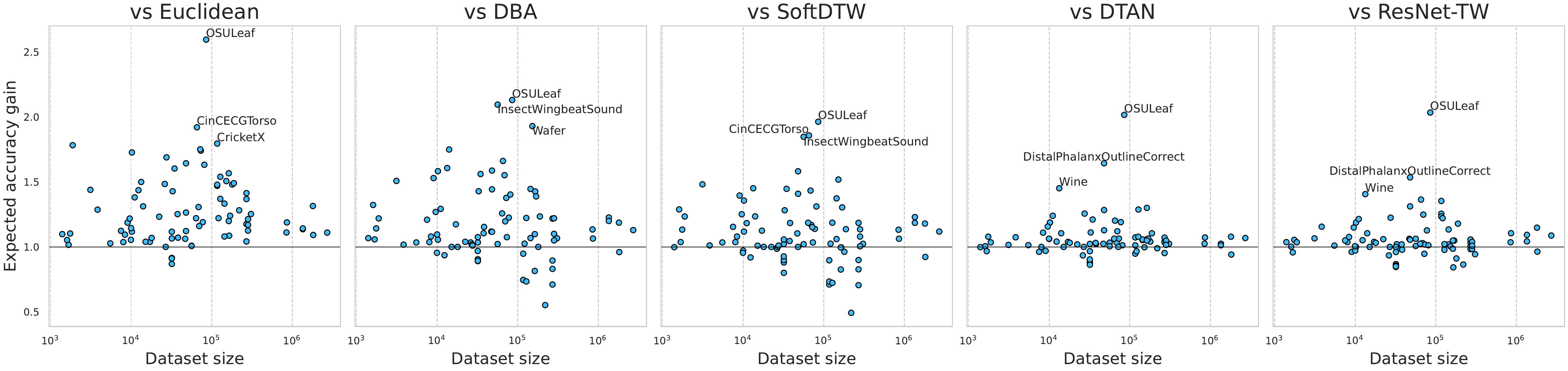}}
    \caption{Dataset size (number of training signals $\times$ time series length) vs expected accuracy gain against the other five competing methods. Each point represents an entire dataset.}
    \label{fig:ucr_dataset_correlation}
    \end{center}
    \vskip -0.2in
\end{figure*}

Finally, we analyze the correlation between the expected accuracy gain and the dataset size. For each competing method, we analyze whether more data availability translates into larger accuracy gains. As visualized in \cref{fig:ucr_dataset_correlation}, there is a lack of correlation between data availability and accuracy improvement across the five competing methods.

\clearpage

\end{document}